\pdfoutput=1
\RequirePackage{fix-cm}%
\documentclass{mmlab-sigs}
\usepackage{longtable,needspace,array,xurl,placeins,multirow,makecell}
\usepackage{utfsym,fontawesome5}
\definecolor{PCwam}{HTML}{4F7FAE}\definecolor{PCvla}{HTML}{294A77}\definecolor{PCagent}{HTML}{6D957E}

\definecolor{M2AMind}{HTML}{416D62}\definecolor{M2AMindT}{HTML}{E6EFEC}
\definecolor{M2AAct}{HTML}{36566F}\definecolor{M2AActT}{HTML}{E5ECF2}
\definecolor{M2ACoord}{HTML}{8A6A2F}\definecolor{M2ACoordT}{HTML}{F3EDE1}
\definecolor{M2AGroup}{HTML}{F1EFE6}   %
\definecolor{M2ASage}{HTML}{EDF1ED}    %
\definecolor{M2AInk}{HTML}{203447}     %
\definecolor{M2ARankA}{HTML}{E2C27E}\definecolor{M2ARankB}{HTML}{EDD7A8}\definecolor{M2ARankC}{HTML}{F6EBD3}
\definecolor{M2AEasy}{HTML}{3F8F5E}\definecolor{M2AMed}{HTML}{B8760F}\definecolor{M2AHard}{HTML}{C1392B}
\definecolor{M2ATrain}{HTML}{8C6E68}\definecolor{M2ATag}{HTML}{C2563C}
\newcommand{\VMind}{\textcolor{M2AMind}{\textbf{Mind}}}
\newcommand{\VAct}{\textcolor{M2AAct}{\textbf{Act}}}
\newcommand{\VMtA}{\textcolor{M2ACoord}{\textbf{Mind2Act}}}
\newcommand{\VMAC}{\textcolor{M2ACoord}{\textbf{Mind-Action Composition}}}
\newcommand{\CMind}[1]{\textcolor{M2AMind}{#1}}\newcommand{\CAct}[1]{\textcolor{M2AAct}{#1}}\newcommand{\CComp}[1]{\textcolor{M2ACoord}{#1}}
\newcommand{\LvEasy}{\textcolor{M2AEasy}{\textbf{Easy}}}
\newcommand{\LvMed}{\textcolor{M2AMed}{\textbf{Medium}}}
\newcommand{\LvHard}{\textcolor{M2AHard}{\textbf{Hard}}}
\newcommand{\LgBest}[1]{{\setlength{\fboxsep}{0.9pt}\colorbox{M2ARankA}{\textbf{#1}}}}
\newcommand{\LgSecond}[1]{{\setlength{\fboxsep}{0.9pt}\colorbox{M2ARankB}{#1}}}
\definecolor{M2APanA}{HTML}{4F7F8C}\definecolor{M2APanB}{HTML}{294B77}\definecolor{M2APanC}{HTML}{B9784A}\definecolor{M2APanS}{HTML}{6D967F}
\newcommand{\PanHead}[2]{\textcolor{#1}{\textbf{#2}}}

\newcommand{\LgZero}{\textcolor{black!35}{0}}
\newcommand{\mtgroup}[2]{\rowcolor{M2AGroup}\multicolumn{#1}{@{}l}{\strut\textbf{\textit{#2}}}}
\newcommand{\mthead}[3]{\cellcolor{#2}\textit{\textcolor{#1}{#3}}}
\newcommand{\rkA}[1]{\cellcolor{M2ARankA}#1}
\newcommand{\rkB}[1]{\cellcolor{M2ARankB}#1}

\definecolor{M2ARand}{HTML}{5E4E86}
\definecolor{M2ALevel}{HTML}{93402F}
\definecolor{M2AReal}{HTML}{2C7C88}
\makeatletter
\newcommand{\axiskeeplines}[1]{\par\begingroup\dimen@\pagegoal\advance\dimen@-\pagetotal
  \ifdim\dimen@<#1\relax\newpage\fi\endgroup}
\makeatother
\newenvironment{axisfinding}[5]{\par\addvspace{7pt plus 1pt}\axiskeeplines{4\baselineskip}\noindent
  {\color{#1}\footnotesize\faIcon{#2}\hspace{0.45em}\sffamily\bfseries #3}%
  \hspace{0.8em}{\footnotesize\itshape\color{black!55}#4}\par\nobreak\vspace{1.2pt}%
  \noindent\textbf{#5}\hspace{0.4em}\ignorespaces}{\par\addvspace{7pt plus 1pt}}

\pdfmapline{+m2apixel <m2apx.enc <m2apixel.ttf}
\usetikzlibrary{arrows.meta,positioning,fit,backgrounds}

\labname{mmlab@SIGS}
\lablockup{\raisebox{-0.3mm}{\includegraphics[height=10mm]{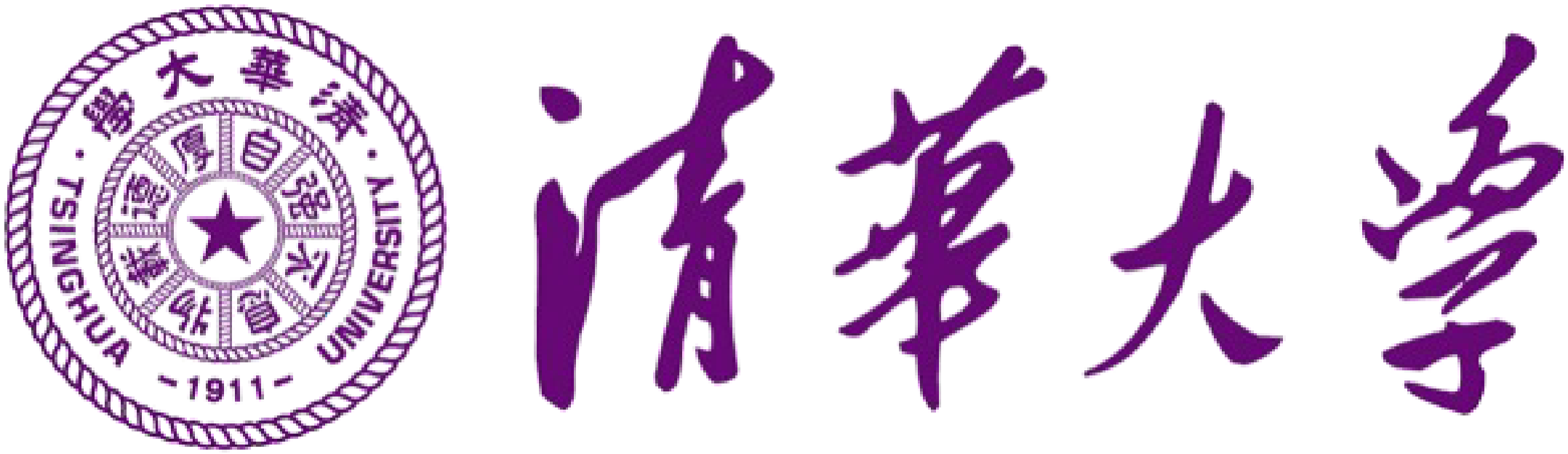}}\hspace{2.6mm}{\color{black!25}\rule[0.6mm]{0.4pt}{8.4mm}}\hspace{2.6mm}\includegraphics[height=9.5mm,trim=7.5bp 8bp 20bp 7.5bp,clip]{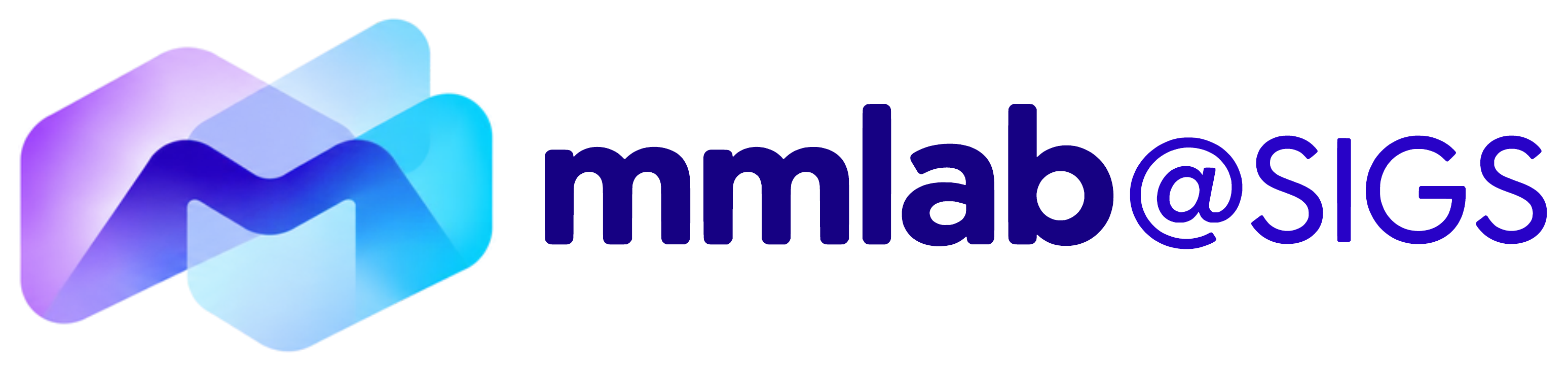}}
\title{Mind2Act World:\\A Diagnostic Benchmark for Embodied Mind, Physical Action,\\and Mind-Action Composition in Robotic Manipulation}
\coverwordmark{assets/cover_wordmark.tex}{\fontsize{18}{21.5}\selectfont A Diagnostic Benchmark for Embodied Mind, Physical Action,\\and Mind-Action Composition in Robotic Manipulation}
\providecommand{\AuMail}{\,{\scriptsize\faEnvelope[regular]}}
\providecommand{\AuSep}{\hspace{0.7em}}
\author{%
\fontsize{9.5}{12.5}\selectfont
\mbox{\textbf{Yuzhi Huang}\textsuperscript{1,14$*\dagger$}}\AuSep
\mbox{\textbf{Jie Wu}\textsuperscript{1$*\dagger$}}\AuSep
\mbox{\textbf{Fanding Huang}\textsuperscript{1$*$}}\AuSep
\mbox{\textbf{Junchao Lin}\textsuperscript{1,14$*$}}\AuSep
\mbox{\textbf{Weichen Zhang}\textsuperscript{1$*$}}\AuSep
\mbox{\textbf{Ziyi Xiong}\textsuperscript{13,14$*$}}\AuSep
\mbox{\textbf{Yang Li}\textsuperscript{12$*$}}\AuSep
\mbox{\textbf{Zhengwei Xie}\textsuperscript{12$*$}}\AuSep
\mbox{\textbf{Zhimin Wang}\textsuperscript{1$*$}}\AuSep
\mbox{\textbf{Zhen Liu}\textsuperscript{1$*$}}\AuSep
\mbox{\textbf{Jiazhen Huang}\textsuperscript{1$*$}}\AuSep
\mbox{\textbf{Yuzheng Zhang}\textsuperscript{12}}\AuSep
\mbox{\textbf{Yuche Liu}\textsuperscript{12}}\AuSep
\mbox{\textbf{Hongyi Yang}\textsuperscript{12}}\AuSep
\mbox{\textbf{Chen Tang}\textsuperscript{8}}\AuSep
\mbox{\textbf{Shixiang Tang}\textsuperscript{3}}\AuSep
\mbox{\textbf{Hakan Aktas}\textsuperscript{2}}\AuSep
\mbox{\textbf{Adam Wu}\textsuperscript{2}}\AuSep
\mbox{\textbf{Duo Wu}\textsuperscript{1}}\AuSep
\mbox{\textbf{Kangye Ji}\textsuperscript{1}}\AuSep
\mbox{\textbf{Shuzhao Xie}\textsuperscript{1}}\AuSep
\mbox{\textbf{Huanan Liu}\textsuperscript{1}}\AuSep
\mbox{\textbf{Weijue Bu}\textsuperscript{11}}\AuSep
\mbox{\textbf{Chenxin Li}\textsuperscript{8}}\AuSep
\mbox{\textbf{Xinyuan Wang}\textsuperscript{6}}\AuSep
\mbox{\textbf{Tianyi Bai}\textsuperscript{7}}\AuSep
\mbox{\textbf{Yinan Mao}\textsuperscript{16}}\AuSep
\mbox{\textbf{Jiawei Wang}\textsuperscript{15}}\AuSep
\mbox{\textbf{Minglei Li}\textsuperscript{15}}\AuSep
\mbox{\textbf{Heng Yu}\textsuperscript{4}}\AuSep
\mbox{\textbf{Ehsan Adeli}\textsuperscript{4$\ddagger$}}\AuSep
\mbox{\textbf{Minchen Li}\textsuperscript{5$\ddagger$}}\AuSep
\mbox{\textbf{Yong-Lu Li}\textsuperscript{10$\ddagger$}}\AuSep
\mbox{\textbf{Cengiz \"{O}ztireli}\textsuperscript{2$\ddagger$}}\AuSep
\mbox{\textbf{Huazhe Xu}\textsuperscript{1$\ddagger$}}\AuSep
\mbox{\textbf{Chao Yu}\textsuperscript{1$\ddagger$}}\AuSep
\mbox{\textbf{Jing Zhang}\textsuperscript{12$\ddagger$}}\AuSep
\mbox{\textbf{Shanghang Zhang}\textsuperscript{9$\ddagger$}}\AuSep
\mbox{\textbf{Weitao Zhou}\textsuperscript{15$\ddagger$}}\AuSep
\mbox{\textbf{Wenwu Zhu}\textsuperscript{1$\ddagger$}}\AuSep
\mbox{\textbf{Cheng Chi}\textsuperscript{12\AuMail}}\AuSep
\mbox{\textbf{Jingyan Jiang}\textsuperscript{1,14\AuMail}}\AuSep
\mbox{\textbf{Zhi Wang}\textsuperscript{1\AuMail}}\\[4pt]
{\fontsize{8.6}{11}\selectfont\bfseries
\mbox{\textsuperscript{1}THU}\quad
\mbox{\textsuperscript{2}UCam}\quad
\mbox{\textsuperscript{3}Oxford}\quad
\mbox{\textsuperscript{4}Stanford}\quad
\mbox{\textsuperscript{5}CMU}\quad
\mbox{\textsuperscript{6}HKU}\quad
\mbox{\textsuperscript{7}HKUST}\quad
\mbox{\textsuperscript{8}CUHK}\\
\mbox{\textsuperscript{9}PKU}\quad
\mbox{\textsuperscript{10}SJTU}\quad
\mbox{\textsuperscript{11}FDU}\quad
\mbox{\textsuperscript{12}RUC}\quad
\mbox{\textsuperscript{13}SZU}\quad
\mbox{\textsuperscript{14}Velabot AI}\quad
\mbox{\textsuperscript{15}Simple AI}\quad
\mbox{\textsuperscript{16}Baidu AI Cloud}\\
\mbox{$^*$Co-first Authors}\quad
\mbox{$^\dagger$Project Leaders}\quad
\mbox{$^\ddagger$Advisors}\quad
\mbox{\AuMail\ Corresponding Authors}
}}
\affiliations{}
\authornotes{}%
\date{18 November 2026}
\paperabstract{\parfillskip=0pt plus 0.65\linewidth\relax\emergencystretch=1.5em\relax
Embodied intelligence aims to enable robots to perform complex physical tasks with human-like adaptability and reliability, which requires three complementary capabilities: \CMind{history-based cognitive reasoning}, \CAct{precise and timely physical execution}, and \CComp{the composition of reasoning and execution through environmental feedback}. 
Yet existing benchmarks often evaluate reasoning and execution within the same tasks, or evaluate them in separate assessments without examining their composition, thus providing limited insight into whether failures arise from reasoning, physical execution, or their composition.
To fill this gap, we introduce \textbf{Mind2Act World}, a diagnostic benchmark comprising 30 simulated manipulation tasks across five real-world application domains. Each task features three difficulty levels and two scene settings (\emph{Clean} and \emph{Random}), yielding 180 configurations. These tasks are organized into three views: \VMind{} emphasizes decisions based on historical observation, \VAct{} assesses precise and timely execution, and \VMtA{} integrates reasoning and physical execution in closed-loop tasks, where action outcomes guide subsequent decisions. By contrasting performance across the three views, Mind2Act World reveals whether a system lacks a capability or fails to compose the capabilities it already has. Furthermore, we develop a diagnosis harness that attributes the failed episodes of three representative systems to \VMind, \VAct, or their composition and summarizes each system in a per-model report called a profile card.
Our evaluation of multimodal agents, vision-language-action models (VLAs), and world-action models (WAMs) shows that competence in a single view does not carry over to Mind2Act tasks. Even the strongest agent succeeds in \CMind{\textbf{52.7\%}} of \VMind{} and \CAct{\textbf{39.0\%}} of \VAct{} episodes but in only \CComp{\textbf{12.0\%}} of \VMtA{} episodes. The strongest agent retains task state but struggles with sustained contact, whereas end-to-end policies execute practiced motions but retain little task state. These findings point to the \CComp{composition} of \CMind{reasoning} and \CAct{execution} through environmental feedback as the main open problem.
}

\paperlinks{\makebox[\linewidth][s]{\fontsize{8.6}{11}\selectfont
\href{https://mind2act-world.com/}{\textcolor[HTML]{2B5C8A}{\faGlobe\enspace\textbf{Website:}\enspace mind2act-world.com}}\hfill
\textcolor[HTML]{22313A}{\faGithub\enspace\textbf{Code:}}\enspace\href{https://github.com/Mind2Act/Mind2Act_World}{\textcolor[HTML]{2B5C8A}{Benchmark}}\hfill
\href{https://mind2act-world.com/benchmark.html}{\textcolor[HTML]{416D62}{\faCubes\enspace\textbf{Tasks}}}\hfill
\href{https://mind2act-world.com/leaderboard.html}{\textcolor[HTML]{A97A16}{\faTrophy\enspace\textbf{Leaderboard}}}\hfill
\href{https://mind2act-world.com/diagnosis.html}{\textcolor[HTML]{B5523B}{\faStethoscope\enspace\textbf{Diagnosis}}}\hfill
\href{https://mind2act-world.com/document.html}{\textcolor[HTML]{6B5A8E}{\faBook\enspace\textbf{Document}}}\hfill
\href{https://mind2act-world.com/community.html}{\textcolor[HTML]{36566F}{\faUsers\enspace\textbf{Community}}}}\vspace{1.2mm}}
\hypersetup{pdftitle={Mind2Act World: A Diagnostic Benchmark for Embodied Mind, Physical Action, and Mind-Action Composition in Robotic Manipulation},pdfauthor={Yuzhi Huang, Jie Wu, Fanding Huang, Junchao Lin, Weichen Zhang, Ziyi Xiong, Yang Li, Zhengwei Xie, Zhimin Wang, Zhen Liu, Jiazhen Huang, Yuzheng Zhang, Yuche Liu, Hongyi Yang, Chen Tang, Shixiang Tang, Hakan Aktas, Adam Wu, Duo Wu, Kangye Ji, Shuzhao Xie, Huanan Liu, Weijue Bu, Chenxin Li, Xinyuan Wang, Tianyi Bai, Yinan Mao, Jiawei Wang, Minglei Li, Heng Yu, Ehsan Adeli, Minchen Li, Yong-Lu Li, Cengiz Öztireli, Huazhe Xu, Chao Yu, Jing Zhang, Shanghang Zhang, Weitao Zhou, Wenwu Zhu, Cheng Chi, Jingyan Jiang, Zhi Wang},pdfsubject={Mind2Act World benchmark and evaluation of agents and VLA policies}}
\coverwordmarkmoon{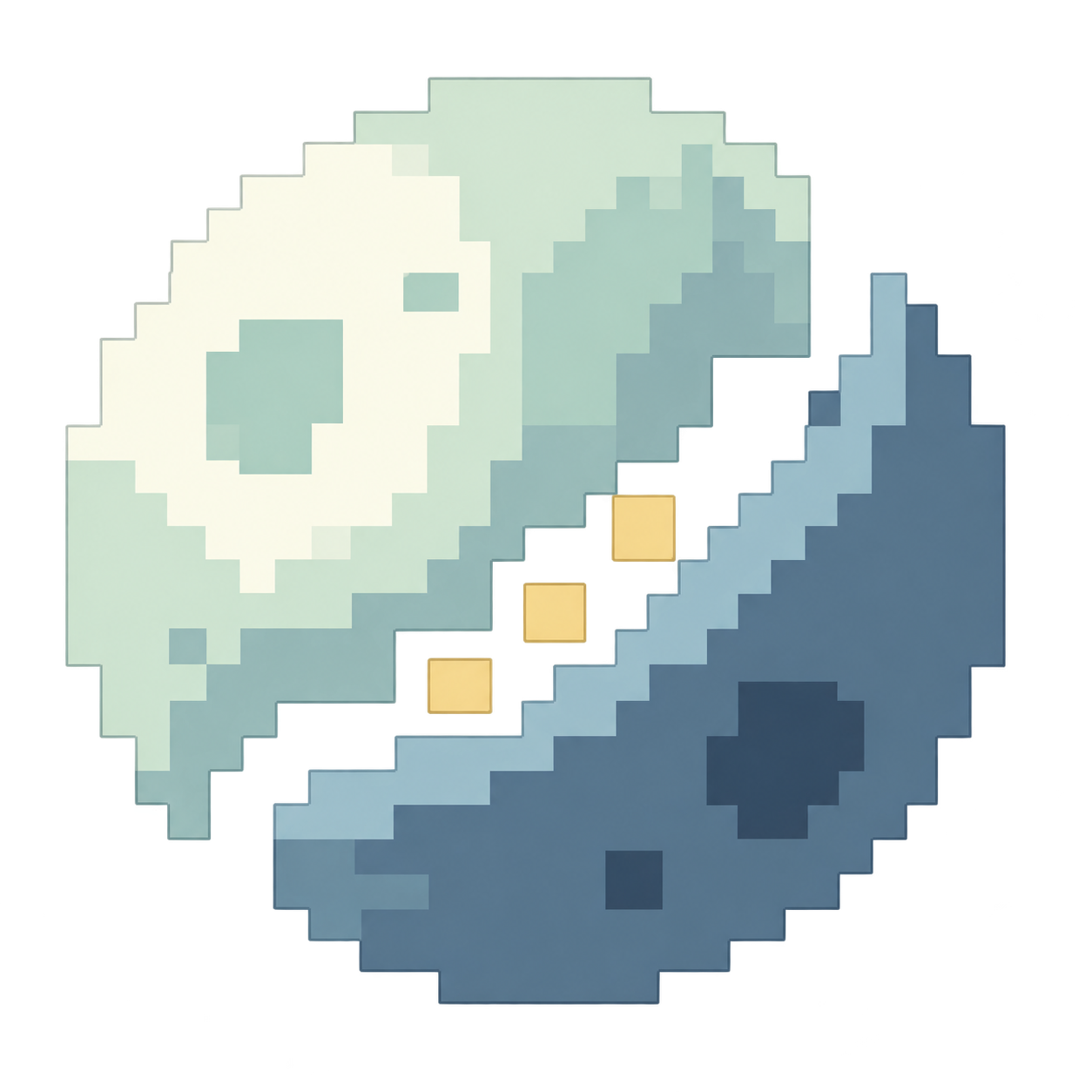}
\paperoverview{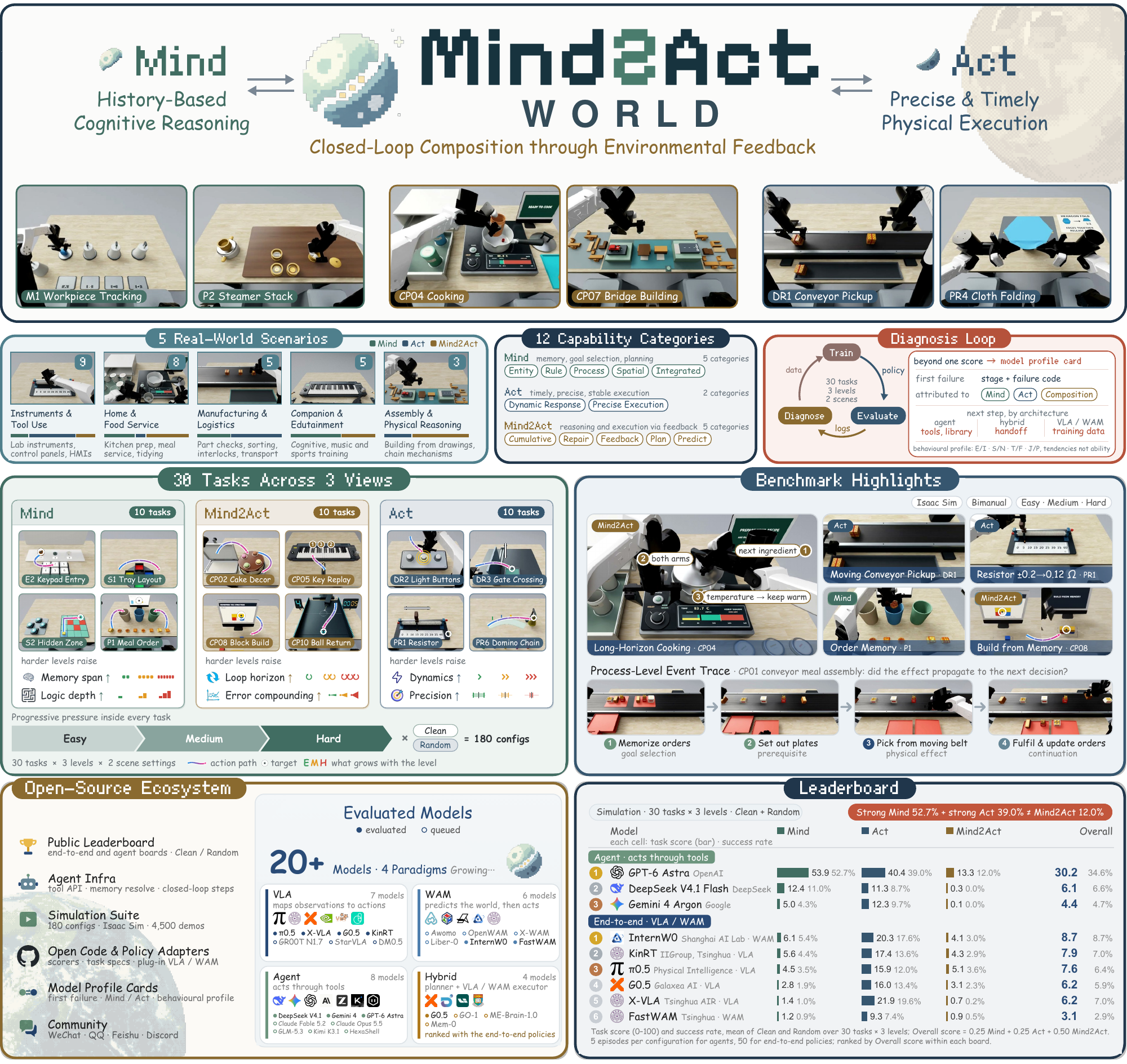}{\textbf{Overview of Mind2Act World.} Mind2Act World is a diagnostic benchmark for embodied mind, physical action and their composition in robotic manipulation. \VMind{} tasks test history-based reasoning, \VAct{} tasks test precise and timely physical execution, and \VMtA{} tasks test their closed-loop composition, where action outcomes guide subsequent decisions. The 30 tasks come from 5 real-world application domains and cover 12 capability categories. Each runs at 3 difficulty levels in \emph{Clean} and \emph{Random} scenes (180 configurations), with 4,500 expert demonstrations for training end-to-end policies. Agents and end-to-end policies share the same tasks, information contract and evaluator, and are ranked on separate leaderboards. A diagnosis loop then attributes the failed episodes of a system to \VMind, \VAct{} or their composition and summarizes them in a model profile card that points to its next improvement. \textbf{Top:} the three views of Mind2Act World. \textbf{Middle:} the application domains, capability categories, tasks and diagnosis loop. \textbf{Bottom:} the open-source ecosystem, the evaluated models and the continuously updated leaderboards.}{fig:report-overview}
\unilogos{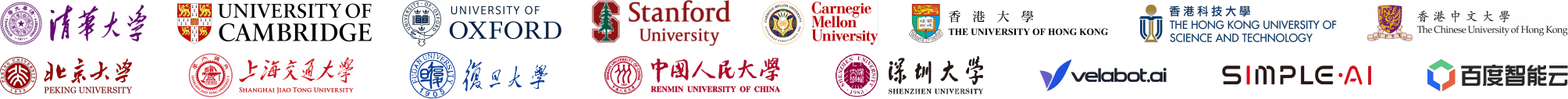}
\newcolumntype{Y}{>{\raggedright\arraybackslash}X}
\renewcommand{\arraystretch}{1.2}
\titlespacing*{\paragraph}{0pt}{0.35em}{0.6em}

\AtBeginDocument{\setlength{\parfillskip}{0pt plus 0.75\textwidth}\setlength{\emergencystretch}{1.5em}}
\makeatletter\g@addto@macro\@floatboxreset{\setlength{\parfillskip}{0pt plus 0.75\textwidth}\setlength{\emergencystretch}{1.5em}}\makeatother
\DeclareCaptionJustification{nolonelyword}{\setlength{\parfillskip}{0pt plus 0.75\linewidth}\setlength{\emergencystretch}{1.5em}}
\begin{document}
\maketitle
\makeatletter\patchcmd{\l@section}{1.0em \@plus\p@}{0.5em \@plus 4\p@}{}{}\pretocmd{\l@subsection}{\ifnum\c@tocdepth>\@ne\addvspace{\z@ \@plus\p@}\fi}{}{}\makeatother
\noindent\vbox to \textheight{{\renewcommand{\baselinestretch}{0.94}\selectfont\tableofcontents}}
\clearpage
\printabstract
\section{Introduction}
\label{sec:scope}
\begin{center}
\small{\bfseries\itshape\color{MMLabBlue}``Knowledge is the beginning of action and action is the completion of knowledge.''}\quad{\color{MMLabBlue}--- Wang Yangming}
\end{center}
\vspace{0.2em}

The central ambition of embodied intelligence is to enable robots to perform complex physical tasks with human-like adaptability and reliability. Studies of human sensorimotor control indicate that adaptive and reliable manipulation involves reasoning over task-relevant information, executing physical actions precisely, and adapting behavior through environmental feedback~\cite{wolpert2011principles,cisek2010neural,shadmehr2010error}. In the cooking task of Fig.~\ref{fig:report-overview}, a robot must \CMind{remember which ingredient to add next}, \CAct{coordinate both arms to add the ingredient}, and \CComp{switch the stove from heating to keep-warm mode based on temperature feedback}. These requirements motivate three capabilities. \VMind{} decides what to do from earlier observations. \VAct{} produces the required physical effect in time. \VMAC{} closes the loop, so that remembered information guides each action and each physical outcome updates the next decision.

Recent embodied systems spanning multimodal agents~\cite{liang2023cap}, vision-language-action models (VLAs)~\cite{kim2024openvla,pi2025pi05}, world-action models (WAMs)~\cite{li2026wallwm,ye2026dreamzero}, and hybrid architectures~\cite{shi2025hirobot,gemini2025robotics15} have substantially advanced robotic manipulation. However, their capabilities in \VMind, \VAct, and \VMAC{} remain poorly understood, which makes it difficult to identify capability bottlenecks and guide further development. Two limitations of existing embodied benchmarks~\cite{libero,robocasa,behavior,robodojo,x2real} explain why. First, conventional benchmarks often assess cognitive reasoning and physical execution within the same tasks and report only whether a task succeeds~\cite{libero,rlbench,robocasa}, so a failure cannot be traced to \VMind{} or \VAct. Second, although recent benchmarks have introduced finer-grained diagnostics and separate assessments of reasoning and physical execution~\cite{robodojo,embodiedplayground,x2real,robocoliseum}, they rarely evaluate \VMAC, particularly whether physical outcomes can effectively inform subsequent decisions. Closing this gap requires tasks that stress one capability while keeping the others present but easy, a judgment of composition based on verified physical outcomes, and a comparison of systems with different action interfaces and inference speeds on equal terms.

\begin{figure}[!htb]\centering
\resizebox{\linewidth}{!}{\input{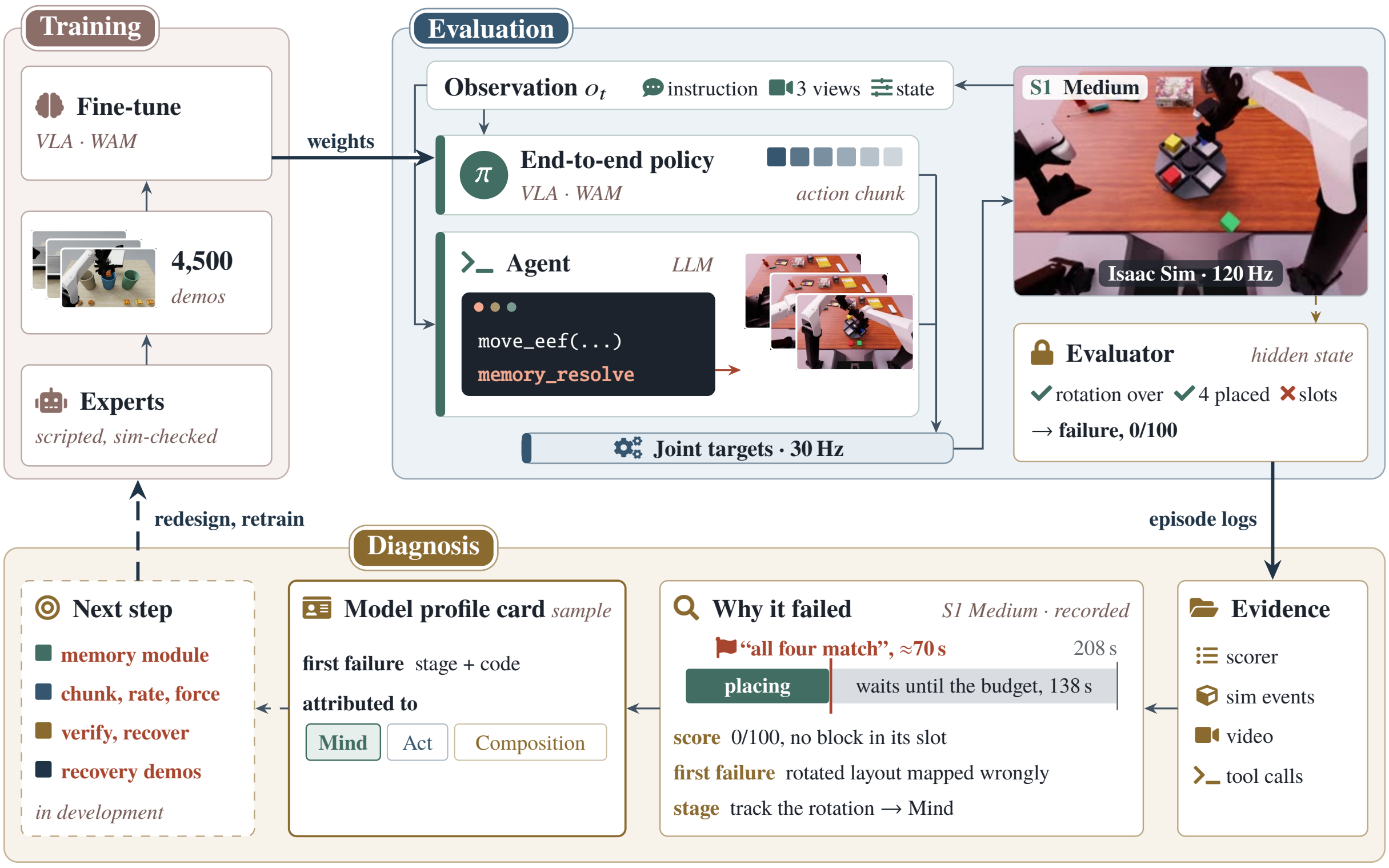}}%
\caption{\textbf{Training, evaluation and diagnosis pipeline.} \textcolor{M2ATrain}{\textbf{Training}}: end-to-end policies are fine-tuned on 4,500 demonstrations generated by scripted experts. \textcolor{M2AAct}{\textbf{Evaluation}}: the policies predict action chunks and the agents call motion tools, while the evaluator decides success from simulator state hidden from both. \textcolor{M2ACoord}{\textbf{Diagnosis}}: episode logs feed a model profile card, which records where each failed episode first breaks down and whether the failure lies in \VMind, \VAct{} or their composition (Section~\ref{subsec:diagnosis-harness}). Dashed elements are the next stage of the loop. The failure codes on a card become concrete changes (Table~\ref{tab:diag-directions}), and the improved model is evaluated again (Section~\ref{sec:limitations}). The evaluation and diagnosis panels trace one recorded GPT-6 Astra episode on S1 Medium, and the card panel is illustrative.}
\label{fig:eval-pipeline}
\end{figure}

To this end, we introduce \textbf{Mind2Act World}, a diagnostic benchmark for \VMind, \VAct, and \VMAC{} in robotic manipulation (Fig.~\ref{fig:report-overview}). Its 30 tasks come from five real-world application domains and are organized into three views of ten tasks each, covering 12 capability categories. Each view emphasizes one demand by design. \VMind{} tasks keep physical operations simple, \VAct{} tasks state the required operation explicitly, and \VMtA{} tasks chain dependent operations so that each physical outcome decides what comes next. Success is judged from physical predicates on the environment, and the evaluator records stages and events within each episode. Each task has three difficulty levels that raise its demands while keeping the objective fixed. Each level runs in a \emph{Clean} and a \emph{Random} scene, which share the same instance but differ in clutter, materials and lighting (180 configurations in total). Agents and end-to-end policies share the same tasks, information contract and evaluator, and 4,500 demonstrations (75.8 hours) are provided for fine-tuning end-to-end policies. A diagnosis harness combines rule-based analysis of the evaluator's stage and event records with an LLM judge to attribute each failed episode of three representative systems to \VMind, \VAct, or their composition. It summarizes each of these systems in a per-model report called a model profile card (Fig.~\ref{fig:eval-pipeline}; Appendix~\ref{subsec:profile-cards}).

Our evaluation on Mind2Act World reveals a clear pattern. Existing architectures are uneven across \VMind{} and \VAct, yet all struggle with \VMAC. The strongest agent retains task state well but struggles to sustain contact, and its timed successes are measured with a simulator that pauses during inference. End-to-end policies respond within a fraction of a second and execute practiced motions, but retain little task state, and the memory of the VLAs largely disappears under scene randomization. Notably, on \VMtA, the strongest system succeeds in only 12.0\% of episodes, far below its 52.7\% on \VMind{} and 39.0\% on \VAct, and no end-to-end policy exceeds 3.6\%. Competence in either capability therefore does not carry over to Mind2Act tasks.

The contributions of this work are summarized as follows:
\begin{itemize}
\item We introduce \textbf{Mind2Act World}, a diagnostic benchmark comprising 30 manipulation tasks grounded in real-world requirements. Organized into three views, it systematically evaluates \VMind, \VAct, and \VMAC.
\item We develop a diagnosis harness that combines rule-based analysis with an LLM judge to identify the stage at which each failed episode first breaks down and to attribute the failure to \VMind, \VAct, or \VMAC. The harness also summarizes three representative models in model profile cards of their failure patterns and behavioral tendencies.
\item We evaluate representative multimodal agents, VLAs, and WAMs on shared tasks and scoring, with separate leaderboards for agents and end-to-end policies. Our results show that systems differ in which of \VMind{} and \VAct{} is weaker, and that all of them struggle with \VMAC.
\end{itemize}

\definecolor{CmpYes}{HTML}{2E7D4F}
\definecolor{CmpNo}{HTML}{D9837A}
\newcommand{\tcmark}{\makebox[0.85em]{\textcolor{CmpYes}{\usym{2713}}}}
\newcommand{\txmark}{\makebox[0.85em]{\textcolor{CmpNo}{$\times$}}}
\definecolor{CmpStat}{HTML}{E3EEF0}
\definecolor{CmpTask}{HTML}{E5F0E6}
\definecolor{CmpCap}{HTML}{E6EBF5}
\definecolor{CmpSys}{HTML}{EEE6F3}
\definecolor{CmpDiag}{HTML}{F3EEE3}
\definecolor{CmpStatT}{HTML}{1F6F78}
\definecolor{CmpTaskT}{HTML}{2F6B35}
\definecolor{CmpCapT}{HTML}{3A4F8C}
\definecolor{CmpSysT}{HTML}{6A3D8C}
\definecolor{CmpDiagT}{HTML}{8C5A1F}
\providecommand{\cmphead}[1]{\rotatebox{90}{#1}}
\providecommand{\cmpgroup}[2]{\textcolor{#1}{\emph{#2}}}
\begin{table}[!t]
\centering
\caption{\textbf{Comparison with existing benchmarks.} Among the compared benchmarks, only Mind2Act World assesses the same task demands both in isolation and in composition, and attributes each failure to reasoning, execution or their composition, with per-model summaries in profile cards. \cmpgroup{CmpStatT}{Scale}: \emph{\# Tasks} in each benchmark's own unit; \emph{Avg.\ demo}, mean length of a training demonstration. \cmpgroup{CmpTaskT}{Task design}: \emph{Workflow}, tasks derived from real-world requirements. \cmpgroup{CmpCapT}{Capability}: \emph{Composition}, decisions conditioned on realized outcomes are tested; \emph{Iso.+Comp.}, the same task semantics are assessed in isolation and in composition. \cmpgroup{CmpSysT}{Systems}: \emph{VLA} includes other end-to-end policies. \cmpgroup{CmpDiagT}{Diagnosis}: \emph{Stage loc.}, failures are localized to the stage at which the episode broke down; \emph{Attribution}, failures are attributed to reasoning, execution or their composition; \emph{Harness}, a dedicated tool analyses episode logs beyond success and score; \emph{Profile card}, a model receives a card summarizing where it fails and its behavioral tendencies. \protect\tcmark{} stated focus, or evaluated in the original paper or on the official leaderboard; \protect\txmark{} not; ``-'' not reported or not applicable.}
\label{tab:related-benchmark-facts}
\begingroup
\fontsize{9}{11}\selectfont
\setlength{\tabcolsep}{3pt}
\renewcommand{\arraystretch}{1.0}
\resizebox{\linewidth}{!}{%
\begin{tabular}{l*{20}{c}}
\toprule
 & \multicolumn{3}{c}{\cellcolor{CmpStat}\cmpgroup{CmpStatT}{Scale}}
 & \multicolumn{6}{c}{\cellcolor{CmpTask}\cmpgroup{CmpTaskT}{Task design}}
 & \multicolumn{4}{c}{\cellcolor{CmpCap}\cmpgroup{CmpCapT}{Capability}}
 & \multicolumn{3}{c}{\cellcolor{CmpSys}\cmpgroup{CmpSysT}{Systems}}
 & \multicolumn{4}{c}{\cellcolor{CmpDiag}\cmpgroup{CmpDiagT}{Diagnosis}} \\
\textbf{Benchmark}
 & \cellcolor{CmpStat}\cmphead{\# Tasks} & \cellcolor{CmpStat}\cmphead{Levels / task} & \cellcolor{CmpStat}\cmphead{Avg. demo (s)}
 & \cellcolor{CmpTask}\cmphead{Memory} & \cellcolor{CmpTask}\cmphead{Dynamic} & \cellcolor{CmpTask}\cmphead{Precision} & \cellcolor{CmpTask}\cmphead{Bimanual} & \cellcolor{CmpTask}\cmphead{Long-horizon} & \cellcolor{CmpTask}\cmphead{Workflow}
 & \cellcolor{CmpCap}\cmphead{Reasoning} & \cellcolor{CmpCap}\cmphead{Execution} & \cellcolor{CmpCap}\cmphead{Composition} & \cellcolor{CmpCap}\cmphead{Iso.+Comp.}
 & \cellcolor{CmpSys}\cmphead{Agent} & \cellcolor{CmpSys}\cmphead{VLA} & \cellcolor{CmpSys}\cmphead{WAM}
 & \cellcolor{CmpDiag}\cmphead{Stage loc.} & \cellcolor{CmpDiag}\cmphead{Attribution} & \cellcolor{CmpDiag}\cmphead{Harness} & \cellcolor{CmpDiag}\cmphead{Profile card} \\
\midrule
\mtgroup{21}{Benchmarks designed mainly for end-to-end policies} \\
RLBench~\cite{rlbench} & 100 & - & - & \txmark & \txmark & \txmark & \txmark & \txmark & \txmark & \txmark & \tcmark & \txmark & \txmark & \txmark & \tcmark & \txmark & \txmark & \txmark & \txmark & \txmark  \\
LIBERO~\cite{libero} & 130 & - & - & \txmark & \txmark & \txmark & \txmark & \tcmark & \txmark & \txmark & \tcmark & \txmark & \txmark & \txmark & \tcmark & \txmark & \txmark & \txmark & \txmark & \txmark  \\
THE COLOSSEUM~\cite{pumacay2024colosseum} & 20 & - & - & \txmark & \txmark & \txmark & \txmark & \txmark & \txmark & \txmark & \tcmark & \txmark & \txmark & \txmark & \tcmark & \txmark & \txmark & \txmark & \txmark & \txmark  \\
Colosseum V2~\cite{morgan2026colosseumv2} & 28 & - & - & \txmark & \txmark & \tcmark & \tcmark & \tcmark & \txmark & \txmark & \tcmark & \txmark & \txmark & \txmark & \tcmark & \txmark & \txmark & \txmark & \txmark & \txmark  \\
RoboLab~\cite{yang2026robolab} & 120 & - & - & \txmark & \txmark & \txmark & \txmark & \tcmark & \txmark & \tcmark & \tcmark & \txmark & \txmark & \tcmark & \tcmark & \tcmark & \tcmark & \txmark & \txmark & \txmark  \\
RoboTwin 2.0~\cite{chen2025robotwin} & 50 & - & 13 & \txmark & \txmark & \txmark & \tcmark & \txmark & \txmark & \txmark & \tcmark & \txmark & \txmark & \txmark & \tcmark & \tcmark & \txmark & \txmark & \txmark & \txmark  \\
BEHAVIOR-1K~\cite{behavior} & 1000 & - & 397 & \txmark & \txmark & \txmark & \tcmark & \tcmark & \tcmark & \tcmark & \tcmark & \txmark & \txmark & \txmark & \tcmark & \txmark & \tcmark & \txmark & \txmark & \txmark  \\
RoboCasa365~\cite{nasiriany2026robocasa365} & 365 & - & 40 & \tcmark & \txmark & \txmark & \txmark & \tcmark & \tcmark & \tcmark & \tcmark & \txmark & \txmark & \tcmark & \tcmark & \tcmark & \txmark & \txmark & \txmark & \txmark  \\
VLABench~\cite{zhang2024vlabench} & 100 & - & - & \txmark & \txmark & \txmark & \txmark & \tcmark & \txmark & \tcmark & \tcmark & \txmark & \txmark & \tcmark & \tcmark & \txmark & \tcmark & \txmark & \txmark & \txmark  \\
GenManip~\cite{gao2025genmanip} & 200 & - & - & \txmark & \txmark & \txmark & \txmark & \tcmark & \txmark & \tcmark & \tcmark & \txmark & \txmark & \tcmark & \tcmark & \txmark & \tcmark & \txmark & \txmark & \txmark  \\
RMBench~\cite{chen2026rmbench} & 9 & - & - & \tcmark & \txmark & \txmark & \tcmark & \txmark & \txmark & \tcmark & \tcmark & \tcmark & \txmark & \txmark & \tcmark & \txmark & \tcmark & \txmark & \txmark & \txmark  \\
RoboMME~\cite{dai2026robomme} & 16 & 3 & - & \tcmark & \tcmark & \txmark & \txmark & \tcmark & \txmark & \tcmark & \tcmark & \tcmark & \txmark & \txmark & \tcmark & \txmark & \txmark & \txmark & \txmark & \txmark  \\
GeoMind~\cite{wang2026geomind} & 20 & - & - & \tcmark & \tcmark & \tcmark & \txmark & \txmark & \txmark & \tcmark & \tcmark & \txmark & \txmark & \tcmark & \tcmark & \txmark & \txmark & \txmark & \txmark & \txmark  \\
MIKASA-Robo-VLA~\cite{cherepanov2026mikasavla} & 90 & - & 28 & \tcmark & \tcmark & \txmark & \txmark & \tcmark & \txmark & \tcmark & \tcmark & \txmark & \txmark & \txmark & \tcmark & \txmark & \txmark & \txmark & \txmark & \txmark  \\
RoboDojo~\cite{robodojo} & 42 & - & 21 & \tcmark & \tcmark & \tcmark & \tcmark & \tcmark & \txmark & \tcmark & \tcmark & \txmark & \txmark & \tcmark & \tcmark & \tcmark & \tcmark & \txmark & \txmark & \txmark  \\
\midrule
\mtgroup{21}{Benchmarks designed mainly for embodied agents or for comparing system types} \\
EmbodiedBench~\cite{yang2025embodiedbench} & 1128 & - & - & \txmark & \txmark & \txmark & \txmark & \tcmark & \txmark & \tcmark & \tcmark & \txmark & \txmark & \tcmark & \txmark & \txmark & \tcmark & \txmark & \txmark & \txmark  \\
RoboCerebra~\cite{han2025robocerebra} & 100 & - & - & \tcmark & \tcmark & \txmark & \txmark & \tcmark & \txmark & \tcmark & \tcmark & \txmark & \txmark & \tcmark & \tcmark & \txmark & \txmark & \txmark & \txmark & \txmark  \\
Embodied Playground~\cite{embodiedplayground} & 410 & - & - & \txmark & \tcmark & \tcmark & \tcmark & \tcmark & \txmark & \tcmark & \tcmark & \tcmark & \txmark & \tcmark & \txmark & \txmark & \tcmark & \txmark & \txmark & \txmark  \\
Embodied First Exam~\cite{embodiedfirstexam} & 1156 & - & - & \txmark & \tcmark & \tcmark & \tcmark & \tcmark & \txmark & \tcmark & \tcmark & \txmark & \txmark & \tcmark & \txmark & \txmark & \tcmark & \txmark & \tcmark & \txmark  \\
X2Real~\cite{x2real} & 44 & - & - & \tcmark & \tcmark & \tcmark & \tcmark & \tcmark & \txmark & \tcmark & \tcmark & \txmark & \txmark & \tcmark & \tcmark & \tcmark & \tcmark & \txmark & \txmark & \txmark  \\
RoboColiseum~\cite{robocoliseum} & 78 & - & - & \txmark & \txmark & \txmark & \tcmark & \txmark & \txmark & \tcmark & \tcmark & \tcmark & \txmark & \tcmark & \tcmark & \tcmark & \tcmark & \txmark & \txmark & \txmark  \\
\midrule
\rowcolor{M2ACoordT}\textbf{Mind2Act World (ours)} & 30 & 3 & 61 & \tcmark & \tcmark & \tcmark & \tcmark & \tcmark & \tcmark & \tcmark & \tcmark & \tcmark & \tcmark & \tcmark & \tcmark & \tcmark & \tcmark & \tcmark & \tcmark & \tcmark  \\
\bottomrule
\end{tabular}}
\endgroup
\end{table}

\section{Related Work}
\label{sec:related}

\subsection{Embodied Policies and Agents}
Robot manipulation policies have evolved from imitation learning~\cite{zhao2023act,chi2023diffusion,ze2024dp3} to large-scale generalist policies. Vision-language-action models such as OpenVLA~\cite{kim2024openvla}, $\pi_{0.5}$~\cite{pi2025pi05}, RDT2~\cite{liu2026rdt2} and G0.5~\cite{galaxea2026g05} improve instruction following and cross-task transfer, while world-action models such as LDA-1B~\cite{lyu2026lda}, Motus~\cite{bi2026motus} and DreamZero~\cite{ye2026dreamzero} couple predictive modeling with control. Agent-based systems take a different route, turning language-model reasoning into executable programs~\cite{liang2023cap} and feeding success detection back into planning~\cite{huang2022innermonologue,liu2023reflect}. Recent work also addresses memory-augmented manipulation~\cite{shi2025memoryvla,torne2026mem,sridhar2025memer}, dynamic manipulation~\cite{black2025rtc,xie2026dynamicvla}, and force- or tactile-aware control~\cite{yu2025forcevla,huang2025tactilevla}.

Despite this progress, each line is usually evaluated where only its own demand is hard: memory methods on tasks with simple execution, dynamic and contact-aware controllers on tasks whose target is given, and agents on tasks where perceiving the outcome is assumed. In practice, a failure may come from holding the wrong task state, missing the intended physical effect, or continuing without checking that the effect occurred. Mind2Act World therefore makes all three demands hard within the same tasks and evaluates agents, VLAs and WAMs on them.

\subsection{Evaluation for Manipulation and Embodied Agents}
Simulation benchmarks provide scalable testbeds for manipulation. RLBench~\cite{rlbench}, CALVIN~\cite{mees2022calvin} and LIBERO~\cite{libero} cover multi-task, language-conditioned and lifelong settings. RoboTwin~2.0~\cite{chen2025robotwin} evaluates bimanual generalization, and THE COLOSSEUM~\cite{pumacay2024colosseum} stresses distribution shift. BEHAVIOR-1K~\cite{behavior} and RoboCasa~\cite{robocasa} pose long household activities. Memory benchmarks such as RMBench~\cite{chen2026rmbench} and RoboMME~\cite{dai2026robomme} isolate history-dependent decisions, the latter with Easy, Medium and Hard levels and timed stopping of moving objects. GeoMind~\cite{wang2026geomind} evaluates spatial reasoning and spatial memory through structured manipulation tasks, with controlled scene variations and diagnostic metrics for spatial understanding and execution failures. RoboDojo~\cite{robodojo} and RobotWorld~\cite{robotworld} diagnose capability dimensions with stage scores or execution traces, and real-world benchmarks such as RoboArena~\cite{atreya2025roboarena} assess deployed policies. For agents built on foundation models, EmbodiedBench~\cite{yang2025embodiedbench} spans high-level and low-level embodied tasks, and Embodied Playground~\cite{embodiedplayground} evaluates closed-loop behavior in which earlier actions set up later prerequisites. X2Real~\cite{x2real} records subtask failure nodes, separates reasoning from manipulation dimensions including dynamic operation, and compares VLA, WAM and agent routes. RoboColiseum~\cite{robocoliseum} reports a gap between task understanding and reliable execution. RLE-Bench~\cite{rle} and Embodied First Exam~\cite{embodiedfirstexam} instead evaluate agents that write and iterate on code to solve robot control and robot learning engineering problems.

Most of these benchmarks entangle reasoning and execution within one task, so a failure cannot be attributed to either, or measure them in separate evaluations, leaving open whether they hold together when one realized outcome sets the conditions for the next decision. Mind2Act World addresses this gap by deriving tasks from real-world requirements and organizing them into Mind, Act and Mind2Act views with graded difficulty, so that the same demands are measured in isolation and in composition across architectures. Table~\ref{tab:related-benchmark-facts} compares Mind2Act World with 21 representative simulation benchmarks. Mind2Act World tasks also have long horizons, with training demonstrations averaging 60.7\,s overall and 99.6\,s in the Mind2Act view. Demonstrations in RoboTwin~2.0, RoboDojo, MIKASA-Robo-VLA and RoboCasa365 average 13 to 40\,s, and only BEHAVIOR-1K, whose tasks are full household activities, has longer ones.

\section{Mind2Act World Benchmark}
\label{sec:benchmark}
\suppressfloats[t]

\begin{figure}[t]\centering
\resizebox{\linewidth}{!}{\input{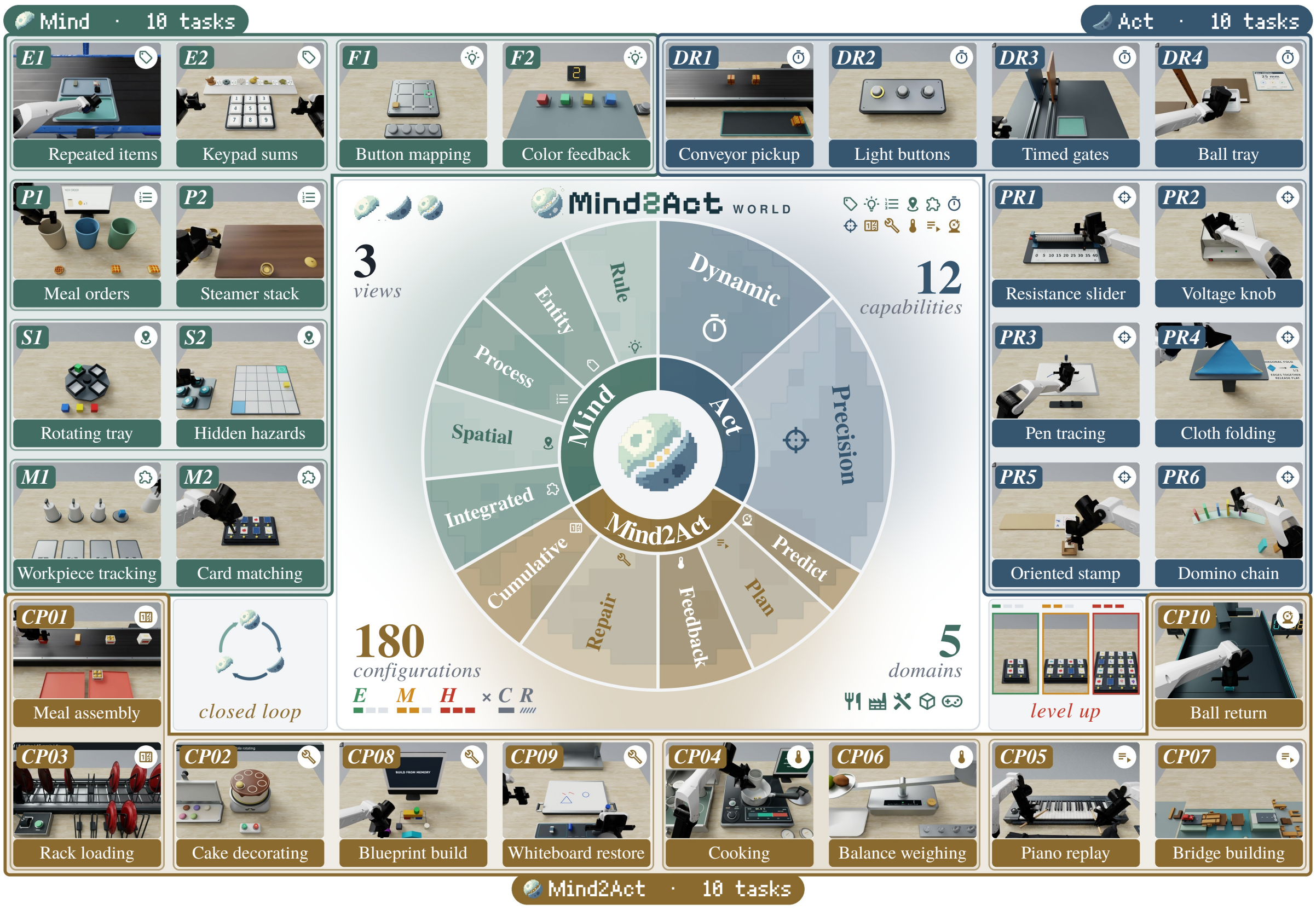}}%
\caption{\textbf{All 30 tasks, grouped by view and capability.} \VMind{} tasks emphasize decisions based on historical observation, \VAct{} tasks precise and timely physical execution, and \VMtA{} tasks the closed-loop composition of the two, in which action outcomes guide subsequent decisions. Each view has ten tasks, and together they cover twelve capability categories and five application domains. The central wheel groups the capability categories by view; the surrounding tiles show each task's ID and scene, framed by view. Each task has three difficulty levels and two scene settings (\emph{Clean} and \emph{Random}), giving 180 configurations.}
\label{fig:skill-diversity}
\end{figure}

Mind2Act World asks whether failures in manipulation tasks arise from incorrect reasoning, imprecise execution, or a poor composition of the two. The benchmark comprises \textbf{30 simulation tasks} on a bimanual ARX X5 platform, organized into \textbf{3 views} that place different demands on a system. \emph{Mind} focuses on history-dependent reasoning, \emph{Act} emphasizes constrained physical execution, and \emph{Mind2Act} focuses on their composition. Each task has Easy, Medium, and Hard levels, and each level is evaluated in a \emph{Clean} and a \emph{Random} scene setting, yielding \textbf{180 configurations}. Fig.~\ref{fig:skill-diversity} presents an overview of our 30 tasks. The following sections describe the design principles (Section~\ref{subsec:design-principles}), task views (Section~\ref{subsec:task-suite-design}), difficulty levels (Section~\ref{subsec:pressure-design}), training data (Section~\ref{subsec:training-data}), and evaluation setting (Section~\ref{subsec:benchmark-evaluation-setting}).

\subsection{Design Principles}
\label{subsec:design-principles}
Four principles guide the benchmark design: (i)~build tasks from real-world requirements, (ii)~evaluate reasoning and execution capabilities separately and in combination, (iii)~verify physical outcomes and record task progress, and (iv)~support evaluation across different model architectures. 

\paragraph{(i) Tasks are built from real-world requirements.}

Embodied agents in industry, homes, and service settings must follow instructions across multiple stages and produce the required physical effects at the right time. We construct tasks in three steps. First, we identify a practical requirement, such as remembering orders after they leave view, maintaining a target instrument reading, intercepting objects on a conveyor, or responding to appliance feedback. We then reproduce this requirement in a tabletop scene and define completion conditions based on physical outcomes. Table~\ref{tab:application-domains} groups the tasks into five application domains: home and food service, manufacturing and logistics, instrument operation and dexterous tool use, assembly and physical reasoning, and companion and edutainment. It also identifies the industrial or service requirement behind each task. In the companion and edutainment domain, structured activities such as card matching and paddle ball represent cognitive and sports training for older adults and children. All thirty tasks are tabletop abstractions of such needs. They isolate the requirement that makes the activity hard for a robot and do not reproduce a deployed workflow in full.
\definecolor{DomMind}{HTML}{416D62}
\definecolor{DomAct}{HTML}{36566F}
\definecolor{DomCoord}{HTML}{8A6A2F}
\definecolor{DomHead}{HTML}{EDF1ED}
\definecolor{DomHeadT}{HTML}{203447}
\newcommand{\tidchip}[2]{{\setlength{\fboxsep}{1.2pt}\colorbox{#1!14}{\textcolor{#1!85!black}{\sffamily\bfseries\scriptsize #2}}}}
\newcommand{\chipM}[1]{\tidchip{DomMind}{#1}}
\newcommand{\chipA}[1]{\tidchip{DomAct}{#1}}
\newcommand{\chipC}[1]{\tidchip{DomCoord}{#1}}
\newcommand{\domain}[2]{\textcolor{DomHeadT}{\faIcon{#1}}\hspace{0.45em}\textbf{#2}}
\begin{table}[!htbp]\centering\small
\caption{\textbf{Application domains of Mind2Act World.} Each task is built from a need in one of five domains. Chips give task IDs colored by view: \chipM{Mind}~\chipA{Act}~\chipC{Mind2Act}.}
\label{tab:application-domains}
\setlength{\tabcolsep}{3pt}
\renewcommand{\arraystretch}{1.35}
\begin{tabularx}{\linewidth}{@{}>{\raggedright\arraybackslash}p{0.34\linewidth}c>{\raggedright\arraybackslash}p{0.24\linewidth}Y@{}}
\toprule
\rowcolor{DomHead}\textcolor{DomHeadT}{\emph{Application domain}} & \textcolor{DomHeadT}{\emph{\#}} & \textcolor{DomHeadT}{\emph{Real-world need}} & \textcolor{DomHeadT}{\emph{Tasks}}\\
\midrule
\domain{utensils}{Home \& Food Service} & 8 & Kitchen prep, meal service, tidying & \chipM{P1} \chipM{P2} \chipM{S1} \chipA{PR4} \chipC{CP01} \chipC{CP02} \chipC{CP03} \chipC{CP04}\\
\domain{industry}{Manufacturing \& Logistics} & 5 & Part checks, traceability, sorting, interlocks, transport & \chipM{E1} \chipM{M1} \chipA{DR1} \chipA{DR3} \chipA{DR4}\\
\domain{tools}{Instruments \& Tool Use} & 9 & Lab instruments, control panels, HMIs, office documents & \chipM{E2} \chipM{F1} \chipA{DR2} \chipA{PR1} \chipA{PR2} \chipA{PR3} \chipA{PR5} \chipC{CP06} \chipC{CP09}\\
\domain{cubes}{Assembly \& Physical Reasoning} & 3 & Building from drawings, limited stock, chain mechanisms & \chipA{PR6} \chipC{CP07} \chipC{CP08}\\
\domain{gamepad}{Companion \& Edutainment} & 5 & Cognitive, music and sports training; puzzle play & \chipM{F2} \chipM{M2} \chipM{S2} \chipC{CP05} \chipC{CP10}\\
\bottomrule
\end{tabularx}
\end{table}

\paragraph{(ii) Capabilities are decoupled, then recoupled.}
Long-horizon tasks often combine memory, planning, precision, and timing, making the source of failure difficult to identify. Mind2Act World evaluates these demands through three views with related task semantics. Mind tasks require only routine execution, so they examine how a system maintains and uses task state. Act tasks specify the required operation to examine its physical execution ability. Mind2Act tasks combine these demands and test composition in both directions. Retained task state guides actions, and the physical outcomes of those actions update that state. The views define observable task demands, not internal policy modules. Each view emphasizes one demand without removing the other. A Mind task still requires routine grasping and placing, and an Act task still requires reading the cue that it responds to. Strong performance on the separate demands with weak performance on their combination is therefore consistent with a composition problem. Because the views contain different tasks and Mind2Act workflows are also longer, view-level differences do not by themselves identify composition failures (Section~\ref{subsec:score-success}). Matched interventions are left to future work (Section~\ref{sec:limitations}).

\paragraph{(iii) Outcomes are physically verified and traced within the episode.}
Task scores depend on simulator states and events. A deposit counts only after an object is released and supported at the target. A dial reading must remain stable after release, and a key strike must produce an effective press. The task records also distinguish incorrect decisions from failed execution, such as selecting the wrong object versus failing to place the correct one. Partial-progress scores and event records show which stages were completed and where the first observable error occurred.

\paragraph{(iv) Evaluation is architecture-agnostic.}
The benchmark supports comparisons across architectures using shared tasks, instances, and scoring rules. End-to-end policies act from observations after training on the shared demonstrations; agents act through declared motion tools. Section~\ref{sec:protocol} specifies what each interface exposes, so that differences in performance can be read against differences in information and control.

\subsection{Task Design}
\label{subsec:benchmark-scope}
\label{subsec:task-suite-design}
\label{subsec:task-mechanisms}

Task design follows the three views introduced above. Mind tests decisions that depend on earlier information (Section~\ref{subsec:mind-dimensions}). Act tests whether a specified operation produces the required physical effect within spatial and temporal constraints (Section~\ref{subsec:act-dimensions}). Mind2Act tests their composition across dependent operations (Section~\ref{subsec:mind2act-dimensions}). Within each view, tasks are grouped by capability. The following subsections explain each category through representative tasks and describe the failures it can reveal. Appendix Table~\ref{tab:task-summary} lists all tasks, their source workflows, categories, and difficulty variables. Appendix~\ref{app:tasks} provides the full specifications.

\subsubsection{Mind: Reasoning over Long-Horizon Dependencies}
\label{subsec:mind-dimensions}
A decision may depend on information that is no longer visible, such as a part on an earlier tray, the effect of an unlabeled control, or the portion of an order already served. These dependencies require a system to retain evidence and update task state throughout an episode. Many end-to-end visuomotor policies use only the current observation or a short history of recent frames~\cite{zhao2023act,chi2023diffusion,kim2024openvla,pi2025pi05}. Recent memory-augmented policies have begun to address this limitation~\cite{li2025cronusvla,shi2025memoryvla,sridhar2025memer,torne2026mem}. Mind tasks examine whether a system can make correct decisions after relevant cues disappear, while requiring only routine execution.

\paragraph{Memory categories.}

The ten Mind tasks cover five memory categories, with two tasks per category. The categories differ in how earlier information supports later decisions. \emph{Entity memory} retains an object's identity or an associated value when its current appearance is insufficient. \emph{Feedback-rule learning} uses interaction to infer and retain a rule or hidden configuration. Examples include learning the effects of unlabeled controls and recovering an arrangement from feedback that reports only the number of correct positions. \emph{Process memory} tracks completed operations and their order, such as fulfilled requests or the original order of disassembled layers. \emph{Spatial memory} retains locations when a reference frame moves or a map is hidden. \emph{Integrated memory} combines identity, location, and order, as in tracking a covered workpiece or matching cards revealed two at a time. Difficulty increases the information to retain, such as bindings, layers, hazard cells, card pairs, or exchanges, while keeping precision demands low.

\paragraph{Representative cases.}

We demonstrate representative cases in Fig.~\ref{fig:process}. In workpiece tracking and route replay (M1), the robot remembers the target location and station route, tracks the hidden target through cover exchanges, and retrieves it to visit the required stations in order. This tests integrated memory of identity, location, and sequence. In steamer-stack order restoration (P2), the robot observes each basket's contents as it unstacks the baskets, then restores their original order and replaces the lid. On Hard, distractor baskets have identical exteriors but different contents. The robot must retain the original layer order and select the correct baskets, testing process memory.

\paragraph{What Mind diagnoses.}
Mind task records distinguish decision errors from execution errors, such as choosing the wrong item versus failing to place the correct one. Because execution is routine, an error on a Mind task usually comes from the task state a system maintains.

\subsubsection{Act: Dynamic Response and Precise Physical Execution}
\label{subsec:act-dimensions}
Correct decisions must be translated into physical effects that meet spatial and temporal constraints. Common examples include intercepting conveyor objects, passing through cyclic gates, maintaining instrument readings, carrying loose loads, and folding fabric. Act tasks specify \emph{what} to do and evaluate \emph{how} it is executed. Act provides a reference for execution demands that also occur in Mind2Act workflows, such as conveyor pickup in meal assembly. This allows performance on separate operations to be compared with performance in a dependent workflow.

\paragraph{Execution categories.}
\emph{Dynamic response} concerns the timing of an action's effect relative to a changing scene. Examples include intercepting objects before they leave the reachable region, pressing a button within a cue window, and passing through a cyclic gate before it closes. Dynamic stability adds a related demand. The robot must control its own motion without disturbing a carried load, such as loose balls on a tray. Timing follows the simulation clock, which pauses during inference. \emph{Precise physical execution} concerns spatial accuracy, contact, and stability. It includes setting instrument readings, maintaining tool contact along a path, controlling stamp orientation, folding cloth that remains flat after release, and spacing dominoes to complete a physical chain to a bell. Difficulty increases through shorter time windows, additional obstacles or loads, tighter tolerances, and less controllable materials or shapes.

\paragraph{Representative cases.}
Cases are demonstrated in Fig.~\ref{fig:process}. In domino chain to a bell (PR6), the robot places dominoes upright along a curved band and pushes the first to trigger a chain that rings a bell. This requires precise spacing and orientation so that contact propagates through the chain. Difficulty increases with more dominoes, greater height variation, and a larger band span. In tray carrying with loose balls (DR4), the robot carries loose balls on a tray past obstacles while keeping each ball within its marked circle. This tests dynamic stability, as the robot must control tray motion to preserve the positions of the balls throughout transport.

\paragraph{What Act diagnoses.}
Act tasks assess endpoint accuracy, stability after release, contact continuity, and timing. By limiting long-horizon reasoning demands, they provide a reference for interpreting execution failures in Mind2Act workflows.

\subsubsection{Mind2Act: Composition across Dependent Operations}
\label{subsec:mind2act-dimensions}
Mind2Act evaluates how reasoning and execution interact across a workflow. Retained information, such as an order, layout, or recipe sequence, determines the next operation and its execution requirements. Each physical outcome then changes the task state on which later decisions depend. A system may handle the separate demands yet struggle to combine them over time. Mind2Act therefore focuses on \emph{outcome-conditioned continuation}, that is, choosing subsequent actions from the state produced by earlier operations. Each task combines a memory or reasoning demand with a physical execution demand, as listed in Appendix Table~\ref{tab:task-summary}.

\paragraph{Composition mechanisms.}
The tasks cover five composition mechanisms. In \emph{cumulative progress} (CP01, CP03), remaining demand or slot occupancy determines the next operation. Progress advances only after successful delivery or stable plate placement. In \emph{state-dependent prerequisites and repair} (CP02, CP08, CP09), later operations require a valid physical state. The robot may need to stop a turntable before covering a cake, rebuild a misplaced base to match a remembered blueprint, or redraw original strokes removed during erasing. In \emph{feedback-driven adjustment} (CP04, CP06), temperature or balance-beam deflection guides the next operation. In \emph{sequence and plan execution} (CP05, CP07), a remembered sequence or a construction plan determines the order of operations. Progress requires effective strikes or stable placements; CP07 also requires a toy car to cross the completed bridge. In \emph{predictive decision under time pressure} (CP10), the robot applies a fixed color rule to each incoming ball: return a white ball and let a red ball pass. It must select and time its response before the ball arrives.

\paragraph{Representative cases.}
In rotating-cake decoration (CP02), the robot remembers a briefly shown macaron layout and reproduces it on a rotating cake. It must stop the turntable and verify that the cake is stationary before placing the cover. In temperature-controlled cooking (CP04), the robot adds ingredients in the remembered recipe order, then uses temperature feedback to switch from HEAT to KEEP WARM within the required time window. Bridge building for a toy car (CP07) requires selecting suitable parts and executing a construction plan. The assembled bridge must remain stable after release and support the car as it crosses the gap, linking plan execution to a physical test of the completed structure. In whiteboard restoration (CP09), the robot compares the current board with a remembered drawing to identify added strokes. It must erase these strokes and redraw any original content removed during erasing, so that the final board matches the remembered colors, shapes, and layout.

\paragraph{What Mind2Act diagnoses.}
Mind2Act examines whether a system uses retained information to guide execution and updates that information from physical outcomes. Comparisons across views, supported by stage and event records, help identify composition failures. 

\begin{figure*}[t]\centering
\resizebox{\linewidth}{!}{\input{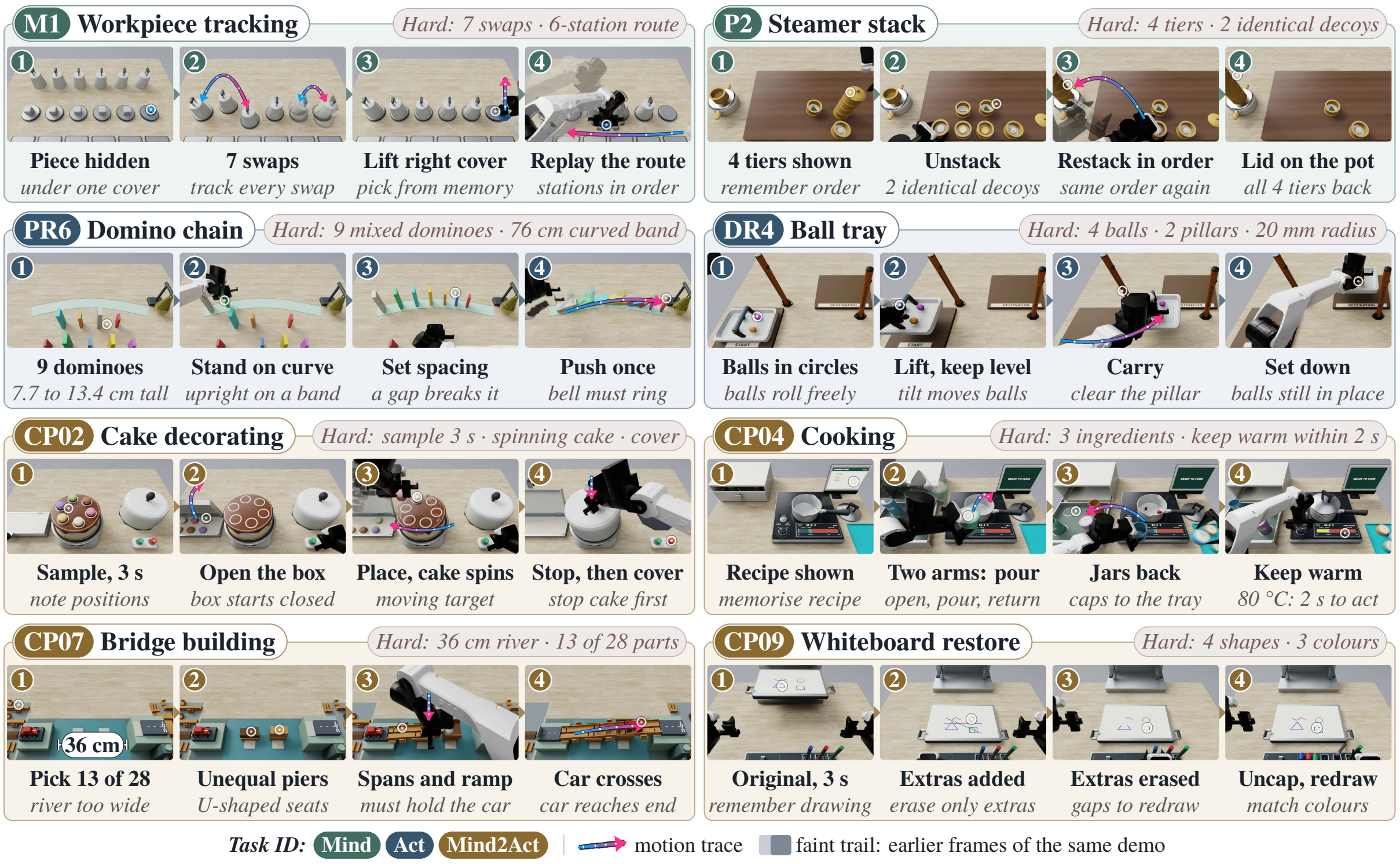}}
\caption{\textbf{Difficult steps inside eight Hard-level tasks.} Each panel follows one Hard-level demonstration through four intermediate steps and shows what makes each step hard: the bold label under a frame names the operation, and the italic line the demand that makes it difficult. A failure at any of these steps blocks the stages after it. The color of the task ID gives its view, and the tag beside the task name the Hard-level setting; Fig.~\ref{fig:all-tasks-difficulties} compares the levels themselves.}
\label{fig:process}
\end{figure*}

\paragraph{Difficulty inside the process.}
Task difficulty often arises during intermediate operations. Fig.~\ref{fig:process} shows four steps from each of eight Hard-level demonstrations, two Mind, two Act and four Mind2Act tasks. At Hard, M1 involves seven cover exchanges, two of them irrelevant, and P2 involves four steamer tiers and two identical decoys. PR6 involves nine dominoes of mixed sizes along a 76\,cm curved band, and DR4 involves four loose balls, two pillars and a marked circle of 20\,mm radius. CP02 shows its sample for 3\,s, CP04 involves bimanual jar handling and a switch from HEAT to KEEP WARM within 2\,s of the pot reaching 80\,\textdegree C, CP07 uses 13 of 28 parts across a 36\,cm river, and CP09 requires uncapping pens with both hands. Failures in these intermediate operations can prevent later stages from succeeding, so we report stage progress alongside final success (Section~\ref{subsec:protocol-metrics}).

\subsection{Difficulty Design}
\label{subsec:pressure-design}
\begin{figure}[t]\centering
\resizebox{\linewidth}{!}{\input{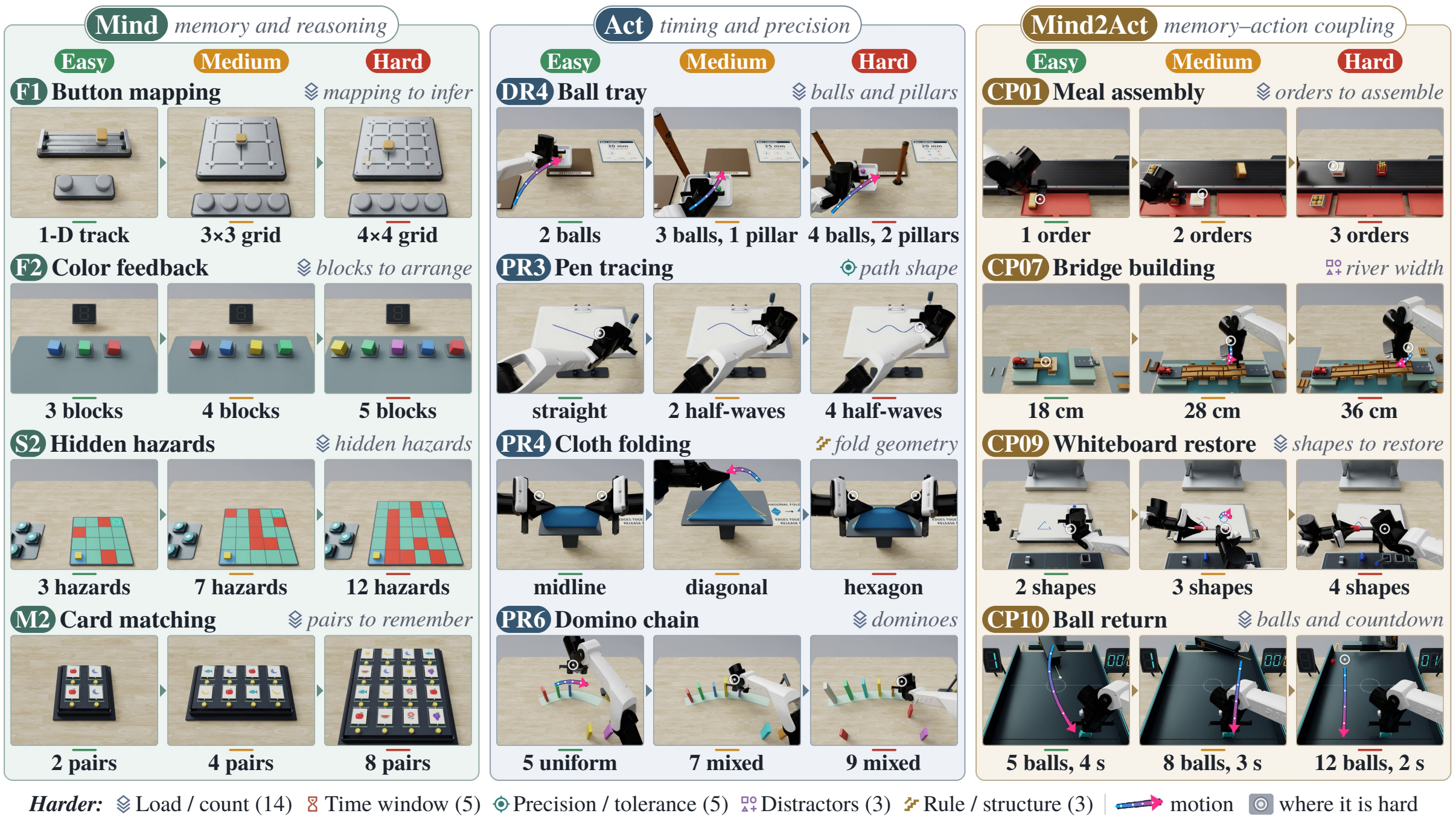}}%
\caption{\textbf{Easy, Medium and Hard levels of twelve tasks.} Four tasks per view at \LvEasy, \LvMed{} and \LvHard. A harder level keeps the goal and raises one or more task demands: \VMind{} tasks add more to remember or infer, \VAct{} tasks require more spatial accuracy and physical stability, and \VMtA{} tasks lengthen the workflow or shorten the response window. The label under each frame gives the parameter that changes, and the icon beside each task name gives the kind of demand (key below the panels; its counts cover all 30 tasks).}
\label{fig:all-tasks-difficulties}
\end{figure}

Aggregate scores can obscure the conditions under which performance declines. We therefore define three difficulty levels for each task. The objective remains fixed, while memory load, reasoning demands, precision, timing, or workflow dependencies increase. \emph{Easy} tests the core requirement under the least demanding conditions. \emph{Medium} increases the information to retain, the required precision, or the number of operations. \emph{Hard} introduces greater demands through tighter constraints, longer dependent sequences, or combinations of these factors. The levels define a progression within each task rather than a common numerical scale across tasks. Comparing success and partial-progress scores across the three levels shows how performance on the same task changes as its demands increase. Fig.~\ref{fig:all-tasks-difficulties} illustrates these variations. Mind tasks increase the information to retain, such as unknown mappings, blocks, hidden hazards, or card pairs, while requiring only routine execution. Act tasks vary physical demands, including load, path shape, fold geometry, and domino count. Mind2Act tasks vary orders, river width, shapes, or balls to increase the number and dependencies of operations. Appendix~\ref{app:tasks} specifies every task and difficulty level.

\subsection{Training Data and Collection}
\label{subsec:training-data}
To compare end-to-end policies fairly, we provide expert demonstrations collected under one protocol for every task and level. Section~\ref{subsec:data-collection} describes how they are generated and checked, and Section~\ref{subsec:data-statistics} summarizes the resulting data.

\subsubsection{Data Collection and Quality Control}
\label{subsec:data-collection}
Each episode has a setup defined at three levels. The task determines the objective and rules, the difficulty level determines the demands, and episode parameters specify variations such as object positions, orders, or target readings. Task scripts define the operation sequence and end-effector targets. Inverse kinematics and motion planning convert these targets into joint trajectories. Quality control checks physical completion against simulator states and events. For example, CP01 requires the requested items to be released into the correct tray and remain supported. Recording checks cover missing or corrupted frames, invalid state or action values, timestamp alignment, and duplicates. Demonstrations must also satisfy task-specific qualification conditions to provide clear examples for learning. A DR3 demonstration must retain at least one second of withdrawal margin, and an M2 demonstration must use only information revealed by actual card flips. These conditions are stricter than task success requirements and apply to demonstration selection rather than policy scoring. Every retained demonstration passes the evaluator used at test time. The collection therefore also shows that each task and level can be completed on this platform by a scripted controller with access to simulator state. Demonstration and evaluation seeds are disjoint.

\subsubsection{Data Statistics}
\label{subsec:data-statistics}
Each episode records the task instruction, RGB images from a head camera and two wrist cameras at $640\times480$, and joint states and targets. Physics runs at 120\,Hz, and the recorded data are sampled at 30\,Hz. Across the thirty tasks, the dataset contains \textbf{50 episodes per task and difficulty level}, resulting in \textbf{4,500 episodes}, \textbf{8.19 million frames per camera}, and \textbf{75.8 hours} of recordings, as detailed in Appendix Table~\ref{tab:training-data-summary}. Mind and Mind2Act episodes average 58.5 and 99.6 seconds, compared with 23.9 seconds for Act. These averages are approximately 2.5 and 4.2 times the Act average, respectively. Hard Mind2Act episodes average 127.6 seconds.

\subsection{Evaluation Setting}
\label{subsec:benchmark-evaluation-setting}
\begin{figure}[t]
\centering
\resizebox{\linewidth}{!}{\input{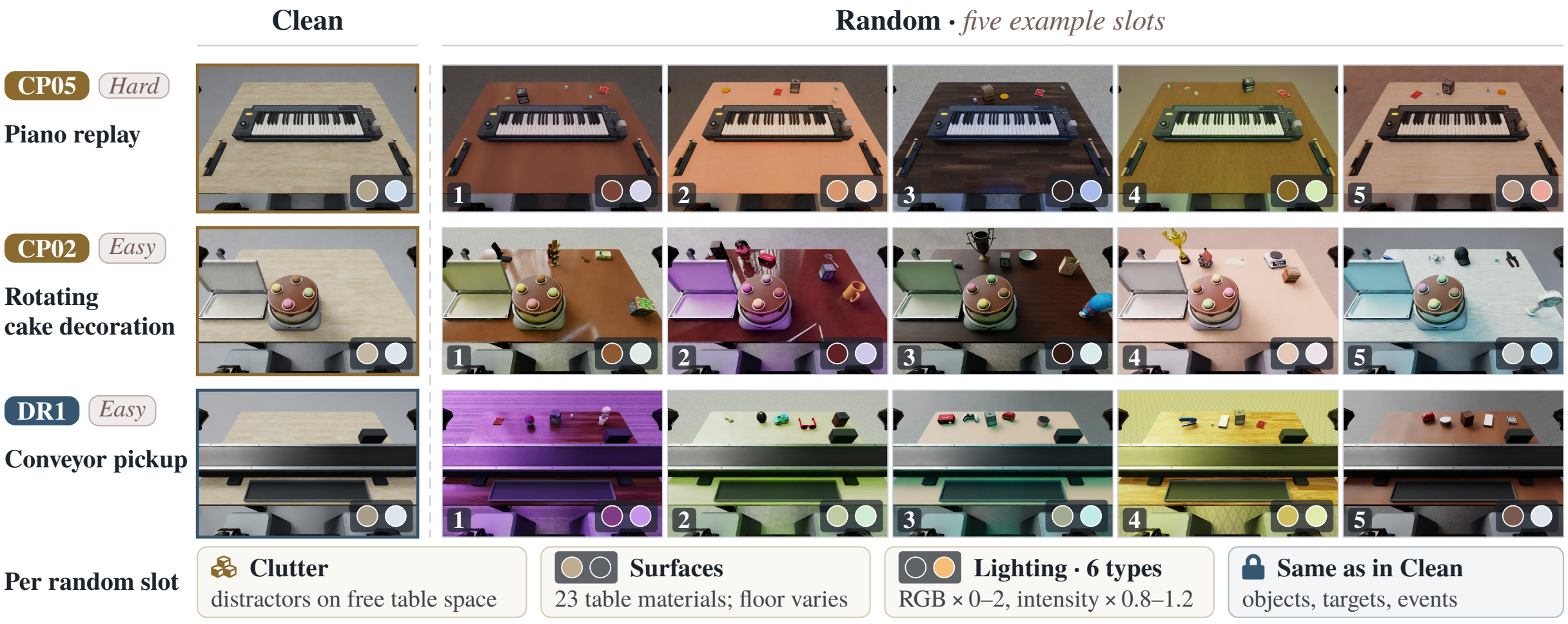}}
\caption{\textbf{Examples of the Clean and Random settings.} Every task is evaluated in both settings; shown are the first head-camera frames of three example tasks. \emph{Clean} scenes use default materials and lighting without clutter, and each \emph{Random} slot adds clutter and varies surfaces and lighting (dots: table color, light tint) while keeping the task's objects, targets and events.}
\label{fig:randomization-reference}\label{fig:clean-random}
\end{figure}
Policies are evaluated through closed-loop interaction with the environment. Each episode starts from a predefined test instance, and the system selects actions using updated observations. Execution ends on task completion, task-defined failure, system termination, or budget exhaustion. Each configuration is defined by a task, a difficulty level, and a scene setting. An end-to-end policy runs all 50 instances of each configuration, and an agent runs a shared subset of 5. Instance parameters are fixed in advance and shared across systems within the same evaluation protocol to control variation between runs. 

We report two metrics. \emph{Final success rate} is the fraction of valid episodes that meet all completion conditions. \emph{Task score} ranges from 0 to 100 and measures partial progress under task-specific rules. High task scores with low final success indicate that a system completes parts of a task but often fails to meet all requirements. This distinction supports analysis of incomplete workflows in Mind2Act. Both metrics are computed from simulator states and events. Invalid or missing runs are recorded separately. Section~\ref{sec:protocol} specifies the interfaces, timing, scoring, and aggregation rules.

\subsubsection{Episode Variation and Randomization}
\label{subsec:instance-variation}
Episodes share the same objective, rules, and difficulty level but vary task-specific initial states and parameters. These include object layouts, initial and target states such as end-effector poses, and event sequences such as piano-key order. Some tasks also vary continuous and hidden parameters. In CP10, launcher position and aim are sampled continuously, while the task seed determines the order of red and white balls. In PR5, the target angle is sampled from a set defined for each difficulty level. The sheet displays a target arrow without revealing the numerical angle or tolerance. These variations test whether a system can adapt its behavior rather than replay a memorized trajectory. Using shared test instances supports controlled comparisons. Each task and difficulty level is evaluated in two scene settings, which we treat as separate configurations, giving 180 configurations in total (30 tasks $\times$ 3 levels $\times$ 2 settings). The \emph{Clean} setting fixes scene appearance. The \emph{Random} setting uses the same task instances and additionally varies clutter, table and floor materials, and lighting as depicted in Fig.~\ref{fig:clean-random}. Because the two settings share task instances, comparing them isolates the effect of visual and scene perturbation from that of task demands. Appendix~\ref{app:clean-random} details how the two settings are generated.

\FloatBarrier
\section{Technical Implementation and Evaluation Protocol}
\label{sec:protocol}

Systems that act through different interfaces must face the same scenes, the same clock and the same notion of success. Building on the task design in Section~\ref{sec:benchmark}, this section describes how policies connect to the tasks, how an episode is executed and timed, and how its outcome is scored and protected against shortcuts. Appendix~\ref{app:impl} collects the implementation details needed for reproduction.

\subsection{Platform and Policy Interfaces}
\label{subsec:simulation-platform}
\label{subsec:environment-interface}
\begin{figure}[t]\centering
\resizebox{\linewidth}{!}{\input{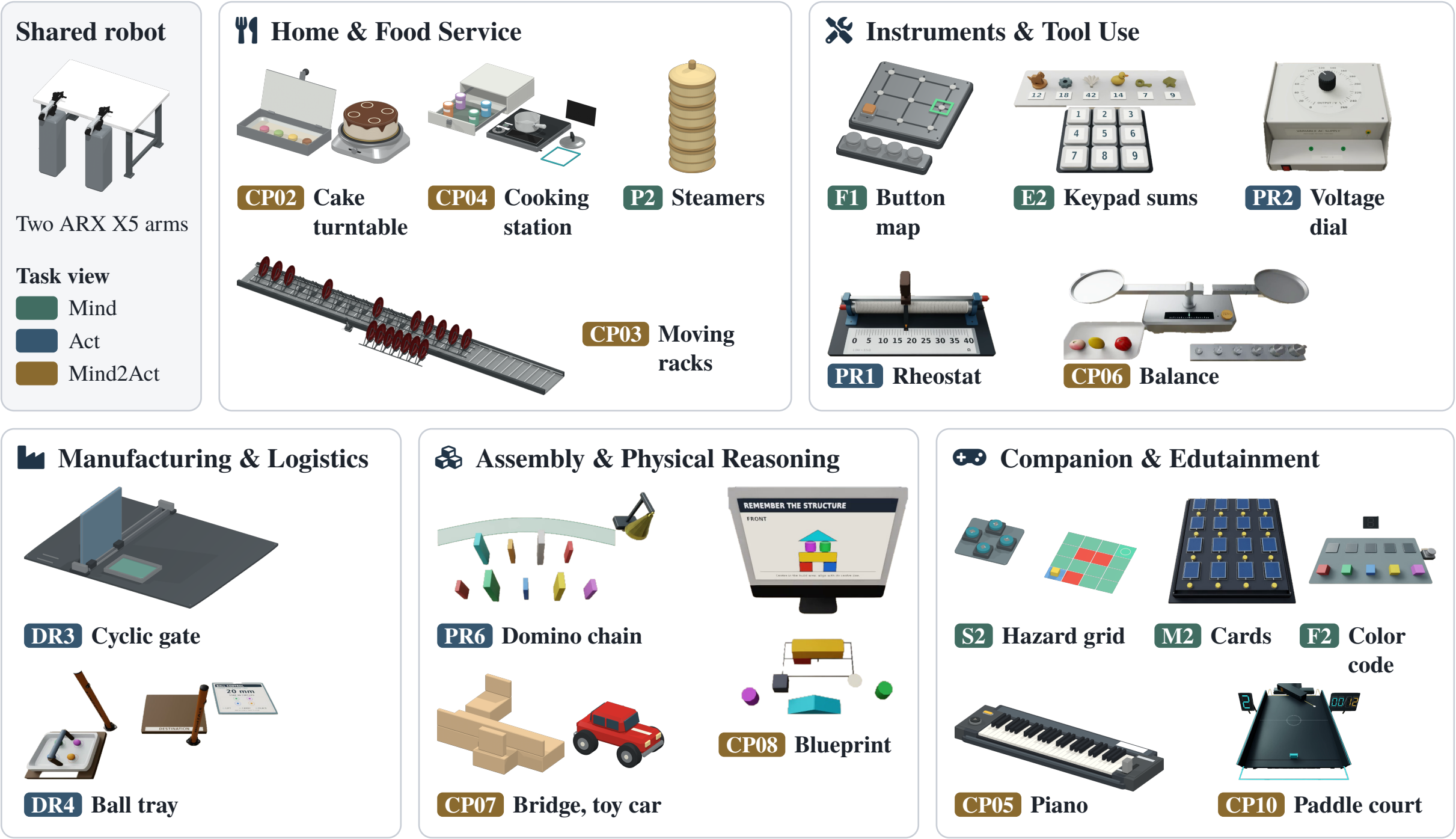}}%
\caption{\textbf{Task setups grouped by application domain.} Nineteen of the thirty tasks are shown with the objects and equipment each one uses. All thirty tasks are performed by the same two-arm ARX X5 robot.}
\label{fig:assets-parallel}
\end{figure}

Comparing agents and end-to-end policies requires a platform that both can drive and that produces the same physical effects for both. We implement Mind2Act World in \textbf{Isaac Sim} with two ARX X5 arms and parallel grippers. A fixed head camera captures the scene layout, while two wrist cameras capture local gripper-object interactions. RGB images are rendered at $640\times480$ for all cameras. Task scenes combine shared robot and workspace assets with task-specific objects and mechanisms as depicted in Fig.~\ref{fig:assets-parallel}. Their geometry, motion, and contact properties are configured to reproduce the physical effects required by each task.

All systems follow a common information contract detailed in Appendix~\ref{app:impl}. Policies receive the task instruction, selected RGB views, and declared robot state. History access is limited to information previously exposed within the same episode. Hidden simulator state is accessible only to the evaluator. This includes unannounced parameters used for scoring, such as the numerical target angle and tolerance in PR5, gate timing in DR3, and allowable ball drift in DR4.

The benchmark supports two action interfaces. \emph{End-to-end policies} predict joint or end-effector actions. These predictions are converted to joint targets, and a configured number of actions is executed from each chunk before the next prediction. \emph{Agents} are multimodal language models that issue structured tool calls. The \texttt{move\_eef} tool accepts an end-effector pose target and gripper command. The backend handles inverse kinematics and motion planning. The \texttt{wait\_steps} tool advances simulation while maintaining the current arm and gripper targets. Agents can use \texttt{memory\_resolve} to retrieve a previously observed image by its observation identifier and add it to the model's context. Retrieval does not advance simulation. An agent can also call \texttt{give\_up} to end the episode itself. Every evaluated agent uses this one fixed harness of tools, planner and history access. The same two interfaces cover every task interaction, including count-only feedback (F2), an unnumbered needle gauge (CP06), unnumbered buttons (M2), keypad input (E2), bimanual uncapping (CP09), and tray carrying with loose balls (DR4). No task requires a task-specific interface.

\subsection{Episode Protocol and Timing}
\label{subsec:episode-lifecycle}
\label{subsec:timing}
Each episode starts from a fixed instance for its task, difficulty level and scene setting, and systems evaluated under the same protocol use the same instance list. Reset restores the robot, objects, task stage, evaluator, and random-number state. Policy memory is cleared between episodes and retained across stages within an episode. Physics simulation advances at \textbf{120~Hz}, while arm and gripper targets are updated at \textbf{30~Hz}. Simulation pauses during model inference. Task deadlines use \emph{simulation time}, while inference latency and total runtime use \emph{wall-clock time}, the elapsed real time. A delayed model response therefore consumes wall-clock time without reducing the simulated response window, so timed tasks measure responses in simulation time rather than real-time performance. For end-to-end policies, each task and difficulty level has an episode horizon set to 1.2 times the 90th percentile of demonstration durations, rounded up to a control step. These simulation-time limits range from 16.27--420.50\,s for Mind, 6.27--156.47\,s for Act, and 34.40--308.50\,s for Mind2Act. Agents receive four times this physics budget and at most 150 decisions per episode, since a tool call spans many control steps. Wall-clock limits are safety stops, not task limits. An episode stopped by one is scored like any other unfinished episode. Episodes end on completion, policy termination, or budget exhaustion. Simulator crashes and incomplete records are not counted as task failures, and infrastructure aborts are rerun. Appendices~\ref{app:platform} and~\ref{app:impl} specify episode horizons, model-call budgets, and scheduling.

\subsection{Metrics and Aggregation}
\label{subsec:protocol-metrics}
\label{subsec:validity-aggregation}
A success rate alone says how often a task is finished but not how far a failed episode got, while a progress score alone can reward work that never completes the task. We therefore report the \textbf{final success rate} (SR), where success requires the complete task predicate, together with a 0--100 \textbf{task score} that credits task-specific progress and equals 100 exactly when the episode succeeds. Partial credit never overrides an unmet terminal requirement. For a task, level and system with $N$ eligible attempts and outcomes $s_i\in\{0,1\}$, the success rate is
\begin{equation}
\hat S=\frac{1}{N}\sum_{i=1}^{N}s_i,
\end{equation}
with protocol timeouts counted as failures and infrastructure aborts rerun. View summaries average the ten tasks of a view, each pooled over its three levels, and the primary result averages the \emph{Clean} and \emph{Random} settings. A robot is useful when it completes whole workflows, and in such workflows remembered information and physical execution depend on each other, so strength in either capability alone is not enough. For this reason, the \emph{Overall} score weights the views as $0.25\,$Mind$\,+\,0.25\,$Act$\,+\,0.50\,$Mind2Act, so that Mind2Act counts as much as the two single views together. The Overall SR is the plain fraction of successful episodes. A Mind2Act task is scored as a whole. Which side failed is assigned by the episode-level diagnosis (Section~\ref{sec:diagnostics}). Partial-credit rules are task-specific, so scores are comparable across systems on one task. For success rates we report 95\% Wilson score intervals over episodes, which capture sampling over the fixed instance lists rather than variation between tasks or repeated runs. As a descriptive reference for composition, we also compare a system's Mind2Act success rate with the product of its Mind and Act success rates, the value expected if a Mind2Act workflow required one independent success of each kind. The task score is defined in Appendix~\ref{app:task-score}, and per-task success predicates are listed in Appendix~\ref{app:metrics}.

\subsection{Evaluation Integrity}
\label{subsec:benchmark-integrity}
\label{subsec:anti-gaming}
Because an agent can declare that it has finished and a policy can repeat an action until something happens, the evaluator has to judge outcomes. Completion is therefore assessed using \emph{environment-side predicates}. Commands and success declarations receive credit only when the required physical conditions are met. Repeating an action earns credit only when it satisfies a new requirement. Task rules distinguish irreversible failures, unscored actions, and valid exploration. For example, in DR4, a ball leaving its circle causes failure even if it later rolls back. In F2, submitting an incomplete layout does not consume an attempt. In M2, revealing a mismatched pair is valid exploration. Any action sequence that satisfies the task rules is valid. A run is classified as a valid failure only when its record is sufficient to evaluate the final success conditions. Unavailable observations and infrastructure interruptions are recorded separately.

\section{Benchmark Results}
\label{sec:results}

Building on the protocol in Section~\ref{sec:protocol}, we evaluate nine systems that span the two main routes to embodied manipulation: multimodal agents that act through tools and end-to-end policies trained on demonstrations. Section~\ref{subsec:experimental-setup} describes the systems and the evaluation procedure. Section~\ref{subsec:main-results} analyzes how well they reason, act and compose the two, where and why they fall short across views, conditions and application domains, and what the evaluation costs in time and money. Section~\ref{subsec:suited} summarizes which system suits which task.

\subsection{Experimental Setup}
\label{subsec:experimental-setup}

\subsubsection{Systems}
We evaluate two system families, each ranked on its own leaderboard. \emph{Agents} are multimodal language models that select motion tools from camera images and robot state, without any training on Mind2Act World. We evaluate GPT-6 Astra~\cite{openai2026gpt6astra}, DeepSeek-V4.1-Flash~\cite{deepseek2026v41flash} and Gemini-4-Argon~\cite{google2026gemini4argon}. \emph{End-to-end policies} map observations and the instruction to robot actions after fine-tuning on the shared demonstrations. We evaluate the VLAs KinRT~\cite{kinrt2026}, $\pi_{0.5}$~\cite{pi2025pi05}, G0.5~\cite{galaxea2026g05} and X-VLA~\cite{zheng2025xvla}, and the WAMs InternW0~\cite{cai2026internw0} and FastWAM~\cite{yuan2026fastwam}. Appendix~\ref{app:baselines} lists the evaluated systems.

\subsubsection{Training and Adaptation}
We fine-tune each end-to-end policy from its public checkpoint on the shared collection of 4,500 demonstrations, fifty per task and difficulty level (Section~\ref{subsec:data-statistics}), with the initialization, global batch size, and training steps of each system detailed in Appendix~\ref{app:baselines}. In contrast, we use the agents directly as provided through their APIs, without fine-tuning.

\subsubsection{Inference and Evaluation Instances}
All evaluations run at 120\,Hz physics and 30\,Hz control/RGB observations, with simulation paused during inference so that task deadlines are measured in simulation time and inference cost in wall-clock time. Across the 180 configurations (30 tasks $\times$ 3 levels $\times$ 2 settings), the nine evaluated systems account for 56,700 scored episodes.

At the start of an episode, each system receives the English task instruction, selected camera views, and declared robot state. End-to-end policies generate action chunks that are converted to robot targets at 30\,Hz, whereas agents issue pose-level motion-tool requests and may retrieve earlier observations from the same episode. For each task and difficulty level, we fix a shared list of 50 instances and evaluate them in both \emph{Clean} and \emph{Random} settings (Table~\ref{tab:evaluation-allocation}). End-to-end policies run on all 100 resulting episodes per task and level (9,000 episodes per policy), while agents run on a shared subset of 5 \emph{Clean} and 5 \emph{Random} episodes (900 episodes per agent).

\emph{Random} instances preserve the underlying task instance while adding table clutter, one of 23 table materials, and one of six light types (with RGB channels scaled in $[0, 2]$ and intensity in $[0.8, 1.2]$). The range is deliberately wide, so \emph{Random} serves as an appearance stress test. End-to-end policies are allotted a simulation-time horizon of $1.2\times$ the 90th percentile of demonstration durations, shared across \emph{Clean} and \emph{Random}, with a 2-hour wall-clock safety limit. Agents receive $4\times$ this physics budget, up to 150 decisions per episode, and a 6-hour wall-clock safety limit. All three agents are evaluated at medium reasoning effort. Appendix~\ref{app:impl} details the remaining run settings.

\begin{table}[!htb]
\centering\small
\caption{\textbf{Evaluation episodes per system.} Every system runs the same 180 configurations: thirty tasks at three difficulty levels, each in a \emph{Clean} and a \emph{Random} scene setting. An end-to-end policy runs ten times as many episodes per configuration as an agent. Counts are per system. The primary result combines the \emph{Clean} and \emph{Random} configurations of each task and difficulty with equal weight.}
\label{tab:evaluation-allocation}
\begingroup
\setlength{\tabcolsep}{6pt}
\renewcommand{\arraystretch}{1.12}
\begin{tabularx}{\linewidth}{@{}Xrr@{}}
\toprule
\textbf{Allocation} & \textbf{Agent} & \textbf{End-to-end policy}\\
\midrule
Tasks / difficulty levels / scene settings & 30 / 3 / 2 & 30 / 3 / 2\\
\textbf{Configurations} & \textbf{180} & \textbf{180}\\
Episodes per configuration & 5 & 50\\
Clean + Random episodes per task and difficulty & 10 & 100\\
Episodes per task, across three difficulties & 30 & 300\\
Episodes per view, ten tasks & 300 & 3,000\\
Clean episodes across the full suite (90 configurations) & 450 & 4,500\\
Random episodes across the full suite (90 configurations) & 450 & 4,500\\
\textbf{Total episodes across the full suite} & \textbf{900} & \textbf{9,000}\\
\bottomrule
\end{tabularx}
\endgroup
\end{table}

\subsubsection{Metrics and Statistical Reporting}
Following Section~\ref{subsec:protocol-metrics}, we report the final success rate (SR) and partial-progress task score (0--100), aggregated across difficulty levels and averaged equally over \emph{Clean} and \emph{Random} settings. Alongside point estimates, we provide 95\% Wilson score intervals in Appendix~\ref{app:full-results}. 

\subsection{Main Results and Analysis}
\label{subsec:main-results}
\label{subsec:task-results}\label{subsec:dimension-results}\label{subsec:setting-difficulty}
\label{subsec:mind-results}\label{subsec:capability-profiles}\label{subsec:act-results}\label{subsec:act-analysis}
\label{subsec:mind2act-results}\label{subsec:score-success}\label{subsec:stage-completion}
\newcommand{\axisref}[3]{\mbox{\color{#1}\footnotesize\faIcon{#2}\hspace{0.3em}\sffamily\bfseries #3}}%
A failure can come from forgetting task state, failing to execute, or failing to connect the two. We first ask how well current systems reason, act and compose the two, taken as a whole (\axisref{M2AInk}{chart-bar}{OVERALL}). We then look at the two capabilities apart before putting them together. \axisref{M2AMind}{brain}{MIND} asks whether a system can remember what it has seen. \axisref{M2AAct}{hand-paper}{ACT} asks whether it can act precisely and on time. \axisref{M2ACoord}{sync-alt}{MIND2ACT} asks whether the two still work together inside one workflow. \axisref{M2ARand}{random}{APPEARANCE} changes how the scene looks, and \axisref{M2ALevel}{signal}{DIFFICULTY} makes the task harder. The last two look beyond the simulator. \axisref{M2AInk}{stopwatch}{LATENCY \& COST} asks whether a system would keep up in real time, and what the evaluation costs. \axisref{M2AReal}{globe-asia}{REAL WORLD} reads the results by application domain and asks how far each is from deployment. Table~\ref{tab:per-task-sr} lists the success rate of every system on each task, grouped by application domain, and Appendix~\ref{app:per-view-detail} gives the full episode counts.

\begin{table}[!tp]
\centering
\caption{\textbf{Mind2Act World leaderboards.} Each cell reports task score (0--100) / success rate (\%) for one view. For each model, the first row pools the three difficulty levels, and \emph{All} is the mean of the \emph{Clean} and \emph{Random} settings. The Overall score is $0.25\,$\VMind$\,+\,0.25\,$\VAct$\,+\,0.50\,$\VMtA, which gives \VMtA{} as much weight as the two single views together (Section~\ref{subsec:protocol-metrics}). The Overall success rate pools all episodes. Agents and end-to-end policies are ranked separately; within each leaderboard, the best and second-best \emph{All} values of each column (by task score) are marked as \LgBest{Best} and \LgSecond{Second Best}.}
\label{tab:main-results}
\begingroup
\fontsize{7.0}{8.8}\selectfont
\setlength{\tabcolsep}{1.6pt}
\renewcommand{\arraystretch}{0.95}
\setlength{\aboverulesep}{0pt}\setlength{\belowrulesep}{0pt}%
\definecolor{ResultMind}{HTML}{416D62}
\definecolor{ResultAct}{HTML}{36566F}
\definecolor{ResultCoord}{HTML}{8A6A2F}
\renewcommand{\mtgroup}[2]{\rowcolor{M2AGroup}\multicolumn{#1}{l}{\strut\textbf{\textit{#2}}}}%
\newcommand{\viewheading}[3]{\raisebox{-1.2mm}{\includegraphics[height=5mm,width=6mm,keepaspectratio]{figures/view_icons/#1}}\hspace{1mm}\textcolor{#2}{\textbf{#3}}}
\newcommand{\subheads}[2]{\mthead{#1}{#2}{\textbf{All}} & \mthead{#1}{#2}{Clean} & \mthead{#1}{#2}{Rand.}}
\begin{tabularx}{\linewidth}{l@{\hspace{6pt}}l@{\hspace{6pt}}l!{\vrule}*{3}{>{\centering\arraybackslash}X}!{\vrule}*{3}{>{\centering\arraybackslash}X}!{\vrule}*{3}{>{\centering\arraybackslash}X}!{\vrule}>{\centering\arraybackslash}p{13mm}}
\toprule
\rule[-2.2mm]{0pt}{7.4mm}\textbf{Model} & \textbf{Type} & & \multicolumn{3}{c!{\vrule}}{\cellcolor{M2AMindT}\viewheading{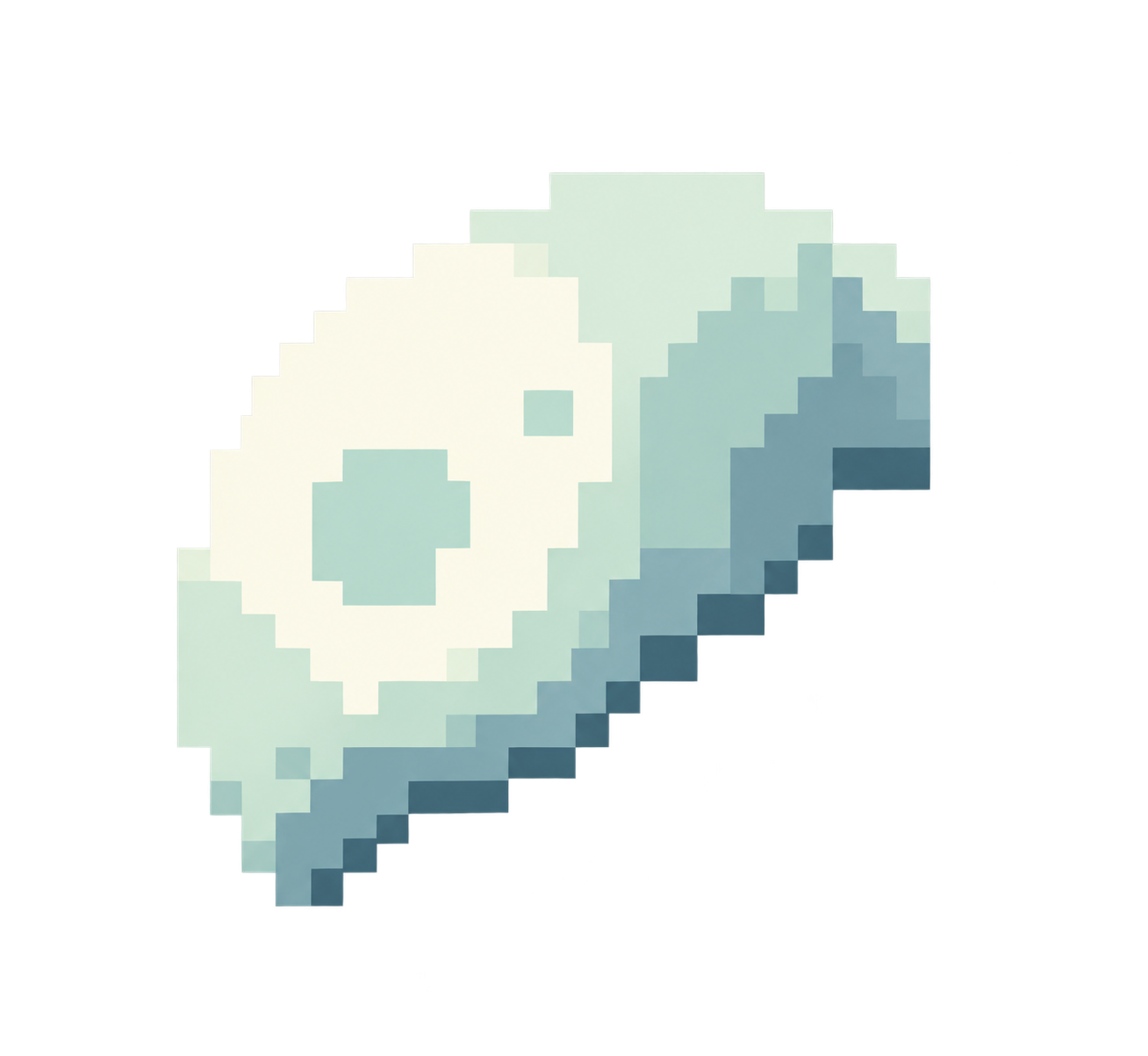}{ResultMind}{Mind}} & \multicolumn{3}{c!{\vrule}}{\cellcolor{M2AActT}\viewheading{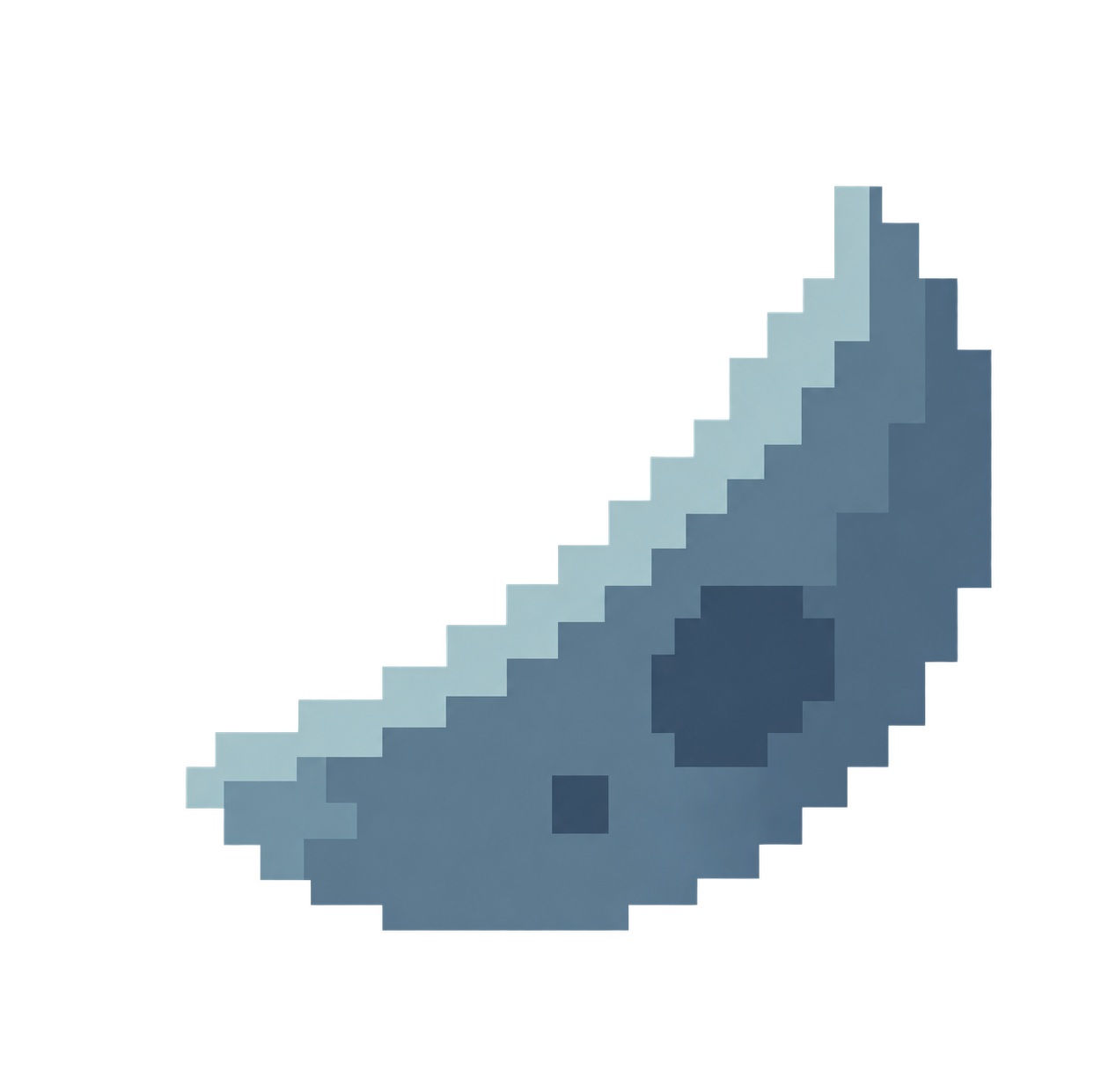}{ResultAct}{Act}} & \multicolumn{3}{c!{\vrule}}{\cellcolor{M2ACoordT}\viewheading{moon.png}{ResultCoord}{Mind2Act}} & \multirow{2}{*}{\textbf{\textit{Overall}}} \\
\cmidrule{4-6}\cmidrule{7-9}\cmidrule{10-12}
\rule[-1.4mm]{0pt}{5.0mm} & & & \subheads{ResultMind}{M2AMindT!45} & \subheads{ResultAct}{M2AActT!45} & \subheads{ResultCoord}{M2ACoordT!45} & \\
\midrule
\mtgroup{13}{Agents under one fixed harness} \\
\multirow{4}{*}{\shortstack[l]{\textbf{GPT-6 Astra}~\cite{openai2026gpt6astra}\\[0.2mm]\textcolor{black!55}{\fontsize{6.2}{7}\selectfont OpenAI, 2026}}} & \multirow{4}{*}{Agent} & \textbf{All} & \rkA{\textbf{53.9/52.7}} & 55.9/54.7 & 51.9/50.7 & \rkA{\textbf{40.4/39.0}} & 44.6/43.3 & 36.3/34.7 & \rkA{\textbf{13.3/12.0}} & 13.3/12.0 & 13.3/12.0 & \rkA{\textbf{30.2/34.6}} \\
 & & Easy & 64.7/64.0 & 65.1/64.0 & 64.3/64.0 & 44.9/44.0 & 47.7/46.0 & 42.0/42.0 & 27.3/26.0 & 27.0/26.0 & 27.5/26.0 & 41.0/44.7 \\
 & & Med. & 52.1/50.0 & 63.2/62.0 & 40.9/38.0 & 41.0/39.0 & 45.2/44.0 & 36.8/34.0 & 8.5/7.0 & 7.8/6.0 & 9.3/8.0 & 27.5/32.0 \\
 & & Hard & 44.9/44.0 & 39.4/38.0 & 50.4/50.0 & 35.5/34.0 & 40.8/40.0 & 30.1/28.0 & 4.2/3.0 & 5.2/4.0 & 3.2/2.0 & 22.2/27.0 \\[3pt]
\multirow{4}{*}{\shortstack[l]{\textbf{DeepSeek-V4.1-Flash}~\cite{deepseek2026v41flash}\\[0.2mm]\textcolor{black!55}{\fontsize{6.2}{7}\selectfont DeepSeek-AI, 2026}}} & \multirow{4}{*}{Agent} & \textbf{All} & \rkB{12.4/11.0} & 11.0/9.3 & 13.8/12.7 & 11.3/8.7 & 11.8/9.3 & 10.9/8.0 & \rkB{0.3/0.0} & 0.2/0.0 & 0.4/0.0 & \rkB{6.1/6.6} \\
 & & Easy & 25.5/24.0 & 20.1/18.0 & 30.9/30.0 & 15.1/11.0 & 19.3/16.0 & 10.9/6.0 & 0.6/0.0 & 0.3/0.0 & 0.9/0.0 & 10.4/11.7 \\
 & & Med. & 7.7/6.0 & 7.8/6.0 & 7.6/6.0 & 10.4/8.0 & 6.7/4.0 & 14.2/12.0 & 0.3/0.0 & 0.3/0.0 & 0.3/0.0 & 4.7/4.7 \\
 & & Hard & 3.9/3.0 & 5.1/4.0 & 2.8/2.0 & 8.5/7.0 & 9.4/8.0 & 7.6/6.0 & 0.1/0.0 & 0.1/0.0 & 0.1/0.0 & 3.2/3.3 \\[3pt]
\multirow{4}{*}{\shortstack[l]{\textbf{Gemini-4-Argon}~\cite{google2026gemini4argon}\\[0.2mm]\textcolor{black!55}{\fontsize{6.2}{7}\selectfont Google DeepMind, 2026}}} & \multirow{4}{*}{Agent} & \textbf{All} & 5.0/4.3 & 5.7/5.3 & 4.2/3.3 & \rkB{12.3/9.7} & 13.2/10.7 & 11.4/8.7 & 0.1/0.0 & 0.1/0.0 & 0.1/0.0 & 4.4/4.7 \\
 & & Easy & 12.1/11.0 & 12.9/12.0 & 11.2/10.0 & 12.5/8.0 & 15.2/10.0 & 9.7/6.0 & 0.3/0.0 & 0.3/0.0 & 0.2/0.0 & 6.3/6.3 \\
 & & Med. & 2.6/2.0 & 4.1/4.0 & 1.1/0.0 & 11.1/9.0 & 11.1/10.0 & 11.2/8.0 & 0.0/0.0 & 0.0/0.0 & 0.1/0.0 & 3.5/3.7 \\
 & & Hard & 0.2/0.0 & 0.1/0.0 & 0.3/0.0 & 13.1/12.0 & 13.1/12.0 & 13.1/12.0 & 0.0/0.0 & 0.0/0.0 & 0.0/0.0 & 3.3/4.0 \\[3pt]
\mtgroup{13}{End-to-end policies: VLAs and WAMs} \\
\multirow{4}{*}{\shortstack[l]{\textbf{KinRT}~\cite{kinrt2026}\\[0.2mm]\textcolor{black!55}{\fontsize{6.2}{7}\selectfont IIGroup, Tsinghua, 2026}}} & \multirow{4}{*}{VLA} & \textbf{All} & \rkB{5.6/4.4} & 7.9/6.5 & 3.4/2.3 & 17.4/13.6 & 17.8/14.0 & 17.0/13.2 & \rkB{4.3/2.9} & 5.3/3.8 & 3.3/2.1 & \rkB{7.9/7.0} \\
 & & Easy & 11.3/9.8 & 15.3/13.6 & 7.3/6.0 & 24.8/19.0 & 25.1/19.8 & 24.4/18.2 & 8.6/6.7 & 10.1/8.2 & 7.2/5.2 & 13.3/11.8 \\
 & & Med. & 4.1/2.7 & 6.4/4.8 & 1.9/0.6 & 14.9/11.1 & 16.2/12.0 & 13.5/10.2 & 2.9/1.6 & 3.9/2.4 & 1.9/0.8 & 6.2/5.1 \\
 & & Hard & 1.5/0.7 & 1.9/1.0 & 1.2/0.4 & 12.6/10.7 & 12.1/10.2 & 13.0/11.2 & 1.4/0.5 & 1.8/0.8 & 1.0/0.2 & 4.2/4.0 \\[3pt]
\multirow{4}{*}{\shortstack[l]{\textbf{\boldmath$\pi_{0.5}$}~\cite{pi2025pi05}\\[0.2mm]\textcolor{black!55}{\fontsize{6.2}{7}\selectfont Physical Intelligence, 2025}}} & \multirow{4}{*}{VLA} & \textbf{All} & 4.5/3.5 & 6.6/5.5 & 2.3/1.5 & 15.9/12.0 & 16.7/12.7 & 15.0/11.3 & \rkA{\textbf{5.1/3.6}} & 6.2/4.5 & 4.0/2.7 & 7.6/6.4 \\
 & & Easy & 9.2/7.8 & 13.6/12.0 & 4.8/3.6 & 22.9/16.6 & 24.0/17.6 & 21.7/15.6 & 10.6/8.6 & 12.9/10.8 & 8.3/6.4 & 13.3/11.0 \\
 & & Med. & 3.2/2.2 & 4.8/3.6 & 1.5/0.8 & 13.7/10.1 & 15.2/11.2 & 12.3/9.0 & 3.6/2.2 & 4.3/2.6 & 2.9/1.8 & 6.0/4.8 \\
 & & Hard & 1.0/0.5 & 1.3/0.8 & 0.6/0.2 & 11.1/9.3 & 11.1/9.2 & 11.1/9.4 & 1.1/0.1 & 1.5/0.2 & 0.7/0.0 & 3.6/3.3 \\[3pt]
\multirow{4}{*}{\shortstack[l]{\textbf{G0.5}~\cite{galaxea2026g05}\\[0.2mm]\textcolor{black!55}{\fontsize{6.2}{7}\selectfont Galaxea Team, 2026}}} & \multirow{4}{*}{VLA} & \textbf{All} & 2.8/1.9 & 3.7/2.7 & 1.9/1.1 & 16.0/13.4 & 16.7/14.1 & 15.3/12.7 & 3.1/2.3 & 3.2/2.3 & 2.9/2.3 & 6.2/5.9 \\
 & & Easy & 6.9/5.6 & 9.1/7.8 & 4.6/3.4 & 25.4/21.9 & 26.0/22.6 & 24.8/21.2 & 7.3/6.2 & 6.8/5.6 & 7.8/6.8 & 11.7/11.2 \\
 & & Med. & 1.0/0.2 & 1.4/0.4 & 0.7/0.0 & 12.5/9.8 & 13.4/10.6 & 11.6/9.0 & 1.2/0.5 & 1.8/1.0 & 0.6/0.0 & 4.0/3.5 \\
 & & Hard & 0.5/0.0 & 0.6/0.0 & 0.5/0.0 & 10.0/8.6 & 10.7/9.2 & 9.4/8.0 & 0.7/0.2 & 0.9/0.4 & 0.4/0.0 & 3.0/2.9 \\[3pt]
\multirow{4}{*}{\shortstack[l]{\textbf{X-VLA}~\cite{zheng2025xvla}\\[0.2mm]\textcolor{black!55}{\fontsize{6.2}{7}\selectfont Zheng et al., 2025}}} & \multirow{4}{*}{VLA} & \textbf{All} & 1.4/1.0 & 2.1/1.6 & 0.7/0.5 & \rkA{\textbf{21.9/19.6}} & 23.2/21.1 & 20.6/18.2 & 0.7/0.2 & 0.9/0.3 & 0.5/0.1 & 6.2/7.0 \\
 & & Easy & 3.3/2.8 & 5.0/4.4 & 1.5/1.2 & 28.2/24.7 & 29.8/26.6 & 26.6/22.8 & 1.5/0.6 & 1.9/0.8 & 1.2/0.4 & 8.6/9.4 \\
 & & Med. & 0.6/0.2 & 0.8/0.2 & 0.5/0.2 & 14.6/12.6 & 16.2/14.2 & 13.1/11.0 & 0.3/0.0 & 0.4/0.0 & 0.2/0.0 & 4.0/4.3 \\
 & & Hard & 0.4/0.1 & 0.5/0.2 & 0.2/0.0 & 22.9/21.6 & 23.6/22.4 & 22.1/20.8 & 0.3/0.0 & 0.4/0.0 & 0.2/0.0 & 5.9/7.2 \\[3pt]
\multirow{4}{*}{\shortstack[l]{\textbf{InternW0}~\cite{cai2026internw0}\\[0.2mm]\textcolor{black!55}{\fontsize{6.2}{7}\selectfont Cai et al., 2026}}} & \multirow{4}{*}{WAM} & \textbf{All} & \rkA{\textbf{6.1/5.4}} & 6.0/5.2 & 6.2/5.5 & \rkB{20.3/17.6} & 20.9/18.2 & 19.8/17.0 & 4.1/3.0 & 4.9/3.7 & 3.4/2.3 & \rkA{\textbf{8.7/8.7}} \\
 & & Easy & 14.4/13.5 & 13.5/12.4 & 15.4/14.6 & 24.6/20.6 & 25.0/21.0 & 24.3/20.2 & 9.4/8.0 & 10.9/9.6 & 7.9/6.4 & 14.5/14.0 \\
 & & Med. & 3.0/2.1 & 3.4/2.4 & 2.6/1.8 & 22.1/19.5 & 21.8/19.2 & 22.3/19.8 & 1.8/0.8 & 2.4/1.2 & 1.3/0.4 & 7.2/7.5 \\
 & & Hard & 1.0/0.5 & 1.2/0.8 & 0.7/0.2 & 14.3/12.7 & 15.9/14.4 & 12.7/11.0 & 1.2/0.3 & 1.3/0.4 & 1.0/0.2 & 4.4/4.5 \\[3pt]
\multirow{4}{*}{\shortstack[l]{\textbf{FastWAM}~\cite{yuan2026fastwam}\\[0.2mm]\textcolor{black!55}{\fontsize{6.2}{7}\selectfont Yuan et al., 2026}}} & \multirow{4}{*}{WAM} & \textbf{All} & 1.2/0.9 & 2.0/1.5 & 0.4/0.3 & 9.3/7.4 & 13.3/10.7 & 5.4/4.1 & 0.9/0.5 & 1.3/0.9 & 0.4/0.1 & 3.1/2.9 \\
 & & Easy & 3.0/2.5 & 4.9/4.2 & 1.0/0.8 & 11.2/7.9 & 14.4/10.2 & 7.9/5.6 & 2.1/1.6 & 3.4/2.8 & 0.8/0.4 & 4.6/4.0 \\
 & & Med. & 0.4/0.2 & 0.7/0.4 & 0.1/0.0 & 10.4/8.7 & 13.3/11.0 & 7.6/6.4 & 0.2/0.0 & 0.2/0.0 & 0.2/0.0 & 2.8/3.0 \\
 & & Hard & 0.2/0.0 & 0.3/0.0 & 0.2/0.0 & 6.4/5.6 & 12.2/11.0 & 0.7/0.2 & 0.2/0.0 & 0.3/0.0 & 0.2/0.0 & 1.8/1.9 \\
\bottomrule
\end{tabularx}
\endgroup
\end{table}

\begin{table}[!tp]
\centering
\caption{\textbf{Final success rate (\%) on every task, grouped by application domain.} Each block lists the tasks of one domain of Table~\ref{tab:application-domains} and closes with the domain mean, the mean of its task rates. In each block, the first three columns are the agents and the other six the end-to-end policies; task IDs are colored by view (\VMind, \VAct, \VMtA) and named in Table~\ref{tab:task-summary}. Each cell pools the \emph{Clean} and \emph{Random} settings and the three difficulty levels: 30 episodes per task for an agent and 300 for an end-to-end policy. Shading, in the view's color, deepens with the success rate, and a gray \LgZero{} marks a task that the system never completes. M1, DR4, PR5, PR6, CP03, CP04, CP07 and CP09 have no success in any of the 56,700 episodes. Appendix Table~\ref{tab:per-task-appendix} gives task scores and the two settings separately.}
\label{tab:per-task-sr}\label{tab:domain-success}
\begingroup
\fontsize{7.4}{8.8}\selectfont
\setlength{\tabcolsep}{0.8pt}
\renewcommand{\arraystretch}{1.08}
\newcommand{\rot}[1]{\rotatebox{90}{#1}}
\begin{tabularx}{\linewidth}{@{}>{\raggedright\arraybackslash}p{7.4mm}*{3}{>{\centering\arraybackslash}X}!{\color{black!30}\vrule}*{6}{>{\centering\arraybackslash}X}@{\hspace{3.2mm}}>{\raggedright\arraybackslash}p{7.4mm}*{3}{>{\centering\arraybackslash}X}!{\color{black!30}\vrule}*{6}{>{\centering\arraybackslash}X}@{}}
\toprule
\textbf{ID} & \rot{GPT-6} & \rot{DeepSeek} & \rot{Gemini} & \rot{KinRT} & \rot{$\pi_{0.5}$} & \rot{G0.5} & \rot{X-VLA} & \rot{InternW0} & \rot{FastWAM} & \textbf{ID} & \rot{GPT-6} & \rot{DeepSeek} & \rot{Gemini} & \rot{KinRT} & \rot{$\pi_{0.5}$} & \rot{G0.5} & \rot{X-VLA} & \rot{InternW0} & \rot{FastWAM}\\
\midrule
\multicolumn{10}{@{}l}{\cellcolor{M2AGroup}\strut\textcolor{M2AInk}{\faIcon{utensils}}\hspace{0.45em}\textbf{\textit{Home \& Food Service}}} & \multicolumn{10}{@{}l}{\cellcolor{M2AGroup}\strut\textcolor{M2AInk}{\faIcon{industry}}\hspace{0.45em}\textbf{\textit{Manufacturing \& Logistics}}}\\
\textcolor{M2AMind}{\textbf{P1}} & \cellcolor{M2AMind!35}43 & \textcolor{black!35}{0} & \textcolor{black!35}{0} & \cellcolor{M2AMind!8}0.7 & \cellcolor{M2AMind!9}1.0 & \textcolor{black!35}{0} & \textcolor{black!35}{0} & \cellcolor{M2AMind!8}0.7 & \cellcolor{M2AMind!8}0.3 & \textcolor{M2AMind}{\textbf{E1}} & \cellcolor{M2AMind!53}73 & \cellcolor{M2AMind!20}20 & \cellcolor{M2AMind!12}6.7 & \cellcolor{M2AMind!12}6.7 & \cellcolor{M2AMind!9}1.0 & \textcolor{black!35}{0} & \textcolor{black!35}{0} & \textcolor{black!35}{0} & \cellcolor{M2AMind!9}1.0\\
\textcolor{M2AMind}{\textbf{P2}} & \textcolor{black!35}{0} & \textcolor{black!35}{0} & \textcolor{black!35}{0} & \cellcolor{M2AMind!17}15 & \cellcolor{M2AMind!17}14 & \cellcolor{M2AMind!9}2.0 & \textcolor{black!35}{0} & \cellcolor{M2AMind!9}1.3 & \textcolor{black!35}{0} & \textcolor{M2AMind}{\textbf{M1}} & \textcolor{black!35}{0} & \textcolor{black!35}{0} & \textcolor{black!35}{0} & \textcolor{black!35}{0} & \textcolor{black!35}{0} & \textcolor{black!35}{0} & \textcolor{black!35}{0} & \textcolor{black!35}{0} & \textcolor{black!35}{0}\\
\textcolor{M2AMind}{\textbf{S1}} & \cellcolor{M2AMind!22}23 & \textcolor{black!35}{0} & \textcolor{black!35}{0} & \textcolor{black!35}{0} & \textcolor{black!35}{0} & \textcolor{black!35}{0} & \cellcolor{M2AMind!8}0.3 & \cellcolor{M2AMind!8}0.7 & \textcolor{black!35}{0} & \textcolor{M2AAct}{\textbf{DR1}} & \cellcolor{M2AAct!56}77 & \cellcolor{M2AAct!12}6.7 & \textcolor{black!35}{0} & \cellcolor{M2AAct!20}19 & \cellcolor{M2AAct!16}13 & \cellcolor{M2AAct!24}25 & \textcolor{black!35}{0} & \cellcolor{M2AAct!22}23 & \cellcolor{M2AAct!9}2.0\\
\textcolor{M2AAct}{\textbf{PR4}} & \textcolor{black!35}{0} & \textcolor{black!35}{0} & \textcolor{black!35}{0} & \cellcolor{M2AAct!19}18 & \cellcolor{M2AAct!14}10 & \cellcolor{M2AAct!19}17 & \cellcolor{M2AAct!48}64 & \cellcolor{M2AAct!12}6.3 & \textcolor{black!35}{0} & \textcolor{M2AAct}{\textbf{DR3}} & \cellcolor{M2AAct!58}80 & \cellcolor{M2AAct!25}27 & \cellcolor{M2AAct!33}40 & \cellcolor{M2AAct!57}79 & \cellcolor{M2AAct!57}79 & \cellcolor{M2AAct!40}51 & \cellcolor{M2AAct!55}76 & \cellcolor{M2AAct!53}73 & \cellcolor{M2AAct!46}61\\
\textcolor{M2ACoord}{\textbf{CP01}} & \cellcolor{M2ACoord!45}60 & \textcolor{black!35}{0} & \textcolor{black!35}{0} & \cellcolor{M2ACoord!13}8.0 & \cellcolor{M2ACoord!13}8.7 & \cellcolor{M2ACoord!13}8.0 & \cellcolor{M2ACoord!9}1.3 & \cellcolor{M2ACoord!9}1.7 & \cellcolor{M2ACoord!11}4.3 & \textcolor{M2AAct}{\textbf{DR4}} & \textcolor{black!35}{0} & \textcolor{black!35}{0} & \textcolor{black!35}{0} & \textcolor{black!35}{0} & \textcolor{black!35}{0} & \textcolor{black!35}{0} & \textcolor{black!35}{0} & \textcolor{black!35}{0} & \textcolor{black!35}{0}\\
\textcolor{M2ACoord}{\textbf{CP02}} & \cellcolor{M2ACoord!19}17 & \textcolor{black!35}{0} & \textcolor{black!35}{0} & \cellcolor{M2ACoord!8}0.3 & \cellcolor{M2ACoord!8}0.7 & \textcolor{black!35}{0} & \textcolor{black!35}{0} & \cellcolor{M2ACoord!8}0.3 & \textcolor{black!35}{0} & \textit{Mean} & \cellcolor{M2AReal!37}\textbf{46.0} & \cellcolor{M2AReal!15}\textbf{10.7} & \cellcolor{M2AReal!14}\textbf{9.3} & \cellcolor{M2AReal!21}\textbf{20.9} & \cellcolor{M2AReal!20}\textbf{18.6} & \cellcolor{M2AReal!17}\textbf{15.2} & \cellcolor{M2AReal!17}\textbf{15.2} & \cellcolor{M2AReal!20}\textbf{19.2} & \cellcolor{M2AReal!16}\textbf{12.8}\\
\textcolor{M2ACoord}{\textbf{CP03}} & \textcolor{black!35}{0} & \textcolor{black!35}{0} & \textcolor{black!35}{0} & \textcolor{black!35}{0} & \textcolor{black!35}{0} & \textcolor{black!35}{0} & \textcolor{black!35}{0} & \textcolor{black!35}{0} & \textcolor{black!35}{0} &  &  &  &  &  &  &  &  &  & \\
\textcolor{M2ACoord}{\textbf{CP04}} & \textcolor{black!35}{0} & \textcolor{black!35}{0} & \textcolor{black!35}{0} & \textcolor{black!35}{0} & \textcolor{black!35}{0} & \textcolor{black!35}{0} & \textcolor{black!35}{0} & \textcolor{black!35}{0} & \textcolor{black!35}{0} & \multicolumn{10}{@{}l}{\cellcolor{M2AGroup}\strut\textcolor{M2AInk}{\faIcon{cubes}}\hspace{0.45em}\textbf{\textit{Assembly \& Physical Reasoning}}}\\
\textit{Mean} & \cellcolor{M2AReal!19}\textbf{17.9} & \textcolor{black!35}{\textbf{0}} & \textcolor{black!35}{\textbf{0}} & \cellcolor{M2AReal!11}\textbf{5.2} & \cellcolor{M2AReal!11}\textbf{4.3} & \cellcolor{M2AReal!10}\textbf{3.4} & \cellcolor{M2AReal!13}\textbf{8.2} & \cellcolor{M2AReal!9}\textbf{1.4} & \cellcolor{M2AReal!8}\textbf{0.6} & \textcolor{M2AAct}{\textbf{PR6}} & \textcolor{black!35}{0} & \textcolor{black!35}{0} & \textcolor{black!35}{0} & \textcolor{black!35}{0} & \textcolor{black!35}{0} & \textcolor{black!35}{0} & \textcolor{black!35}{0} & \textcolor{black!35}{0} & \textcolor{black!35}{0}\\
 &  &  &  &  &  &  &  &  &  & \textcolor{M2ACoord}{\textbf{CP07}} & \textcolor{black!35}{0} & \textcolor{black!35}{0} & \textcolor{black!35}{0} & \textcolor{black!35}{0} & \textcolor{black!35}{0} & \textcolor{black!35}{0} & \textcolor{black!35}{0} & \textcolor{black!35}{0} & \textcolor{black!35}{0}\\
\multicolumn{10}{@{}l}{\cellcolor{M2AGroup}\strut\textcolor{M2AInk}{\faIcon{tools}}\hspace{0.45em}\textbf{\textit{Instruments \& Tool Use}}} & \textcolor{M2ACoord}{\textbf{CP08}} & \cellcolor{M2ACoord!16}13 & \textcolor{black!35}{0} & \textcolor{black!35}{0} & \cellcolor{M2ACoord!10}2.7 & \cellcolor{M2ACoord!12}6.0 & \textcolor{black!35}{0} & \textcolor{black!35}{0} & \cellcolor{M2ACoord!15}12 & \textcolor{black!35}{0}\\
\textcolor{M2AMind}{\textbf{E2}} & \cellcolor{M2AMind!47}63 & \cellcolor{M2AMind!20}20 & \cellcolor{M2AMind!12}6.7 & \cellcolor{M2AMind!8}0.7 & \textcolor{black!35}{0} & \textcolor{black!35}{0} & \cellcolor{M2AMind!9}1.0 & \cellcolor{M2AMind!8}0.3 & \cellcolor{M2AMind!8}0.3 & \textit{Mean} & \cellcolor{M2AReal!11}\textbf{4.4} & \textcolor{black!35}{\textbf{0}} & \textcolor{black!35}{\textbf{0}} & \cellcolor{M2AReal!9}\textbf{0.9} & \cellcolor{M2AReal!9}\textbf{2.0} & \textcolor{black!35}{\textbf{0}} & \textcolor{black!35}{\textbf{0}} & \cellcolor{M2AReal!10}\textbf{4.0} & \textcolor{black!35}{\textbf{0}}\\
\textcolor{M2AMind}{\textbf{F1}} & \cellcolor{M2AMind!56}77 & \cellcolor{M2AMind!19}17 & \cellcolor{M2AMind!12}6.7 & \cellcolor{M2AMind!10}3.0 & \cellcolor{M2AMind!12}5.7 & \cellcolor{M2AMind!13}8.3 & \cellcolor{M2AMind!11}5.0 & \cellcolor{M2AMind!14}9.3 & \cellcolor{M2AMind!10}4.0 &  &  &  &  &  &  &  &  &  & \\
\textcolor{M2AAct}{\textbf{DR2}} & \cellcolor{M2AAct!64}90 & \cellcolor{M2AAct!12}6.7 & \cellcolor{M2AAct!10}3.3 & \cellcolor{M2AAct!9}1.0 & \textcolor{black!35}{0} & \cellcolor{M2AAct!31}37 & \cellcolor{M2AAct!16}13 & \cellcolor{M2AAct!28}32 & \textcolor{black!35}{0} & \multicolumn{10}{@{}l}{\cellcolor{M2AGroup}\strut\textcolor{M2AInk}{\faIcon{gamepad}}\hspace{0.45em}\textbf{\textit{Companion \& Edutainment}}}\\
\textcolor{M2AAct}{\textbf{PR1}} & \cellcolor{M2AAct!70}100 & \cellcolor{M2AAct!33}40 & \cellcolor{M2AAct!33}40 & \cellcolor{M2AAct!20}19 & \cellcolor{M2AAct!18}16 & \cellcolor{M2AAct!10}3.0 & \cellcolor{M2AAct!35}43 & \cellcolor{M2AAct!33}40 & \cellcolor{M2AAct!14}10 & \textcolor{M2AMind}{\textbf{F2}} & \cellcolor{M2AMind!41}53 & \textcolor{black!35}{0} & \cellcolor{M2AMind!10}3.3 & \cellcolor{M2AMind!12}5.7 & \cellcolor{M2AMind!10}3.7 & \cellcolor{M2AMind!8}0.7 & \textcolor{black!35}{0} & \cellcolor{M2AMind!17}14 & \textcolor{black!35}{0}\\
\textcolor{M2AAct}{\textbf{PR2}} & \cellcolor{M2AAct!35}43 & \cellcolor{M2AAct!12}6.7 & \cellcolor{M2AAct!16}13 & \cellcolor{M2AAct!8}0.3 & \cellcolor{M2AAct!9}1.0 & \cellcolor{M2AAct!9}1.7 & \cellcolor{M2AAct!8}0.7 & \cellcolor{M2AAct!10}2.7 & \cellcolor{M2AAct!8}0.3 & \textcolor{M2AMind}{\textbf{M2}} & \cellcolor{M2AMind!66}93 & \cellcolor{M2AMind!28}33 & \cellcolor{M2AMind!14}10 & \cellcolor{M2AMind!13}7.3 & \cellcolor{M2AMind!9}2.0 & \cellcolor{M2AMind!12}6.7 & \cellcolor{M2AMind!10}4.0 & \cellcolor{M2AMind!19}18 & \cellcolor{M2AMind!10}3.3\\
\textcolor{M2AAct}{\textbf{PR3}} & \textcolor{black!35}{0} & \textcolor{black!35}{0} & \textcolor{black!35}{0} & \textcolor{black!35}{0} & \cellcolor{M2AAct!9}1.0 & \textcolor{black!35}{0} & \textcolor{black!35}{0} & \textcolor{black!35}{0} & \textcolor{black!35}{0} & \textcolor{M2AMind}{\textbf{S2}} & \cellcolor{M2AMind!70}100 & \cellcolor{M2AMind!20}20 & \cellcolor{M2AMind!14}10 & \cellcolor{M2AMind!11}4.7 & \cellcolor{M2AMind!13}7.3 & \cellcolor{M2AMind!9}1.7 & \textcolor{black!35}{0} & \cellcolor{M2AMind!14}10 & \textcolor{black!35}{0}\\
\textcolor{M2AAct}{\textbf{PR5}} & \textcolor{black!35}{0} & \textcolor{black!35}{0} & \textcolor{black!35}{0} & \textcolor{black!35}{0} & \textcolor{black!35}{0} & \textcolor{black!35}{0} & \textcolor{black!35}{0} & \textcolor{black!35}{0} & \textcolor{black!35}{0} & \textcolor{M2ACoord}{\textbf{CP05}} & \cellcolor{M2ACoord!20}20 & \textcolor{black!35}{0} & \textcolor{black!35}{0} & \textcolor{black!35}{0} & \textcolor{black!35}{0} & \textcolor{black!35}{0} & \textcolor{black!35}{0} & \textcolor{black!35}{0} & \textcolor{black!35}{0}\\
\textcolor{M2ACoord}{\textbf{CP06}} & \cellcolor{M2ACoord!14}10 & \textcolor{black!35}{0} & \textcolor{black!35}{0} & \cellcolor{M2ACoord!19}17 & \cellcolor{M2ACoord!21}21 & \cellcolor{M2ACoord!15}12 & \cellcolor{M2ACoord!8}0.7 & \cellcolor{M2ACoord!18}16 & \cellcolor{M2ACoord!8}0.3 & \textcolor{M2ACoord}{\textbf{CP10}} & \textcolor{black!35}{0} & \textcolor{black!35}{0} & \textcolor{black!35}{0} & \cellcolor{M2ACoord!9}1.3 & \cellcolor{M2ACoord!8}0.3 & \cellcolor{M2ACoord!10}2.7 & \textcolor{black!35}{0} & \cellcolor{M2ACoord!8}0.7 & \cellcolor{M2ACoord!8}0.7\\
\textcolor{M2ACoord}{\textbf{CP09}} & \textcolor{black!35}{0} & \textcolor{black!35}{0} & \textcolor{black!35}{0} & \textcolor{black!35}{0} & \textcolor{black!35}{0} & \textcolor{black!35}{0} & \textcolor{black!35}{0} & \textcolor{black!35}{0} & \textcolor{black!35}{0} & \textit{Mean} & \cellcolor{M2AReal!41}\textbf{53.3} & \cellcolor{M2AReal!15}\textbf{10.7} & \cellcolor{M2AReal!11}\textbf{4.7} & \cellcolor{M2AReal!10}\textbf{3.8} & \cellcolor{M2AReal!10}\textbf{2.7} & \cellcolor{M2AReal!9}\textbf{2.4} & \cellcolor{M2AReal!8}\textbf{0.8} & \cellcolor{M2AReal!13}\textbf{8.5} & \cellcolor{M2AReal!8}\textbf{0.8}\\
\textit{Mean} & \cellcolor{M2AReal!34}\textbf{42.6} & \cellcolor{M2AReal!14}\textbf{10.0} & \cellcolor{M2AReal!13}\textbf{7.8} & \cellcolor{M2AReal!11}\textbf{4.6} & \cellcolor{M2AReal!11}\textbf{5.0} & \cellcolor{M2AReal!12}\textbf{6.9} & \cellcolor{M2AReal!12}\textbf{7.0} & \cellcolor{M2AReal!15}\textbf{11.1} & \cellcolor{M2AReal!9}\textbf{1.7} &  &  &  &  &  &  &  &  &  & \\
\bottomrule
\end{tabularx}
\endgroup
\end{table}

\begin{axisfinding}{M2AInk}{chart-bar}{OVERALL}{How well do current systems reason, act and compose the two?}{Current systems can perform single skills but rarely combine them.}
Table~\ref{tab:main-results} ranks agents and end-to-end policies on separate leaderboards because the two families are evaluated under different settings (Section~\ref{subsec:experimental-setup}). On both leaderboards, the leading systems succeed in only a minority of episodes, with 34.6\% for GPT-6 Astra among the agents and 8.7\% for InternW0 among the end-to-end policies. The remaining seven systems stay between 2.9\% and 7.0\%, far below either leader. These low rates are not explained by the tasks themselves, because under the same evaluator a scripted controller with access to simulator state completes every task at every level (Section~\ref{subsec:data-collection}). Nor is the agent harness the only limit, since the three agents share one harness and GPT-6 Astra still succeeds more than five times as often as the next agent. The shortfall therefore most likely lies in the models themselves.

This progress is also narrow, as Mind and Act tasks, each of which tests a single capability, supply between 81\% and 100\% of the successes of every system. Within these tasks, the successes cluster on a few practiced motions. Gate passage (DR3) is the clearest case, as this single timed task supplies between 36\% and 41\% of all successes of KinRT, $\pi_{0.5}$ and X-VLA and about 70\% of those of FastWAM. At the other extreme, eight of the thirty tasks (M1, DR4, PR5, PR6, CP03, CP04, CP07 and CP09) are completed by no system in any of the 56,700 episodes.

The overall ranking hides this, because a system can lead it on single skills alone. For example, InternW0 leads the end-to-end leaderboard because of its Mind and Act results, yet its Mind2Act success rate (3.0\%) is below that of $\pi_{0.5}$ (3.6\%). The per-view results show that the two families fail in different ways, as the strongest agent retains task state but struggles to sustain contact, whereas the end-to-end policies execute practiced motions but retain little task state. Both families break down once the two capabilities must work together in one workflow. The VLAs also lose much of their memory when the scene looks different, and Mind2Act workflows collapse most steeply as the difficulty rises.
\end{axisfinding}

\begin{axisfinding}{M2AMind}{brain}{MIND}{Can it remember what it has seen?}{Agents can remember what they have seen and end-to-end policies seldom can, but no system reliably keeps track of objects that are moved.}
The Mind view asks a system to decide what to do from earlier observations, such as an item seen before, a rule learned from feedback, a layout, or the order of earlier operations. On this view, GPT-6 Astra leads by its widest margin and succeeds in 52.7\% of Mind episodes (Table~\ref{tab:main-results}). By comparison, the other two agents reach only 11.0\% and 4.3\%, and the end-to-end policies stay between 0.9\% and 5.4\%. Even these few successes of the end-to-end policies come mostly at \emph{Easy}, where there is least to remember, so these policies retain little task state of any kind.

However, GPT-6 Astra does not remember equally well on every Mind task. It solves hidden-hazard navigation (S2) in every episode and card matching (M2) in 93\%, and it also does well on button mapping (F1, 77\%) and repeated-item recognition (E1, 73\%). In all four tasks, what the agent remembers stays fixed once seen, and the agent uses it through button presses or simple placements. In contrast, the agent completes no episode of workpiece tracking (M1) or steamer-stack restoration (P2), and M1 remains unsolved by every system. Unlike the other four, these two tasks move the objects that the agent must remember. In M1 the covers are swapped while they hide the workpiece, and in P2 the baskets are unstacked and must be restacked in their original order. As a result, the agent must keep identities, positions and orders up to date while the robot itself lifts and moves the objects. Retrieving its earlier observations does not help here, because a stored image no longer shows where a swapped cover or a moved basket now is.

P2 is also the only Mind task on which the end-to-end policies outperform all three agents, as four of the six policies complete some episodes and KinRT and $\pi_{0.5}$ reach 15\% and 14\%. Because these policies retain little task state on the other Mind tasks, their P2 successes more likely come from practiced handling of the baskets than from memory. This also suggests that handling, and not memory alone, limits the agent on P2.
\end{axisfinding}

\begin{axisfinding}{M2AAct}{hand-paper}{ACT}{Can it execute precisely and on time?}{End-to-end policies act on time and the strongest agent also acts precisely, but neither family sustains control through contact.}
The Act view isolates physical execution, since each task states what to do and success depends on doing it precisely and on time. Every end-to-end policy performs best on this view, where the four VLAs reach 12.0 to 19.6\% (Table~\ref{tab:main-results}), but GPT-6 Astra still leads at 39.0\%. For the VLAs, timing is much easier than precision, as on average they succeed in 22 to 28\% of the episodes on the four dynamic tasks but in only 4 to 18\% on the six precision tasks.

GPT-6 Astra succeeds where the target can be read from the scene and reached in a few deliberate moves. It completes every resistance-adjustment episode (PR1) and 77 to 90\% of the episodes on the three timed tasks DR1 to DR3, and it also reaches 43\% on voltage setting (PR2), on which no end-to-end policy exceeds 3\%. The end-to-end policies, by contrast, succeed where one practiced motion fits the task. On gate passage (DR3), KinRT, $\pi_{0.5}$ and X-VLA succeed in 76 to 79\% of episodes, close to the agent's 80\%, and X-VLA folds the cloth (PR4) in 64\% of episodes, where the agent never succeeds.

Both kinds of success break down once a task requires sustained control, meaning that the robot must keep adjusting while it stays in contact with an object. Tasks such as DR4, PR3 and PR6 demand such control, because every small slip changes the state that the next motion must respond to. GPT-6 Astra completes none of the five tasks of this kind (DR4, PR3, PR4, PR5 and PR6), and its profile card shows that the failures occur during the contact itself (Appendix~\ref{subsec:profile-cards}). All 30 of its DR4 episodes are lost while the tray is lifted, and 21 of its 30 PR6 episodes end with dominoes knocked over during placement. Apart from PR4, which one practiced fold can cover, the end-to-end policies fare no better, and no system completes DR4, PR5 or PR6 in any of the 56,700 episodes. These failures point to missing correction during contact rather than to a misunderstanding of the task. Both families execute open-loop between observations, as the agent's motion tools turn each target into a complete planned trajectory and each end-to-end policy executes a fixed number of actions from each chunk before predicting again (Appendix~\ref{app:impl}). A ball that starts to roll or a domino that starts to tip therefore goes uncorrected until the next observation.
\end{axisfinding}

\begin{figure}[!htb]\centering
\resizebox{\linewidth}{!}{\input{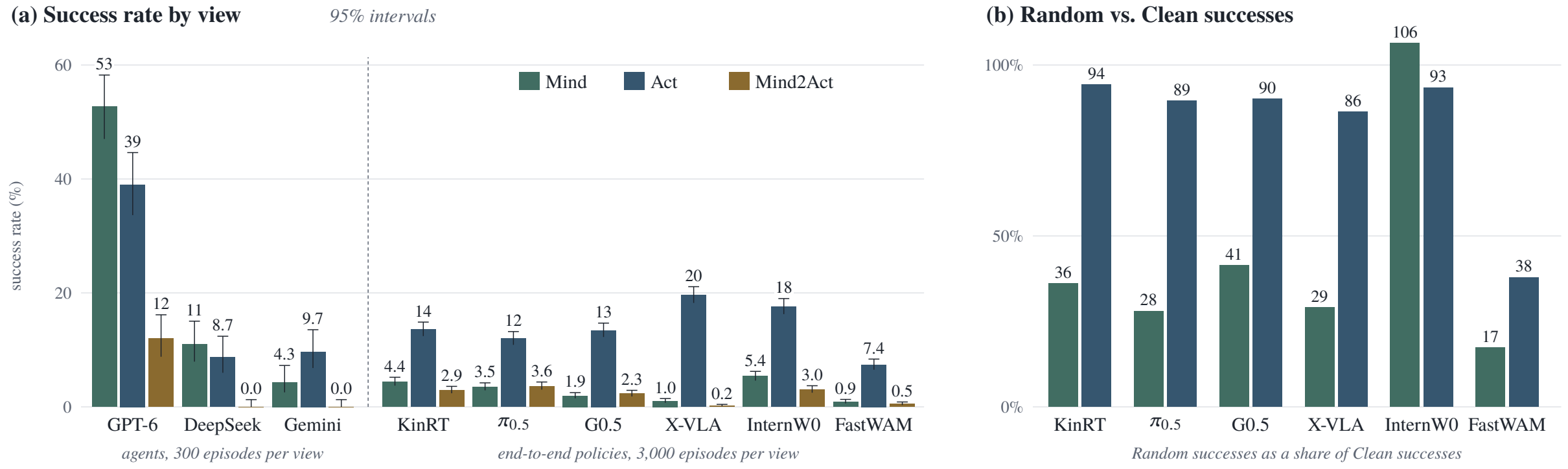}}
\caption{\textbf{Success rates by view, and \emph{Random} versus \emph{Clean} successes.} \textbf{(a)~Success rate by view.} Final success rate on \VMind, \VAct{} and \VMtA, with 95\% Wilson intervals. \textbf{(b)~Random vs.\ Clean successes.} For the six end-to-end policies, the ratio of successful \emph{Random} episodes to successful \emph{Clean} episodes on the same instances.}
\label{fig:gap-bars}
\end{figure}

\begin{axisfinding}{M2ACoord}{sync-alt}{MIND2ACT}{Do reasoning and execution work together?}{Mind2Act workflows mostly end at their first steps, before reasoning and execution ever have to work together.}
The Mind2Act view chains dependent operations into one workflow, in which remembered information guides each action and each physical outcome decides what comes next. Table~\ref{tab:main-results} and Fig.~\ref{fig:gap-bars}a show that every system succeeds less often on this view than on Act. GPT-6 Astra drops from 52.7\% on Mind and 39.0\% on Act to 12.0\% on Mind2Act, while the other two agents complete no Mind2Act episode. The end-to-end policies reach only 0.2 to 3.6\%, close to their Mind rates and far below their Act rates. For example, X-VLA, the strongest VLA on Act, drops from 19.6\% to 0.2\%. This shows that strength on a single view does not carry over to workflows that need both, and that for the end-to-end policies the missing memory, rather than execution, sets the limit.

The few Mind2Act successes come from workflows that a system can finish with steps it already performs on the single views. GPT-6 Astra succeeds reliably only on conveyor meal assembly (CP01), in 60\% of its episodes, and the picks in CP01 resemble the conveyor pickup of DR1, on which it succeeds in 77\%. CP01 alone supplies 18 of its 36 Mind2Act successes, whereas the agent completes no episode of CP03, CP04, CP07, CP09 or CP10. The end-to-end policies, by contrast, succeed mostly at \emph{Easy}, where a single order or item turns the workflow into little more than an Act task. For KinRT, $\pi_{0.5}$ and G0.5, produce weighing (CP06) and CP01 supply 81 to 88\% of all Mind2Act successes, and $\pi_{0.5}$ succeeds in 8.6\% of its Mind2Act episodes at \emph{Easy} but in only 0.1\% at \emph{Hard}. These successes therefore reflect practiced steps rather than reasoning and execution working together.

Partial scores further indicate that most Mind2Act episodes fail early rather than near completion. For every system, the Mind2Act score exceeds the success rate by at most 1.5 points, and on CP07 and CP09 no system reaches even the first scored stage. For example, G0.5 never grasps a piece in 202 of its 300 CP07 episodes (Appendix~\ref{subsec:profile-cards}). Errors in the composition itself appear only when an episode gets that far, as when InternW0 switches on the heat before all ingredients are in the pot in 271 of its 300 CP04 episodes. This suggests that Mind2Act failures are driven less by the coordination of reasoning and execution than by single steps that reasoning or execution gets wrong. Reliable composition therefore first requires each capability to hold up at every step of a workflow, which no current system achieves.
\end{axisfinding}

\begin{axisfinding}{M2ARand}{random}{APPEARANCE}{Does it hold up when the scene looks different?}{A change of appearance leaves the strongest agent's memory largely intact, whereas the VLAs lose most of theirs.}
\label{subsec:difficulty-results}%
The \emph{Clean} and \emph{Random} settings use the same task instances, and \emph{Random} only adds clutter and changes the table materials and the lighting, so comparing the two shows how much a change of appearance alone affects each system. Because all demonstrations were recorded in \emph{Clean} scenes, \emph{Random} scenes also look different from what the end-to-end policies saw in training. In overall score the change matters little, since no system gains or loses more than about three points (Appendix Table~\ref{tab:compute}).

The four fine-tuned VLAs, however, keep most of their Act successes and lose most of their Mind successes. They still complete 86 to 94\% as many Act episodes in \emph{Random} scenes as in \emph{Clean} scenes, but only 28 to 41\% as many Mind episodes (Fig.~\ref{fig:gap-bars}b). Their Mind successes fall from 97 to 35 for KinRT, from 82 to 23 for $\pi_{0.5}$, from 41 to 17 for G0.5 and from 24 to 7 for X-VLA. An Act task shows its target in the current image, so the policy can still find it among new clutter and lighting. A Mind task instead requires the policy to decide from earlier observations, and in \emph{Random} scenes those observations look unlike the demonstrations it was trained on. The VLAs therefore seem to remember how a scene looked in training rather than what it contained.

GPT-6 Astra behaves the other way round, as its Mind success rate only moves from 54.7\% to 50.7\% and its Mind2Act rate stays at 12.0\%. It appears to remember what the scene contains, such as which items it has seen and where they are, and new textures, lighting and clutter do not change that. Its Act successes drop more, from 65 to 52 of 150, so for the agent a new appearance hurts execution more than memory. The two world-action models differ from each other, as InternW0 completes 83 Mind episodes in \emph{Random} against 78 in \emph{Clean} and keeps 93\% of its Act successes, whereas FastWAM falls from 198 to 67 successes overall. For the world-action models, coping with a new appearance therefore depends on the individual model rather than on the model type.
\end{axisfinding}

\begin{axisfinding}{M2ALevel}{signal}{DIFFICULTY}{Does it hold up when the task gets harder?}{Harder levels hurt most where they add dependent operations, so Mind2Act workflows collapse while timed Act tasks barely change.}
Each harder level keeps the goal of a task and raises the demand its view is built around (Section~\ref{subsec:pressure-design}), so Mind tasks give more earlier observations to decide from, Act tasks require more precise or timely execution, and Mind2Act tasks chain more dependent operations. Under these heavier demands every system succeeds less often, but four of the six end-to-end policies lose far more than the strongest agent. Over all tasks, GPT-6 Astra keeps 60\% of its success rate from \emph{Easy} to \emph{Hard}, falling from 44.7\% to 27.0\%, whereas KinRT, $\pi_{0.5}$, G0.5 and InternW0 keep only 26 to 34\%.

This loss is not spread evenly across the views but concentrates where a harder level adds dependent operations. GPT-6 Astra falls from 64\% to 44\% on Mind and from 44\% to 34\% on Act, but from 26\% to 3\% on Mind2Act. This order follows the average length of the demonstrations, which is 23.9 seconds on Act, 58.5 seconds on Mind and 99.6 seconds on Mind2Act (Section~\ref{subsec:data-statistics}). Longer tasks contain more operations that can fail, and each harder Mind2Act level adds more orders, shapes or balls, so a workflow fails as soon as one added operation fails. The end-to-end policies follow the same pattern, with Act holding up best. For example, InternW0 falls from 13.5\% to 0.5\% on Mind and from 8.0\% to 0.3\% on Mind2Act but still reaches 12.7\% on Act at \emph{Hard}. Robustness to appearance does not protect against this loss, as InternW0's Mind rate and GPT-6 Astra's Mind2Act rate barely change between \emph{Clean} and \emph{Random} but fall to 0.5\% and 3\% at \emph{Hard}.

The timed Act tasks behave differently, because their harder levels mainly shorten the time in which the physical effect must happen. In DR2 the time allowed to press a lit button drops from 4.0 to 2.6 seconds, yet G0.5 still succeeds in 37, 39 and 34 of 100 episodes at the three levels. DR3 behaves the same way, as the gates stay open for less time while $\pi_{0.5}$ succeeds in 76, 81 and 79 (Appendix Table~\ref{tab:difficulty-counts}). A policy with a practiced motion can simply perform it faster, and less time adds no operation that could go wrong. On the precision tasks, success can even depend on the fold a level asks for more than on its difficulty. X-VLA folds the cloth (PR4) in 97 \emph{Easy}, 1 \emph{Medium} and 94 \emph{Hard} episodes of 100, so its learned motion seems to suit the rectangle and hexagon folds but not the diagonal fold of \emph{Medium}. Harder levels therefore hurt most when they add dependent operations and much less when they mainly shorten the time.
\end{axisfinding}

\begin{table}[!tp]
\centering
\caption{\textbf{Evaluation time and cost per system.} Time per call is the mean duration of one API request of an agent or one action prediction of an end-to-end policy, run locally on one GPU; episode time is the mean wall-clock time of an episode, over all tasks and by view. API cost is the mean cost of an agent episode at the providers' list prices and the cost per successful episode.}
\label{tab:efficiency-main}
\begingroup
\fontsize{7.0}{8.8}\selectfont
\setlength{\tabcolsep}{1.6pt}
\renewcommand{\arraystretch}{0.95}
\setlength{\aboverulesep}{0pt}\setlength{\belowrulesep}{0pt}%
\definecolor{ResultMind}{HTML}{416D62}\definecolor{ResultAct}{HTML}{36566F}\definecolor{ResultCoord}{HTML}{8A6A2F}
\begin{tabularx}{\linewidth}{@{}l@{\hspace{6pt}}l!{\vrule}>{\centering\arraybackslash}X!{\vrule}*{4}{>{\centering\arraybackslash}X}!{\vrule}*{2}{>{\centering\arraybackslash}X}@{}}
\toprule
\rule[-1.6mm]{0pt}{5.6mm}\textbf{Model} & \textbf{Type} & \textbf{Time} & \multicolumn{4}{c!{\vrule}}{\textbf{Episode time (min)}} & \multicolumn{2}{c}{\textbf{API cost (USD)}}\\
\cmidrule(lr){4-7}\cmidrule(l){8-9}
\rule[-1.4mm]{0pt}{5.0mm} & & \textit{/ call} & \textit{All} & \mthead{ResultMind}{M2AMindT!45}{\textbf{Mind}} & \mthead{ResultAct}{M2AActT!45}{\textbf{Act}} & \mthead{ResultCoord}{M2ACoordT!45}{\textbf{Mind2Act}} & \textit{/ episode} & \textit{/ success}\\
\midrule
\mtgroup{9}{Agents under one fixed harness (900 episodes each)} \\
\textbf{GPT-6 Astra} & Agent & 11.0\,s & 22.2 & 19.1 & 16.8 & 30.7 & 25.13 & 72.7\\
\textbf{DeepSeek-V4.1-Flash} & Agent & 35.6\,s & 96.1 & 91.9 & 93.0 & 103.2 & 1.24 & 18.9\\
\textbf{Gemini-4-Argon} & Agent & 27.8\,s & 18.0 & 15.2 & 15.8 & 22.9 & 1.12 & 24.0\\
\mtgroup{9}{End-to-end policies: VLAs and WAMs (9,000 episodes each)} \\
\textbf{KinRT} & VLA & 110.9\,ms & 4.8 & 3.5 & 3.3 & 7.5 & -- & --\\
$\boldsymbol{\pi_{0.5}}$ & VLA & 77.1\,ms & 8.6 & 5.7 & 6.5 & 13.5 & -- & --\\
\textbf{G0.5} & VLA & 170.3\,ms & 6.7 & 7.5 & 4.6 & 8.0 & -- & --\\
\textbf{X-VLA} & VLA & 153.0\,ms & 16.5 & 16.0 & 6.5 & 27.2 & -- & --\\
\textbf{InternW0} & WAM & 488.1\,ms & 19.2 & 17.6 & 7.4 & 32.5 & -- & --\\
\textbf{FastWAM} & WAM & 262.1\,ms & 12.4 & 10.9 & 4.5 & 21.9 & -- & --\\
\bottomrule
\end{tabularx}
\endgroup
\end{table}

\begin{axisfinding}{M2AInk}{stopwatch}{LATENCY \& COST}{Would it keep up in real time, and what does each success cost?}{Agent episodes take longer and cost more as the workflow grows, while end-to-end policies respond within a fraction of a second.}
\label{subsec:evaluation-efficiency}%
Simulation pauses while a model computes, so a timed task measures whether the right action is chosen and executed, not whether it is chosen in time. The pause hides a speed gap of one to nearly three orders of magnitude. As shown in Table~\ref{tab:efficiency-main}, an agent call takes 11.0 to 35.6\,s, whereas an end-to-end policy takes 77.1 to 488.1\,ms, and model requests fill 67 to 86\% of an agent episode's wall-clock time, against 2 to 29\% for the end-to-end policies.

For an agent, this time goes to deliberation and grows with how much a task leaves to be decided. GPT-6 Astra spends 16.8 minutes on an Act episode and 19.1 on a Mind episode, but 30.7 on a Mind2Act episode, where each physical outcome changes the next decision and its calls take longest (13.0\,s against 9.2\,s on Act), yet it completes only 12.0\% of these workflows. A longer run does not by itself bring more success, as DeepSeek-V4.1-Flash is the slowest agent (96.1 minutes per episode) but completes only 6.6\% of its episodes, against 34.6\% for GPT-6 Astra. GPT-6 Astra, in turn, costs about twenty times as much per episode as the other two agents (USD~25.13 against USD~1.12 to 1.24 at list price), but because it succeeds far more often, its cost per success (USD~72.7) is only about four times that of DeepSeek-V4.1-Flash (USD~18.9).

End-to-end policies are much faster, and X-VLA, for example, predicts in 153\,ms, about seventy times faster than a GPT-6 Astra call. It finishes an Act episode in 6.5 rather than 16.8 minutes and reaches 19.6\% on Act, the highest among end-to-end policies and half the agent's 39.0\%. Speed also varies among the end-to-end policies, as InternW0 succeeds slightly more often than KinRT (8.7\% against 7.0\%) but needs 19.2 rather than 4.8 minutes per episode. The agent's lead on timed tasks depends on the paused clock, because many of GPT-6 Astra's Act successes come from conveyor pickup (DR1, 77\%) and light-cued response (DR2, 90\%), while its mean request time on Act tasks (9.2\,s) is more than three times the 2.6\,s response window of the \emph{Hard} light-cued task. In a real setting, the conveyor would keep moving and the light keep changing during that time.
\end{axisfinding}

\begin{axisfinding}{M2AReal}{globe-asia}{REAL WORLD}{How far is each application domain from deployment?}{Domains built on pressing and placing reach at best about half of episodes, and kitchens and assembly remain far from deployment.}
Each task is built from a need in one of five application domains (Table~\ref{tab:application-domains}), so the per-task results can also be read by domain (\emph{Mean} rows of Table~\ref{tab:per-task-sr}). GPT-6 Astra succeeds in 53.3\% of the companion and edutainment episodes and in 46.0\% and 42.6\% of those in manufacturing and logistics and in instrument and tool use, whose tasks mostly come down to pressing buttons, setting a slider or placing an item. It falls to 17.9\% in home and food service and to 4.4\% in assembly, where at least half of the tasks are Mind2Act workflows.

The end-to-end policies rank the domains differently, as they do best in manufacturing and logistics (KinRT 20.9\%), where conveyor pickup and gate passage repeat one practiced motion, and stay below 12\% in every other domain. Four of the five domains each hold two tasks that no system completes, namely kitchen loading and cooking (CP03, CP04), workpiece tracking and tray carrying (M1, DR4), oriented stamping and whiteboard restoration (PR5, CP09), and the domino chain and bridge building (PR6, CP07).

The best domain value, 53.3\%, still means failing about every second attempt, so no domain is yet close to deployment even in simulation. Agents come closest in edutainment and manufacturing, end-to-end policies come closest in repetitive factory motions, and neither comes close in kitchens or assembly.
\end{axisfinding}

\subsection{Which System Suits Which Task}
\label{subsec:suited}
The leaderboards and the analysis of Section~\ref{subsec:main-results} show how the systems rank and where each of them succeeds and fails. Together they show that agents and end-to-end policies currently handle different kinds of task demand.
\paragraph{Agents for reasoning-heavy steps.} A strong agent is suited to the steps of a task that turn on what was seen before and on what to do next. GPT-6 Astra remembers and reasons well (52.7\% on Mind), keeps that memory when the scene changes, and solves tasks whose target can be read and reached in a few deliberate moves. Its limits lie in contact and timing. It completes none of the five contact-rich tasks, and its timed successes are measured with a simulator that waits for it. The other two agents use the same tools yet differ widely on Mind, so this profile is specific to the model.

\paragraph{End-to-end policies for execution-heavy steps.} End-to-end policies are suited to steps that one practiced motion can cover. The four evaluated VLAs retain 86--94\% of their Act successes under scene randomization, and each has its own strength: G0.5 on timed reactions, X-VLA on resistance adjustment and cloth folding, and KinRT and $\pi_{0.5}$ on memory, where the VLAs show little. They retain almost no task state, so any step that depends on earlier information is better left to another system.

\paragraph{World-action models differ widely.} InternW0 is the only end-to-end policy whose memory survives scene changes, and it nearly matches the agent on the blueprint task. It still follows its learned sequence without checking preconditions, as in cooking. FastWAM drops in both views under scene randomization, so this advantage is not inherent to world-action models.

Some demands are met by no system yet. No evaluated system succeeds on M1, and GPT-6 Astra completes no episode of P2. No system sustains contact under force (DR4, PR5 and PR6 are never completed), and none completes Mind2Act workflows reliably. GPT-6 Astra reaches 12.0\% on Mind2Act, and no end-to-end policy exceeds 3.6\%. Neither family yet covers both reasoning and execution.

\section{Diagnostic Analysis}
\label{sec:diagnostics}

Success rates measure task completion, and task scores measure partial progress. But they cannot identify where or why a system fails within an episode, which is what model developers need to know. In this section, we examine where tasks fail, which errors recur across systems, and how these findings inform improvement priorities. Episode-level diagnosis and model profile cards (Section~\ref{subsec:diagnosis-harness}) connect failure patterns to task progress, budget use, and stopping decisions (Section~\ref{subsec:card-findings}), and the failure codes point to directions for improvement (Section~\ref{subsec:diag-directions}). Inference latency, runtime and API cost are analyzed alongside success rates in Section~\ref{subsec:evaluation-efficiency}.

\subsection{Diagnosis Harness and Model Profile Cards}
\label{subsec:diagnosis-harness}
\paragraph{Diagnosis harness design.} We classify failures by the first unrecoverable error identified in the episode, using 21 failure codes (Appendix~\ref{subsec:error-taxonomy}). Mind errors concern perception and task reasoning, Act errors concern physical execution, and composition errors concern using physical outcomes to guide subsequent actions. Three episode-level codes capture timeouts, stalls, and cases with insufficient evidence. The label follows the observed error, regardless of the task's view. A Mind failure involves an error in perception or task reasoning. Examples include selecting the wrong object based on a misremembered order (\texttt{M-MEMORY}) and incorrectly predicting how the scene will change (\texttt{M-PREDICT}). An Act failure occurs when the system selects an appropriate action but executes it incorrectly. Examples include losing a grasp (\texttt{A-GRASP}) and executing the correct action too early or too late (\texttt{A-TIMING}). A composition failure occurs when the system proceeds without verifying a required outcome or fails to update its next action based on feedback. For example, the system may continue pressing the same button after the light cue changes (\texttt{C-NO-UPDATE}).

\begin{figure}[!t]\centering
\resizebox{\linewidth}{!}{\input{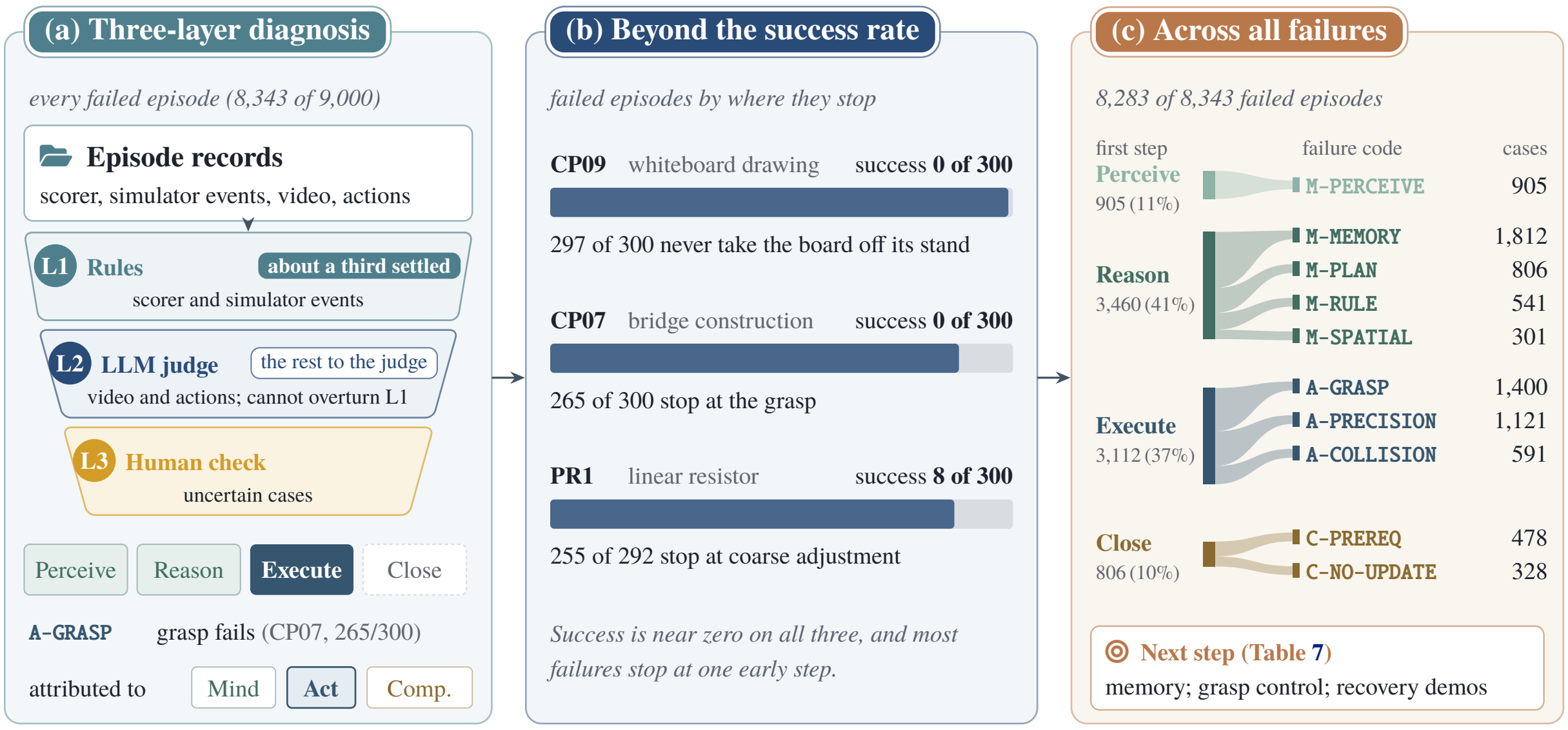}}
\caption{\textbf{Failure diagnosis on an example profile card.} An illustrative example, the same as in Fig.~\ref{fig:profile-card-example} and on the project website; the measured cards of three representative systems are in Appendix~\ref{subsec:profile-cards}. \PanHead{M2APanA}{(a)~Three-layer diagnosis.} The workflow as a funnel: in this example, the rule layer settles about a third of the failed episodes, and the rest pass to the LLM judge and, where it is uncertain, to the human check. \PanHead{M2APanB}{(b)~Beyond the success rate.} The example end-to-end policy rarely completes these three tasks, and most of its failures on each task occur at a single early step. \PanHead{M2APanC}{(c)~Across all failures.} The card follows the failed episodes from the step at which each first breaks to its failure code.}
\label{fig:diagnosis-card}
\end{figure}

The harness combines three layers of review as in Fig.~\ref{fig:diagnosis-card}a. First, deterministic rules use scorer outputs and simulator events to isolate the failure or bound the step at which it occurred. For cases left open by rules, an LLM judge inspects the video and action or tool-call history without overturning rule-based verdicts, flagging ambiguous cases as uncertain. Finally, human reviewers settle the cases the judge marks as uncertain. Fig.~\ref{fig:diagnosis-card} shows the workflow on an illustrative example, and Appendix~\ref{subsec:profile-cards} reports the results of this procedure for every failed episode of three representative systems.

\begin{figure}[!htb]\centering
\resizebox{\linewidth}{!}{\input{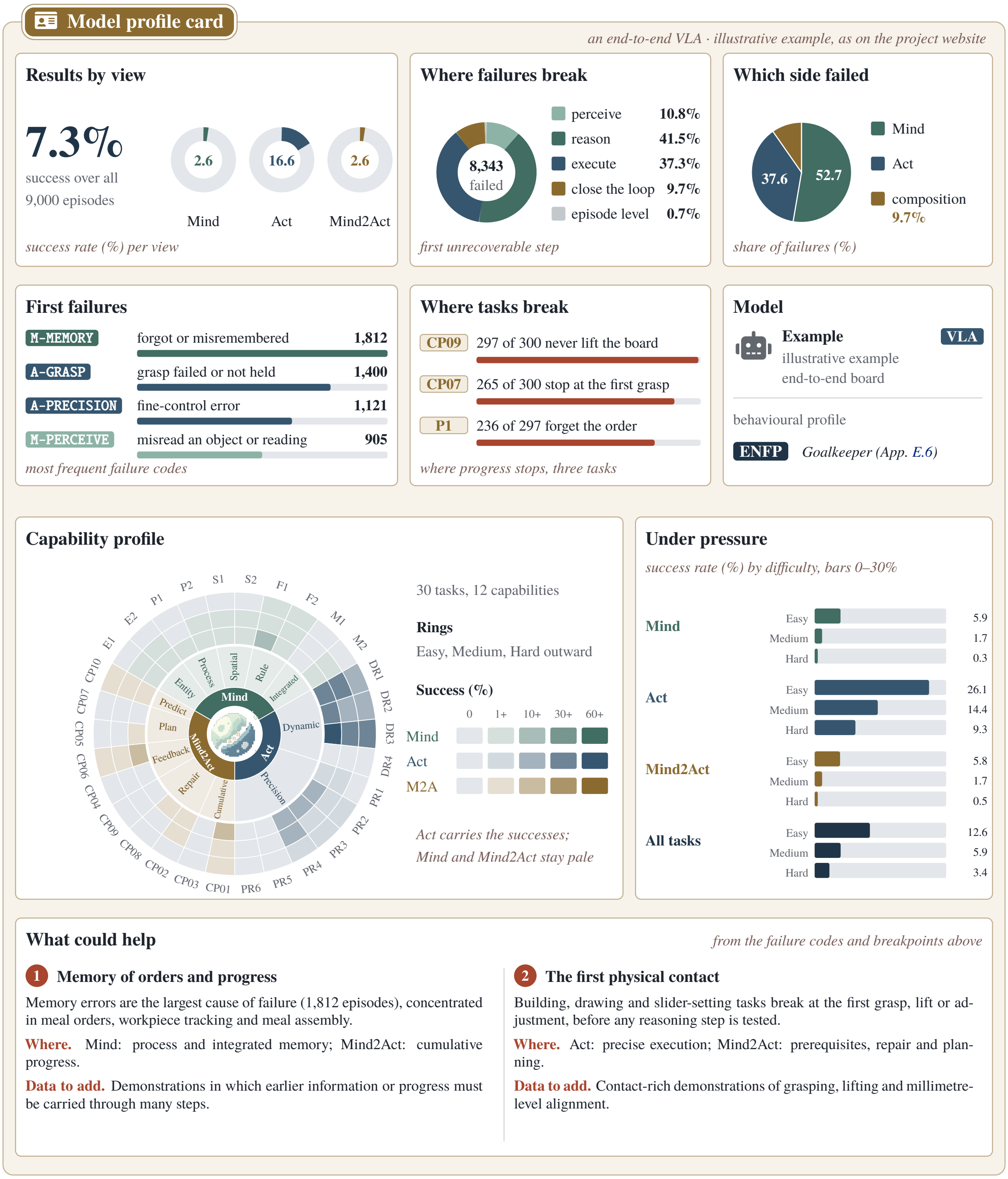}}
\caption{\textbf{Overview of the model profile card.} The card reads from top to bottom, as on the project website. \emph{Results by view} gives the success rate in each view. \emph{Where failures break} and \emph{Which side failed} give the step at which failed episodes first break down and whether \VMind, \VAct{} or their composition failed. \emph{First failures} and \emph{Where tasks break} give the most frequent failure codes and where typical tasks stop. \emph{Model} gives a four-letter behavioral profile (Appendix~\ref{app:profiles-fun}). \emph{Capability profile} and \emph{Under pressure} give the success of each task at Easy, Medium and Hard, grouped by capability and view, and the success rate by difficulty. \emph{What could help} turns the codes into directions for improvement. The numbers are an illustrative example; the measured cards of three representative systems are given in Appendix~\ref{subsec:profile-cards}.}
\label{fig:profile-card-example}
\end{figure}

\paragraph{Model profile cards.} A model profile card summarizes the failure profile of a single system. Fig.~\ref{fig:diagnosis-card} walks through the diagnosis of an illustrative end-to-end policy, and Fig.~\ref{fig:profile-card-example} shows the full card format with illustrative numbers. The cards of three representative evaluated systems are given in Appendix~\ref{subsec:profile-cards}. 

\subsection{Failure Patterns of the Evaluated Systems}
\label{subsec:card-findings}
We read the measured cards of G0.5, InternW0 and GPT-6 Astra (Appendix~\ref{subsec:profile-cards}), together with the run logs of the agents.

\paragraph{The overall pattern.}
Physical execution is the largest diagnosed failure group in all three systems, accounting for 41.4\% of G0.5's failures, 37.2\% of InternW0's, and 55.3\% of GPT-6 Astra's. GPT-6 Astra's failures are particularly concentrated in execution, whereas G0.5 and InternW0 also show substantial perception and reasoning failures, at 27.7\% and 32.8\%, respectively; 6--11\% of each system's failures occur while closing the loop. These proportions refer to the first unrecoverable error identified in each failed episode. The rule layer alone would overstate Mind-side errors for the end-to-end policies (Appendix Table~\ref{tab:rule-layer-counts}), because scorer records reveal a misremembered item more readily than video reveals a motor fault. A card therefore combines the layers.

The dominant execution errors differ across systems. Failed or lost grasps are the most frequent errors for G0.5 and GPT-6 Astra, with 1,324 and 120 cases, while fine-control errors lead InternW0's profile with 1,132 cases. The two end-to-end policies also record 924 and 1,011 memory failures. These profiles identify distinct priorities: grasp reliability for G0.5 and GPT-6 Astra, fine control for InternW0, and task-state retention for both end-to-end policies.

\paragraph{Differences across tasks.}
The dominant failure also varies by task, even within the same system. For G0.5, 202 of 300 bridge-building episodes (CP07) fail to grasp a construction piece (Fig.~\ref{fig:profile-card-g05}), making object acquisition the main obstacle to further progress. In light-cued pressing (DR2), however, 90 of 190 failures involve a misread light (\texttt{M-PERCEIVE}) and 88 involve a response that is not updated after the light changes (\texttt{C-NO-UPDATE}). Only eight concern the press itself or its timing. In incremental meal orders (P1), memory errors account for 210 of G0.5’s 300 failed episodes and 245 of InternW0’s 298 failed episodes.

\paragraph{Where progress stops.}
No evaluated system completes whiteboard restoration (CP09) or bridge building (CP07), and failures on such workflows often occur before their reasoning stages are reached. On CP09, 299 of 300 G0.5 episodes stop before board removal. GPT-6 Astra fails during the lift in all 30 tray-carrying (DR4) episodes (Fig.~\ref{fig:profile-card-astra}). Zero success on these tasks is therefore consistent with early physical bottlenecks rather than failures in downstream planning or repair. Most episodes fail in Mind or Act before closed-loop composition is tested.

\paragraph{Analysis of budget use.}\label{subsec:failure-recovery}
Under the same decision budget, the agent logs show distinct termination patterns (Appendix Table~\ref{tab:agent-stops}). Gemini-4-Argon voluntarily stops in 829 of 900 episodes and uses approximately 33 decisions per episode. DeepSeek-V4.1-Flash uses 111.8 decisions on average, voluntarily stops in 385 episodes, and exhausts the decision and physics budgets in 286 and 94 episodes, respectively. GPT-6 Astra averages 79.6 decisions per episode. It voluntarily stops in 460 episodes, and only one of them succeeds, a gate-passage episode (DR3) whose task was already complete when it stopped. The other 459 account for 77.9\% of its 589 failed episodes, and in 264 of them the agent has used no more than 100 of the 150 allowed decisions. Among these 459 stops, the first unrecoverable error is attributed to execution in 289 cases, perception or reasoning in 127, and composition in 28. The remaining 15 cases fall outside these three capability groups.

These results distinguish resource limits from the errors that prevent task completion. DeepSeek frequently exhausts its budget. GPT-6 Astra shows a different pattern: many unsuccessful episodes end with substantial decision budget remaining, and execution errors dominate its voluntary stops, often after its motion planner rejects the requested poses. For episodes affected by execution errors, additional decisions are useful only if they enable effective retries or corrective actions (Table~\ref{tab:diag-directions}).

\subsection{From Failure Codes to Directions}
\label{subsec:diag-directions}
A diagnosis helps a model developer only if it says what to change, and the profiles narrow broad improvement goals to specific behaviors. Because each failure code names the step and the side of the loop at which an episode broke, a group of codes points to the capability that most needs to improve. An evaluation speaks most directly to two levers that a developer can adjust without redesigning the model. One is the training data of an end-to-end policy. The other is the harness of an agent, that is, the tools and control loop through which it acts. Table~\ref{tab:diag-directions} maps the three groups of the profile card, and the breakpoints shared by all of them, to such directions. The paragraphs below give the evidence behind each.

\begin{table}[!htb]\centering\footnotesize
\caption{\textbf{Directions for improvement by failure group.} Grouped as on the profile card (\VMind, \VAct{} and \textcolor{M2ACoord}{\textbf{Composition}}). The directions concern the training data of end-to-end policies and the harness of agents. Evidence is from the measured cards of Appendix~\ref{subsec:profile-cards} and the results of Section~\ref{sec:results}.}
\label{tab:diag-directions}
\setlength{\tabcolsep}{3.5pt}
\begin{tabularx}{\linewidth}{@{}>{\raggedright\arraybackslash}p{24mm}>{\raggedright\arraybackslash}p{44mm}>{\raggedright\arraybackslash}X@{}}\toprule
Failure group & Evidence in this report & Direction\\\midrule
\textcolor{M2AMind}{\textbf{Mind}}: \texttt{M-MEMORY}, \texttt{M-SPATIAL}, \texttt{M-RULE} & 924 misremembered-information failures of G0.5 and 1,011 of InternW0; VLAs lose 59--72\% of Mind successes under \emph{Random} scenes & \textbf{Task state across stages}: keeping what was seen and recognizing items under a new appearance. \emph{Data}: varied scenes and longer gaps between when information is shown and when it is needed\\\addlinespace[2pt]
\textcolor{M2AAct}{\textbf{Act}}: \texttt{A-GRASP}, \texttt{A-PRECISION}, \texttt{A-TIMING} & 1,324 grasp failures of G0.5 and 1,132 fine-control failures of InternW0; 202 of 300 G0.5 episodes on CP07 never grasp a piece & \textbf{Grasp, contact and timing}: reliable grasps and fine contact, and responses within the windows of DR2 and DR3. \emph{Data}: demonstrations weighted toward these segments, with more variation in pose, contact and timing. \emph{Harness}: finer motion tools\\\addlinespace[2pt]
\textcolor{M2ACoord}{\textbf{Composition}}: \texttt{C-NO-VERIFY}, \texttt{C-NO-REPAIR}, \texttt{C-PREREQ} & 761 precondition failures of InternW0, 271 of them on CP04, which heats the pot before all ingredients are in; agents that stop with budget left (Section~\ref{subsec:failure-recovery}) & \textbf{Checking and repairing outcomes}: checking a step before the next one and repairing it when it failed. \emph{Data}: demonstrations that include failed steps and their repair. \emph{Harness}: the outcome of each step returned, a check required before moving on, and a retry before stopping\\\addlinespace[2pt]
\textbf{All groups}, by breakpoint & 299 of 300 G0.5 episodes on CP09 end before the board is lifted & \textbf{Targeted data for end-to-end policies}: demonstrations that start from the states in which episodes stop and from perturbed states near them, instead of new full-length trajectories\\\bottomrule
\end{tabularx}
\end{table}

\paragraph{\VMind{} failures: task state across stages.} Misremembered information is the most frequent \VMind-side code of both measured end-to-end policies, with 924 failures of G0.5 and 1,011 of InternW0, and the VLAs lose 59--72\% of their \VMind{} successes when the scene is randomized (Section~\ref{subsec:task-results}). These failures point to maintaining task-relevant identities, order, and progress across operations, and to recognizing an item under a new appearance. Closing this gap calls for data that stress what the failures expose, with varied scene appearance and longer gaps between the moment information is shown and the moment it is needed.

\paragraph{\VAct{} failures: grasp, contact and timing.} Failed or lost grasps account for 1,324 failures of G0.5 and fine-control errors for 1,132 of InternW0. On bridge building (CP07), 202 of 300 G0.5 episodes never grasp a piece. G0.5's frequent grasp failures motivate training and control changes that improve contact acquisition and object retention. InternW0's fine-control errors motivate closer examination of action resolution and feedback during manipulation. Responses must also fall within the short windows of timed tasks such as the light-cued button (DR2) and the cyclic gates (DR3). On DR3, most failures of the two end-to-end policies come from the withdrawal after delivery rather than from the gate timing. The arm is still inside the gate corridor when the gate closes, or touches the gate while it closes. Demonstrations can be weighted toward these segments, with more variation in object pose, contact and timing, instead of being spread evenly over whole episodes.

\paragraph{Composition failures: checking and repairing outcomes.} InternW0 has 761 precondition failures, 271 of them on cooking (CP04), where it switches on the heat before all ingredients are in the pot. Each step is executed, but its outcome is not checked before the next one starts. Such failures call for checking whether the required physical state has been achieved before proceeding, and repairing a step that failed. The scripted demonstrations show only steps that succeed, so a policy never sees a failed step followed by its repair. Recovery demonstrations can begin at frequently observed failures, such as a slipped grasp or an incomplete placement, and show how to restore progress.

\paragraph{Breakpoints: targeted data for end-to-end policies.} These diagnoses also guide data collection. Because an end-to-end policy learns its behavior from demonstrations, the cards show where new data for it should be built. On whiteboard restoration (CP09), 299 of 300 G0.5 episodes end before the board is lifted off its stand, so the later stages of the task are never reached. Adding full-length demonstrations would spend most of their steps on later stages, which the policy almost never reaches. Demonstrations constructed around the breakpoint instead, starting from the states in which episodes stop and from perturbed states near them, concentrate the training data where the policy fails.

\paragraph{Agents: the harness.} For agents, the same codes also point to the harness. GPT-6 Astra breaks down mostly during execution and often stops after its motion planner rejects the requested poses (Section~\ref{subsec:card-findings}), so finer motion tools and a retry step before stopping come first. Tool feedback should help distinguish an ineffective request from a completed operation and support an appropriate retry or alternative action. The harness can also return the outcome of each step and require a check before the next one, which addresses composition failures. It can also keep a record of earlier failures, so that an agent recognizes a breakpoint it has met before.

\FloatBarrier

\Needspace{9\baselineskip}
\section{Limitations and Future Directions}
\label{sec:limitations}
\subsection{Limitations}
\paragraph{Limited causal attribution.} Each view is a set of tasks, not an intervention on one system. A Mind task still requires routine manipulation, an Act task still requires the robot to perceive its cue, and the Mind2Act workflows are longer than the single-view tasks and differ from them in content (Section~\ref{subsec:score-success}). A low Mind2Act result therefore cannot yet be separated from the effect of a longer horizon or of an early execution error. The diagnosis harness attributes failures within each episode, where rules settle 22 to 38\% of the failures of the three diagnosed systems and the LLM judge, with human review of uncertain cases, attributes the rest.

\paragraph{Paused clock and a fixed agent harness.} Simulation pauses during inference, so timed tasks test whether an action is chosen and timed correctly in simulation time, not in real time (Section~\ref{subsec:evaluation-efficiency}). The agents also receive an inverse-kinematics and motion-planning service and four times the physics budget of the end-to-end policies, and they are not trained on the benchmark. All three agents use the same fixed harness, so the results do not show what a model would achieve with its own tools, controllers or memory. We therefore rank agents and end-to-end policies on separate leaderboards and treat the lead of the strongest agent as a result under the paused clock rather than as real-time ability.

\paragraph{Limited system coverage.} The results cover three agents, four VLAs and two WAMs, and they include no hybrid or memory-augmented system. This is enough to compare the two families but too few for a broader analysis, for example of how model size, training data or architecture shapes the results on Mind, Act and Mind2Act. Each agent is also evaluated on only ten episodes per task and level, so its results by difficulty level are coarse. Some episodes were run again after infrastructure failures (Appendix~\ref{app:impl}).

\paragraph{Limited embodiments and settings.} All tasks are tabletop manipulation tasks in simulation on one bimanual platform with parallel grippers, and each task and level uses one fixed English instruction. Other embodiments and settings, such as dexterous hands, mobile manipulation, navigation and humanoid robots, are not yet covered. The \emph{Random} setting changes appearance over a wide range and is not meant to model a deployment scene. Each capability category contains between one and six tasks, which is too few to support claims about a single category. How the results in simulation relate to results on the physical robot has not yet been measured.

\subsection{Future Directions}
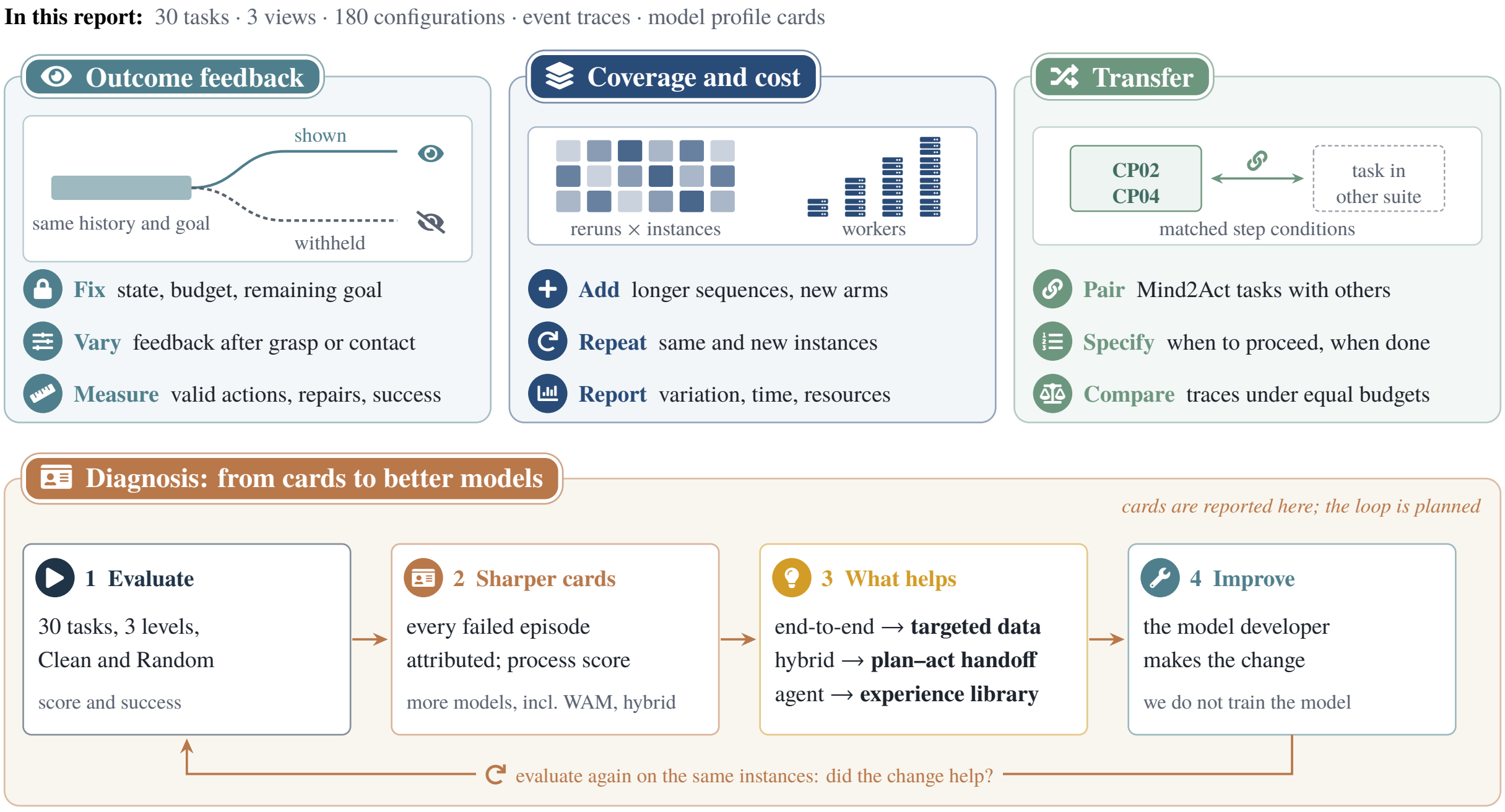
\begin{figure}[!t]\centering
\resizebox{\linewidth}{!}{\begingroup%
\definecolor{fInk}{HTML}{1D232C}\definecolor{fNavy}{HTML}{203447}%
\definecolor{fMind}{HTML}{416D62}\definecolor{fAct}{HTML}{36566F}%
\definecolor{fGrey}{HTML}{5A6270}\definecolor{fWait}{HTML}{D9DCE1}%
\definecolor{cA}{HTML}{4F7F8C}\definecolor{cAT}{HTML}{EEF4F6}%
\definecolor{cB}{HTML}{294B77}\definecolor{cBT}{HTML}{EDF2F6}%
\definecolor{cS}{HTML}{6D967F}\definecolor{cST}{HTML}{EEF4F1}%
\definecolor{cC}{HTML}{B9784A}\definecolor{cCT}{HTML}{F7F2E8}%
\definecolor{cAmber}{HTML}{D29C27}%
\fontencoding{T1}\fontfamily{qtm}\selectfont%
\def\fsB{\fontsize{8.4pt}{9.8pt}\selectfont}%
\def\fsD{\fontsize{7pt}{8.2pt}\selectfont}%
\def\fsE{\fontsize{6.2pt}{7.2pt}\selectfont}%
\def\FdPill#1#2{\tikz[baseline=(p.base)]\node[fill=#1,text=white,rounded corners=1.4mm,inner xsep=1.6mm,inner ysep=0.7mm,
  outer sep=0pt,font=\fsB\bfseries](p){#2};}%
\begin{tikzpicture}[x=1mm,y=-1mm,every node/.style={inner sep=0pt,outer sep=0pt},
  panel/.style={draw=#1!50,line width=0.5pt,rounded corners=1.6mm,fill=#1T!70},
  tab/.style={draw=#1!65,fill=white,line width=0.5pt,rounded corners=2mm,inner xsep=0.5mm,inner ysep=0.5mm,anchor=west},
  sketch/.style={draw=#1!40,line width=0.45pt,fill=white,rounded corners=0.9mm},
  txt/.style={font=\fsD,text=fInk,anchor=base west},
  small/.style={font=\fsE,text=fGrey},
  ic/.style={circle,fill=#1,text=white,minimum size=4.2mm,font=\fsE},
  flow/.style={draw=#1,line width=0.7pt,-{Stealth[length=1.6mm,width=1.4mm]}}]

\path (0,-0.8) -- (160,-0.8);
\node[anchor=base west,font=\fsD] at (0,2.4) {\textbf{In this report:}\enspace\textcolor{fGrey}{30 tasks \textperiodcentered{} 3 views \textperiodcentered{} 180 configurations \textperiodcentered{} event traces \textperiodcentered{} model profile cards}};

\def\FdRows#1#2#3#4{\foreach \i/\l/\t [count=\k] in {#4}{
  \pgfmathsetmacro\yy{#2+(\k-1)*5.6}
  \node[ic=#3] at (#1+2.1,\yy-0.9) {\i};
  \node[txt] at (#1+5.4,\yy) {\textbf{\textcolor{#3}{\l}}\enspace\t};}}

\draw[panel=cA] (0,8) rectangle (52,45);
\node[tab=cA] at (1.8,8) {\FdPill{cA}{\faEye\enspace Outcome feedback}};
\draw[sketch=cA] (2,12.2) rectangle (50,27.8);
\fill[cA!55,rounded corners=0.4mm] (5,18.6) rectangle (20,21.2);
\node[small,anchor=base] at (12.5,24.4) {same history and goal};
\draw[cA,line width=0.8pt] (20,19.9) .. controls (25,19.9) and (25,16.0) .. (30,16.0) -- (42,16.0);
\draw[fGrey,line width=0.8pt,dash pattern=on 1.4pt off 1pt] (20,19.9) .. controls (25,19.9) and (25,23.4) .. (30,23.4) -- (42,23.4);
\node[font=\fsD,text=cA] at (45.6,16.2) {\faEye};\node[small,anchor=base west,text=cA] at (31,15.0) {shown};
\node[font=\fsD,text=fGrey] at (45.6,23.6) {\faEyeSlash};\node[small,anchor=base west] at (31,26.5) {withheld};
\FdRows{2}{31.6}{cA}{\faLock/Fix/{state, budget, remaining goal},\faSlidersH/Vary/{feedback after grasp or contact},\faRuler/Measure/{valid actions, repairs, success}}

\draw[panel=cB] (54,8) rectangle (106,45);
\node[tab=cB] at (55.8,8) {\FdPill{cB}{\faLayerGroup\enspace Coverage and cost}};
\draw[sketch=cB] (56,13.4) rectangle (104,26);
\foreach \r in {0,1,2}\foreach \c in {0,...,5}{
  \pgfmathsetmacro\op{0.25+0.15*mod(\r*3+\c*2,5)}
  \fill[cB,opacity=\op,rounded corners=0.25mm] (59+\c*3.3,14.8+\r*2.7) rectangle ++(2.6,2.3);}
\node[small,anchor=base] at (68.6,25.0) {reruns $\times$ instances};
\foreach \i/\x in {0/87,1/91,2/95,3/99}{\foreach \j in {0,...,\i}{\node[font=\fsE,text=cB] at (\x,22.0-\j*2.2) {\faServer};}}
\node[small,anchor=base] at (93,25.0) {workers};
\FdRows{56}{31.6}{cB}{\faPlus/Add/{longer sequences, new arms},\faRedo/Repeat/{same and new instances},\faChartBar/Report/{variation, time, resources}}

\draw[panel=cS] (108,8) rectangle (160,45);
\node[tab=cS] at (109.8,8) {\FdPill{cS}{\faRandom\enspace Transfer}};
\draw[sketch=cS] (110,13.4) rectangle (158,26);
\draw[cS,line width=0.45pt,fill=cST,rounded corners=0.6mm] (114,15.4) rectangle (128,22.4);
\node[font=\fsE\bfseries,text=fMind] at (121,18.0) {CP02};\node[font=\fsE\bfseries,text=fMind] at (121,20.8) {CP04};
\draw[fGrey!60,line width=0.45pt,fill=white,rounded corners=0.6mm,dash pattern=on 1.2pt off 0.8pt] (140,15.4) rectangle (154,22.4);
\node[font=\fsE,text=fGrey] at (147,18.0) {task in};\node[font=\fsE,text=fGrey] at (147,20.8) {other suite};
\draw[cS,line width=0.7pt,{Stealth[length=1.3mm,width=1.1mm]}-{Stealth[length=1.3mm,width=1.1mm]}] (129,18.9) -- (139,18.9);
\node[font=\fsE,text=cS] at (134,17.0) {\faLink};
\node[small,anchor=base] at (134,25.0) {matched step conditions};
\FdRows{110}{31.6}{cS}{\faLink/Pair/{Mind2Act tasks with others},\faListOl/Specify/{when to proceed, when done},\faBalanceScale/Compare/{traces under equal budgets}}

\draw[panel=cC] (0,51) rectangle (160,86);
\node[tab=cC] at (1.8,51) {\FdPill{cC}{\faIdCard\enspace Diagnosis: from cards to better models}};
\node[small,anchor=base east,text=cC] at (158,54.6) {\textit{cards are reported here; the loop is planned}};
\def\FdStation#1#2#3#4{%
  \draw[draw=#2!55,line width=0.5pt,fill=white,rounded corners=0.9mm] (#1,58) rectangle ++(35,20.4);
  \node[ic=#2] at (#1+3.4,61.6) {#3};
  \node[txt,font=\fsD\bfseries,text=#2] at (#1+6.6,62.5) {#4};}
\FdStation{2}{fNavy}{\faPlay}{1\enspace Evaluate}
\node[txt] at (3.6,67.6) {30 tasks, 3 levels,};
\node[txt] at (3.6,71.2) {Clean and Random};
\node[small,anchor=base west] at (3.6,75.6) {score and success};
\draw[flow=cC] (37.2,68.2) -- (41.0,68.2);
\FdStation{41.4}{cC}{\faIdCard}{2\enspace Sharper cards}
\node[txt] at (43,67.6) {every failed episode};
\node[txt] at (43,71.2) {attributed; process score};
\node[small,anchor=base west] at (43,75.6) {more models, incl.\ WAM, hybrid};
\draw[flow=cC] (76.6,68.2) -- (80.4,68.2);
\FdStation{80.8}{cAmber}{\faLightbulb}{3\enspace What helps}
\node[txt] at (82.4,67.6) {end-to-end $\rightarrow$ \textbf{targeted data}};
\node[txt] at (82.4,71.2) {hybrid $\rightarrow$ \textbf{plan--act handoff}};
\node[txt] at (82.4,74.8) {agent $\rightarrow$ \textbf{experience library}};
\draw[flow=cC] (116,68.2) -- (119.8,68.2);
\FdStation{120.2}{cA}{\faWrench}{4\enspace Improve}
\node[txt] at (121.8,67.6) {the model developer};
\node[txt] at (121.8,71.2) {makes the change};
\node[small,anchor=base west] at (121.8,75.6) {we do not train the model};
\draw[flow=cC] (137.7,78.4) -- (137.7,82.6) -- (19.5,82.6) -- (19.5,78.8);
\node[small,anchor=base,fill=cCT!70,inner xsep=1mm,text=cC] at (78.6,83.5) {\faRedo\enspace evaluate again on the same instances: did the change help?};
\end{tikzpicture}%
\endgroup%
}%
\caption{\textbf{Planned work beyond this report.} \textbf{Top:} \PanHead{M2APanA}{Outcome feedback}, matched runs with and without it; \PanHead{M2APanB}{Coverage and cost}, broader and repeated evaluation with its cost; \PanHead{M2APanS}{Transfer}, the analysis carried to other simulators. \textbf{Bottom:} \PanHead{M2APanC}{Diagnosis}, sharper profile cards for more models and the planned loop from a card to a better model that is evaluated again. The small drawings are schematic.}
\label{fig:future-directions}
\end{figure}

We group the planned extensions into four directions that match the panels of Fig.~\ref{fig:future-directions}, and the leaderboards will be updated with new systems, tasks and results as these extensions are added.
\paragraph{Composition and outcome feedback.} The current views show where a system fails, and the next step is to show why. On the Mind2Act tasks, an oracle for Mind will supply the task state that would otherwise have to be remembered or inferred, and an oracle for Act will let a system call scripted skills for the physical operations. The gain from each oracle shows how much that capability contributes, and the gap that remains with both oracles shows the cost of composition. Each Mind2Act workflow will also be paired with the Mind and Act tasks that share its demands, and progress will be reported for each stage as the share of episodes that pass it among those that reach it, which removes the effect of horizon length. Matched pairs of runs that start from the same task state and differ only in whether the system receives feedback on a grasp, a placement or a contact will then measure how much the outcome of an action guides the next decision.

\paragraph{Broader and fairer comparison.} The leaderboards will cover more systems and settings, so that the comparison between the two families no longer rests on one fixed setup. The task suite will also grow beyond tabletop bimanual manipulation to dexterous hands, mobile manipulation, navigation and humanoid robots, so that the three views can be tested on other embodiments. Human teleoperation and a scripted expert acting through each interface will give every task a reachable upper bound, and a version of the protocol that adds inference time to the simulation clock will show how much of the agents' result on timed tasks remains in real time. A sandbox setting with its own leaderboard will let an agent bring its own harness, which runs in an isolated container with access only to the cameras, the robot state and joint-level control. The evaluated systems will extend to memory-augmented end-to-end policies, which test whether the weak memory of the VLAs comes from their architecture, and to hybrid systems that pair a language-model planner with a learned executor, which test whether combining the memory of the best agent with the practiced motions of the VLAs closes part of the gap. Repeated runs of the same instances will show how much results vary, together with the time and cost of each evaluation, and the difficulty levels of tasks on which the levels currently make no difference will be recalibrated.

\paragraph{From diagnosis to better models.} The diagnosis will move from explaining failures to guiding improvement. It will cover every evaluated system rather than three representative ones, and its rules and prompts will be released. The current harness reports only the first unrecoverable error of each episode, which shows where an episode breaks but not what happens after it. The diagnosis will therefore trace the whole course of a failed episode and record multiple breakpoints, including how an early error leads to later ones and whether the system notices a mistake and recovers from it. Episodes will also start from verified intermediate states of the scripted demonstrations, so that each stage of a workflow is measured even when an earlier stage fails. A developer can then change a model in response to its profile card and evaluate the new model on the same instances to check whether the change helped.

\paragraph{Transfer to real robots.} The last direction tests whether the findings hold outside simulation. Mind2Act World will add real-robot experiments in which its tasks are mapped to physical counterparts, starting with Mind2Act tasks such as CP02 and CP04 and extending to a wider set of tasks, so that the Mind, Act and Mind2Act results can be verified on physical hardware. The same tasks will also be paired with similar tasks in other simulators, which shows how much of the failure analysis carries across platforms.

\FloatBarrier

\section{Conclusion}
We introduce \textbf{Mind2Act World}, a benchmark that measures what a robot retains and decides, how well it executes, and whether the two hold together when each physical outcome sets the next decision. Thirty tasks in three views, each with three difficulty levels and paired \emph{Clean} and \emph{Random} scenes, yield 180 configurations, all scored from physical outcomes by the same evaluator for agents and end-to-end policies.

Our evaluation of three agents, four VLAs, and two WAMs over 56,700 episodes yields three main findings. First, the strongest agent leads every view without task-specific training, operating under a paused clock and with a motion planner, while the four fine-tuned VLAs have similar Overall success rates and differ mainly in the tasks they handle best. Second, single-view ability does not carry over to Mind2Act tasks: the best system falls from 52.7\% on Mind and 39.0\% on Act to 12.0\% on Mind2Act, and no end-to-end policy exceeds 3.6\%. Third, the VLAs are limited even before composition, by memory that mostly disappears when the scene changes and by the first physical contact of the longer workflows. Eight of the thirty tasks are completed by no system.

Wang Yangming held that knowing and acting are one. In current embodied systems, they remain separate abilities, unevenly developed and not yet joined. Mind2Act World makes the distance between them measurable. The interventions, references and additional systems described in Section~\ref{sec:limitations} are our next steps toward explaining it.

\section*{Acknowledgments}
We thank Chang Wu and Jiaen Huang for the additional manual review in our evaluation loop, and Yuankun Lv, Zixu Lin, and Yegang Tang for their work on optimizing the project website. We are grateful to Velabot AI, Simple AI, and Baidu AI Cloud for providing computing resources, covering API token costs, and optimizing our evaluation infrastructure.

\clearpage
\renewcommand{\refname}{References}

\clearpage
\appendix
\addtocontents{toc}{\protect\setcounter{tocdepth}{0}}
\section*{Appendix Overview}
\addcontentsline{toc}{section}{Appendix Overview}
The appendices give the material needed to rebuild and use the benchmark. Appendices~\ref{app:platform}--\ref{app:baselines} describe the platform, protocol and systems, Appendix~\ref{app:full-results} reports results by task and setting with confidence intervals, Appendix~\ref{app:diagnostics-detail} gives the diagnostic details, and Appendix~\ref{app:tasks} is the reference for the thirty tasks.
\begin{itemize}[leftmargin=*,itemsep=1pt,topsep=2pt]
\item \textbf{Appendix~\ref{app:platform}, Simulation platform and data:} the simulator, assets and parallel evaluation, how demonstrations are collected and qualified, and how \emph{Clean} and \emph{Random} test instances are generated.
\item \textbf{Appendix~\ref{app:impl}, Evaluation protocol:} what a system observes and how it acts, episode horizons and budgets, how the task score is computed, and how results are selected and kept comparable.
\item \textbf{Appendix~\ref{app:baselines}, Systems and training:} the initialization, global batch size and training steps of each evaluated end-to-end policy, and the API setting of the agents.
\item \textbf{Appendix~\ref{app:full-results}, Detailed results:} outcomes by setting, success counts with confidence intervals, scores and success rates for every task, and successes by difficulty level.
\item \textbf{Appendix~\ref{app:diagnostics-detail}, Diagnostic details:} how a failed episode is assigned to a stage and a code, task-level results by view, the share of failures settled by each layer, the model profile cards of three representative systems, the stop reasons of the agent runs, and the behavioral profiles.
\item \textbf{Appendix~\ref{app:tasks}, Task details:} the real-world needs behind the tasks, a summary table and the task cards.
\end{itemize}

\section{Simulation Platform and Data Details}
\label{app:platform}
This appendix complements Section~\ref{subsec:simulation-platform} with the details needed to rebuild the task scenes and to regenerate the demonstration data.

\subsection{Simulation Platform}
Mind2Act World runs in Isaac Sim with two ARX X5 arms and parallel grippers. A fixed head camera and two wrist cameras provide RGB images. Physics is stepped at 120\,Hz and robot targets are updated at 30\,Hz. Simulation pauses while a model computes, so task deadlines are measured in simulation time and cost in wall-clock time. Each task is defined by a specification that fixes the scene, the objects and mechanisms, the per-level parameters, the episode randomization and the evaluator. Changing the level or the seed changes only these parameters, never the objective or the scoring rule.

\subsection{Assets and Parallel Evaluation}
As shown in Fig.~\ref{fig:assets-parallel}, task scenes combine shared robot and workspace assets with task-specific objects, instruments, and mechanisms. Their geometry, motion, and contact properties are configured so that every required effect is physically realized: a button travels and releases, a dial retains its reading after release, cloth stays folded only if it is laid flat, and a balance beam deflects under the actual load. Physical completion checks compare simulator states and events with task requirements, including release and support after placement.

Episodes run in isolated simulator processes. A scheduler distributes task-level conditions across the available GPUs and checks the first result of each condition before launching its remaining seeds. A valid policy failure lets evaluation continue, whereas a runtime failure blocks the remaining seeds for inspection. Model services may persist across episodes, and requests from concurrent episodes are kept apart, so that each episode observes its own visual-action sequence.

\subsection{Demonstration Collection}
Demonstrations are collected under one protocol for every task and level: 50 episodes per task and level for all thirty tasks (4,500 in total), all recorded in \emph{Clean} scenes. Each episode records the instruction, three RGB streams at 30\,Hz, and the joint states, with joint targets logged at 120\,Hz. A demonstration qualifies only if the evaluator marks it successful. For timing-critical tasks it must also keep a margin, for example at least 1\,s between withdrawal and gate closure in DR3. Experts are scripted: task scripts set the operation sequence and end-effector targets, and inverse kinematics and motion planning produce the joint trajectories (Section~\ref{subsec:data-collection}). Recording checks identify missing or corrupted frames, invalid state or action values, misaligned timestamps, and duplicate recordings. Together with the physical completion checks, they ensure that demonstrations show the required behavior and yield consistent training records.

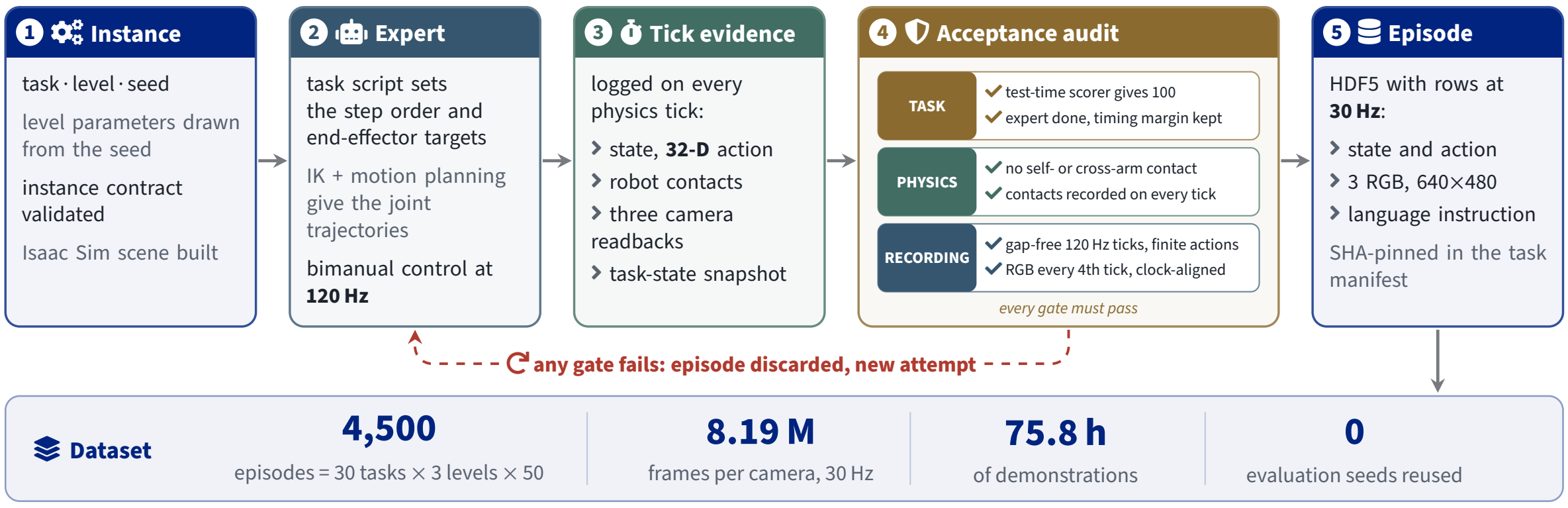
\begin{figure}[!htb]\centering
\resizebox{\linewidth}{!}{\begingroup
\sffamily
\definecolor{DPRed}{HTML}{B03A2E}\definecolor{DPInk}{HTML}{1D232C}\definecolor{DPMuted}{HTML}{5A6270}
\def\dpH{3.95}
\newcommand\dphead[6]{%
  \fill[#3!5,rounded corners=5pt] (#1,0) rectangle ++(#2,-\dpH);
  \fill[#3,rounded corners=5pt] (#1,0) rectangle ++(#2,-0.62);
  \fill[#3] (#1,-0.34) rectangle ++(#2,-0.28);
  \draw[#3!80,line width=.8pt,rounded corners=5pt] (#1,0) rectangle ++(#2,-\dpH);
  \node[circle,fill=white,text=#3,inner sep=0pt,minimum size=.34cm,font=\scriptsize\bfseries] at (#1+.30,-.31) {#5};
  \node[anchor=west,text=white,font=\footnotesize\bfseries,inner sep=0pt] at (#1+.56,-.31) {#4\hspace{.3em}#6};
}
\newcommand\dpcard[7]{%
  \dphead{#1}{#2}{#3}{#4}{#5}{#6}
  \node[anchor=north west,text width=#2 cm-.40cm,align=left,inner sep=0pt,text=DPInk,
    font=\scriptsize] at (#1+.20,-.84) {\hyphenpenalty=10000\exhyphenpenalty=10000 #7};
}
\newcommand\dpgate[5]{%
  \fill[white,rounded corners=3pt] (10.80,#1) rectangle ++(4.72,-.84);
  \draw[#2!70,line width=.5pt,rounded corners=3pt] (10.80,#1) rectangle ++(4.72,-.84);
  \fill[#2,rounded corners=3pt] (10.80,#1) rectangle ++(1.22,-.84);
  \node[text=white,font=\tiny\bfseries,inner sep=0pt] at (11.41,#1-.42) {#3};
  \node[anchor=west,font=\tiny,text=DPInk,align=left,inner sep=0pt] at (12.12,#1-.42)
    {\textcolor{#2}{\faCheck}~#4\\[2pt]\textcolor{#2}{\faCheck}~#5};
}
\newcommand\dpi{\textcolor{DPMuted}{\faAngleRight}~}
\begin{tikzpicture}[>={Stealth[length=5pt,width=5pt]}]
\dpcard{0}{3.10}{MMLabBlue}{\faCogs}{1}{Instance}{%
  task\,\textperiodcentered\,level\,\textperiodcentered\,seed\\[4pt]
  \textcolor{DPMuted}{level parameters drawn from the seed}\\[4pt]
  instance contract validated\\[4pt]
  \textcolor{DPMuted}{Isaac Sim scene built}}
\dpcard{3.52}{3.10}{M2AAct}{\faRobot}{2}{Expert}{%
  task script sets the step order and end-effector targets\\[4pt]
  \textcolor{DPMuted}{IK + motion planning give the joint trajectories}\\[4pt]
  bimanual control at \textbf{120\,Hz}}
\dpcard{7.04}{3.10}{M2AMind}{\faStopwatch}{3}{Tick evidence}{%
  logged on every physics tick:\\[4pt]
  \dpi state, \textbf{32-D} action\\[2pt]
  \dpi robot contacts\\[2pt]
  \dpi three camera readbacks\\[2pt]
  \dpi task-state snapshot}
\dphead{10.56}{5.20}{M2ACoord}{\faShield*}{4}{Acceptance audit}
\dpgate{-0.80}{M2ACoord}{TASK}{test-time scorer gives 100}{expert done, timing margin kept}
\dpgate{-1.74}{M2AMind}{PHYSICS}{no self- or cross-arm contact}{contacts recorded on every tick}
\dpgate{-2.68}{M2AAct}{RECORDING}{gap-free 120\,Hz ticks, finite actions}{RGB every 4th tick, clock-aligned}
\node[font=\tiny\itshape,text=M2ACoord,inner sep=0pt] at (13.16,-3.74) {every gate must pass};
\dpcard{16.18}{3.10}{MMLabBlue}{\faDatabase}{5}{Episode}{%
  HDF5 with rows at \textbf{30\,Hz}:\\[4pt]
  \dpi state and action\\[2pt]
  \dpi 3 RGB, 640$\times$480\\[2pt]
  \dpi language instruction\\[4pt]
  \textcolor{DPMuted}{SHA-pinned in the task manifest}}
\foreach \a/\b in {3.10/3.52,6.62/7.04,10.14/10.56,15.76/16.18}
  \draw[->,line width=1.1pt,DPInk!60] (\a+.03,-1.90) -- (\b-.03,-1.90);
\draw[->,DPRed,line width=.9pt,dashed,rounded corners=6pt]
  (13.16,-\dpH-.04) -- (13.16,-4.41) -- (5.07,-4.41) -- (5.07,-\dpH-.04);
\node[fill=white,inner sep=2.5pt,font=\scriptsize\bfseries,text=DPRed] at (9.11,-4.41)
  {\faRedo~any gate fails: episode discarded, new attempt};
\draw[->,line width=1.1pt,DPInk!60] (17.73,-\dpH-.04) -- (17.73,-4.77);
\fill[MMLabBlue!6,rounded corners=5pt] (0,-4.81) rectangle (19.28,-6.11);
\draw[MMLabBlue!45,line width=.6pt,rounded corners=5pt] (0,-4.81) rectangle (19.28,-6.11);
\node[anchor=west,font=\footnotesize\bfseries,text=MMLabBlue,inner sep=0pt] at (.35,-5.46) {\faLayerGroup\hspace{.4em}Dataset};
\foreach \x/\num/\lab in {
  4.75/{4,500}/{episodes = 30 tasks $\times$ 3 levels $\times$ 50},
  9.35/{8.19\,M}/{frames per camera, 30\,Hz},
  13.00/{75.8\,h}/{of demonstrations},
  16.70/{0}/{evaluation seeds reused}}{
  \node[font=\Large\bfseries,text=MMLabBlue,inner sep=0pt] at (\x,-5.25) {\num};
  \node[font=\scriptsize,text=DPMuted,inner sep=0pt] at (\x,-5.78) {\lab};
}
\foreach \x in {7.20,11.20,14.85} \draw[MMLabBlue!25,line width=.5pt] (\x,-5.01) -- (\x,-5.91);
\end{tikzpicture}
\endgroup}
\caption{\textbf{Demonstration collection and quality control.} A scripted expert executes each task instance at 120\,Hz while the robot state, contacts and camera readbacks are logged on every physics tick. An episode is kept only if the test-time scorer gives it full score, timing-critical tasks keep the required margin, the arms make no self- or cross-arm contact, and the recording is gap-free and clock-aligned. Otherwise it is discarded and collected again. Kept episodes are stored as 30\,Hz HDF5 files.}
\label{fig:demo-pipeline}
\end{figure}

\begin{table}[t]\centering\small
\caption{\textbf{Training dataset statistics by benchmark view and difficulty level.} Statistics cover all 30 tasks, with 50 episodes per task at each difficulty level. Frames count synchronized samples per camera; durations use the recorded sampling frequency of 30\,Hz.}
\label{tab:training-data-summary}
\begingroup
\setlength{\tabcolsep}{3pt}
\renewcommand{\arraystretch}{1.15}
\begin{tabularx}{\linewidth}{@{}l *{5}{>{\centering\arraybackslash}X}@{}}\toprule
\rowcolor{M2ASage}\textit{Difficulty} & \textit{Frames} & \textit{\shortstack{Duration\\(h)}} & \textit{Episodes} & \textit{\shortstack{Avg. frames\\/ episode}} & \textit{\shortstack{Avg. duration\\/ episode (s)}}\\\midrule
\rowcolor{M2AMindT}\multicolumn{6}{@{}l}{\strut\textbf{\textit{\textcolor{M2AMind}{Mind}}}}\\
Easy & 549,776 & 5.09 & 500 & 1,099.55 & 36.65\\
Medium & 868,075 & 8.04 & 500 & 1,736.15 & 57.87\\
Hard & 1,216,463 & 11.26 & 500 & 2,432.93 & 81.10\\
\textit{Total} & 2,634,314 & 24.39 & 1,500 & 1,756.21 & 58.54\\
\rowcolor{M2AActT}\multicolumn{6}{@{}l}{\strut\textbf{\textit{\textcolor{M2AAct}{Act}}}}\\
Easy & 280,443 & 2.60 & 500 & 560.89 & 18.70\\
Medium & 356,496 & 3.30 & 500 & 712.99 & 23.77\\
Hard & 436,343 & 4.04 & 500 & 872.69 & 29.09\\
\textit{Total} & 1,073,282 & 9.94 & 1,500 & 715.52 & 23.85\\
\rowcolor{M2ACoordT}\multicolumn{6}{@{}l}{\strut\textbf{\textit{\textcolor{M2ACoord}{Mind2Act}}}}\\
Easy & 1,067,955 & 9.89 & 500 & 2,135.91 & 71.20\\
Medium & 1,502,661 & 13.91 & 500 & 3,005.32 & 100.18\\
Hard & 1,913,316 & 17.72 & 500 & 3,826.63 & 127.55\\
\textit{Total} & 4,483,932 & 41.52 & 1,500 & 2,989.29 & 99.64\\
\midrule
\rowcolor{M2AGroup}\textbf{All views} & \textbf{8,191,528} & \textbf{75.85} & \textbf{4,500} & \textbf{1,820.34} & \textbf{60.68}\\\bottomrule
\end{tabularx}
\endgroup
\end{table}

\subsection{Instance Generation: Clean and Random Settings}
\label{app:clean-random}
Every task and level has a frozen set of 50 evaluation slots, and each slot can be run in two settings that share the task instance and differ only in the scene around it. Each setting counts as a separate configuration, so the thirty tasks, three levels and two settings give the 180 configurations of the benchmark. Evaluation seeds are not used for training.

\paragraph{Clean.} The slot's task seed draws the task-relevant parameters described in Section~\ref{subsec:instance-variation}: object layouts, initial and target states, event orders and hidden parameters such as the PR5 stamp angle. Appearance is fixed to the default scene, with its table and floor materials and lighting, and no unrelated objects are present.

\paragraph{Random.} The same slot keeps its task seed, so objects, targets and events are identical to the clean instance. A separate environment seed adds three kinds of scene perturbation. (i)~Clutter objects are drawn from a shared library and placed on free table space, and their poses are recorded after they settle under physics. (ii)~Table and floor materials are drawn from a texture library. (iii)~The light type is drawn from six (dome, rectangular, distant, sphere, cylinder, disk), its intensity is scaled by 0.8--1.2, and each RGB channel is scaled by 0--2. The amount of clutter is set per task and, for most tasks, does not change with the difficulty level. Clutter is placed only on free table space, never on a conveyor belt. The Random scene of a slot is fixed, so systems run on the same slot start from the same scene.

Fig.~\ref{fig:clean-random} gives examples of the two settings. All nine evaluated systems cover all thirty tasks in both settings. Reproducing a result requires the setting as well as the instance identifier.

\FloatBarrier
\section{Evaluation Protocol Details}
\label{app:impl}
\label{app:repro}
With the scenes and test instances fixed by Appendix~\ref{app:platform}, this appendix specifies how a system interacts with them, expanding Section~\ref{sec:protocol}: what it observes and how it acts, how long an episode may run, how the episode is scored, and how results are selected.

\subsection{Observations, Actions and Hidden State}
A policy receives the task instruction, the selected RGB cameras with arm joint positions and the declared gripper state, and earlier public observations and permitted feedback. Hidden task state is excluded, including unannounced targets and tolerances such as the PR5 angle, the DR3 gate timing and the DR4 drift radius. The evaluator uses it only to check contact, support, timing and completion.

End-to-end policies predict joint or end-effector actions. Joint predictions are denormalized and relative offsets converted to absolute targets. Cartesian poses are transformed into the controller frame and mapped to joint targets by inverse kinematics. Each policy executes its configured chunk before predicting again. Agents instead select Cartesian targets through motion tools, which a local inverse-kinematics and motion-planning service turns into trajectories using declared robot and fixed-scene geometry. Depth and pixel-to-world queries are disabled, and agents may retrieve earlier public images without advancing the simulation. The planner, history access and any adapter fallback are therefore part of the evaluated system (Table~\ref{tab:interface}).

\begin{table}[!htb]\centering\small
\caption{\textbf{System interfaces.} What each system family receives and returns under the shared protocol.}
\label{tab:interface}
\setlength{\tabcolsep}{4pt}
\begin{tabularx}{\linewidth}{@{}l YYY@{}}\toprule
Family & Input & Output & Must disclose\\\midrule
VLA / WAM & Instruction, 1--3 RGB views, joint state & Joint or EEF action chunk & Checkpoint, chunk, control period, data\\
Agent & Instruction, RGB, history, tool feedback & Motion-tool calls & Model, harness, tools, budget, latency\\
Hybrid & As above, split across modules & Planner calls and controller actions & Module boundaries, fallbacks, cost\\
\midrule
\multicolumn{4}{@{}l}{\footnotesize\textit{Not available to any system: hidden targets and tolerances, gate timing, drift radii, depth and pixel-to-world queries.}}\\
\bottomrule\end{tabularx}
\end{table}

\subsection{Horizons and Budgets}
The end-to-end task horizon uses the 90th percentile of demonstration duration at each level, multiplied by 1.2 and rounded up to a control step. \emph{Clean} and \emph{Random} share it. Agents use this physics budget multiplied by four and at most 150 decisions per episode (Table~\ref{tab:horizons}). Rejected and invalid actions count as decisions. Network retries do not, with up to 12 retries per decision. An agent episode ends when the task is complete, when the agent stops, or when either budget is used. Wall-clock limits are safety stops. An episode stopped by one is scored like any other unfinished episode.
\begin{table}[!htbp]\centering\small
\caption{\textbf{Run budgets.} $B$ denotes the task- and level-specific end-to-end budget in physics steps.}
\label{tab:horizons}
\begin{tabular}{@{}lrrl@{}}\toprule
Leaderboard & Physics steps & Decisions & Wall-clock safety limit\\\midrule
End-to-end & $B$ & unlimited & 2 hours\\
Agent & $4B$ & 150 & 6 hours\\\bottomrule
\end{tabular}
\end{table}

\subsection{Task Score}
\label{app:task-score}
The task evaluator produces a score between 0 and 100 using the task-specific rule shown on the task pages of the project website. A successful episode scores 100. Any other outcome receives the partial credit defined for that task, or zero. For example, CP01 gives credit for each requested item released into the correct tray and remaining supported, and full success additionally requires every order to be complete and the meal trays to be retrieved. Partial-credit scales differ between tasks, so scores are compared across systems on one task and averaged within a view, but not compared across tasks.

For each task, the evaluator marks an episode as successful when it meets all completion criteria, satisfies any prerequisites, and violates no rule that prevents success. We calculate each system’s success rate by dividing the number of episodes marked as successful by the total number of evaluated episodes.

\subsection{Evaluation Integrity and Result Selection}
Every system on a leaderboard receives the same frozen list of task, level and seed instances. Reset restores the robot, objects, task stage, evaluator and random state, and clears policy memory. Each episode writes a manifest that binds the task, level, seed and configuration hashes to the observation profile, action interface, model or checkpoint version, clocks, budgets, termination reason and scorer output, together with an event record that separates the issued action from the evaluator's completion check. All evaluations in this report were run by the benchmark team on its own machines.

\paragraph{Retained attempts.} 
Each system receives one result per test instance and scene setting. Infrastructure failures can trigger a retry. A valid run in which the policy fails the task still counts as a failure. Earlier attempts may remain in the logs for cost tracking, but each retained result counts only once.

\FloatBarrier
\section{Systems and Training Details}
\label{app:baselines}
For reproducibility, this appendix reports the initialization, global batch size and number of training steps of every evaluated end-to-end policy, the four VLAs and the two WAMs, each fine-tuned on the 4,500 demonstrations of Mind2Act World (Section~\ref{subsec:data-statistics}). The three agents, GPT-6 Astra, DeepSeek-V4.1-Flash and Gemini-4-Argon, require no training: each is called zero-shot through its API with medium reasoning effort.

\textbf{G0.5.} G0.5~\cite{galaxea2026g05} is initialized from its public pretrained checkpoint and fine-tuned with a global batch size of 240 for 136,526 steps.

\textbf{$\pi_{0.5}$.} $\pi_{0.5}$~\cite{pi2025pi05} is initialized from its public pretrained checkpoint and fine-tuned with a global batch size of 256 for 264,305 steps.

\textbf{KinRT.} KinRT~\cite{kinrt2026} is initialized from its public pretrained checkpoint and fine-tuned with a global batch size of 256 for 264,305 steps.

\textbf{X-VLA.} X-VLA~\cite{zheng2025xvla} is initialized from its public pretrained checkpoint and fine-tuned with a global batch size of 2,048 for 41,747 steps.

\textbf{InternW0.} The world action model InternW0~\cite{cai2026internw0} is initialized from its public pretrained checkpoint and fine-tuned with a global batch size of 2,048 for 32,000 steps.

\textbf{FastWAM.} The world action model FastWAM~\cite{yuan2026fastwam} is initialized from its public pretrained checkpoint and fine-tuned with a global batch size of 512 for 79,210 steps.

\FloatBarrier
\section{Detailed Results}
\label{app:full-results}\label{subsec:resource-coverage}\label{subsec:result-coverage}
This appendix gives the numbers behind Section~\ref{sec:results}. Table~\ref{tab:compute} reports Overall scores and success counts in each setting, Table~\ref{tab:view-ci} the success counts of each view with confidence intervals, Table~\ref{tab:per-task-appendix} the task scores and setting-level success rates of every task, and Table~\ref{tab:difficulty-counts} the successes of two end-to-end policies by difficulty level. Evaluation time and API cost are reported in Table~\ref{tab:efficiency-main}.
\begin{table}[!htbp]\centering\small
\caption{\textbf{Outcomes by setting.} Overall scores ($0.25\,$\VMind$\,+\,0.25\,$\VAct$\,+\,0.50\,$\VMtA) and numbers of successful episodes in each setting.}
\label{tab:compute}
\label{tab:resource-use}
\begin{tabular}{@{}lrrrr@{}}\toprule
System & Clean Overall & Clean successes & Random Overall & Random successes \\\midrule
GPT-6 Astra & 31.8 & 165/450 & 28.7 & 146/450 \\
DeepSeek-V4.1-Flash & 5.8 & 28/450 & 6.4 & 31/450 \\
Gemini-4-Argon & 4.8 & 24/450 & 3.9 & 18/450 \\
KinRT & 9.1 & 364/4,500 & 6.8 & 264/4,500 \\
$\pi_{0.5}$ & 8.9 & 340/4,500 & 6.3 & 234/4,500 \\
G0.5 & 6.7 & 288/4,500 & 5.8 & 242/4,500 \\
X-VLA & 6.8 & 344/4,500 & 5.6 & 282/4,500 \\
InternW0 & 9.2 & 407/4,500 & 8.2 & 373/4,500 \\
FastWAM & 4.5 & 198/4,500 & 1.7 & 67/4,500 \\
\bottomrule\end{tabular}
\end{table}

\begin{table}[!htbp]\centering\small
\caption{\textbf{Success counts and 95\% confidence intervals.} Each cell gives the number of successful episodes and, in brackets, the Wilson score interval of the success rate in percent. Intervals treat episodes as independent draws and therefore describe sampling over the fixed instance lists; they do not include variation between tasks or between repeated runs.}
\label{tab:view-ci}
\setlength{\tabcolsep}{5pt}
\begin{tabularx}{\linewidth}{@{}l*{4}{>{\centering\arraybackslash}X}@{}}\toprule
\rowcolor{M2ASage}System & Mind & Act & Mind2Act & All episodes\\\midrule
\mtgroup{5}{Agents (300 episodes per view)} \\
GPT-6 Astra & 158 \;[47.0, 58.2] & 117 \;[33.7, 44.6] & 36 \;[8.8, 16.2] & 311 \;[31.5, 37.7] \\
DeepSeek-V4.1-Flash & 33 \;[7.9, 15.0] & 26 \;[6.0, 12.4] & 0 \;[0.0, 1.3] & 59 \;[5.1, 8.4] \\
Gemini-4-Argon & 13 \;[2.5, 7.3] & 29 \;[6.8, 13.5] & 0 \;[0.0, 1.3] & 42 \;[3.5, 6.2] \\
\mtgroup{5}{End-to-end policies (3,000 episodes per view)} \\
KinRT & 132 \;[3.7, 5.2] & 408 \;[12.4, 14.9] & 88 \;[2.4, 3.6] & 628 \;[6.5, 7.5] \\
$\pi_{0.5}$ & 105 \;[2.9, 4.2] & 360 \;[10.9, 13.2] & 109 \;[3.0, 4.4] & 574 \;[5.9, 6.9] \\
G0.5 & 58 \;[1.5, 2.5] & 403 \;[12.3, 14.7] & 69 \;[1.8, 2.9] & 530 \;[5.4, 6.4] \\
X-VLA & 31 \;[0.7, 1.5] & 589 \;[18.3, 21.1] & 6 \;[0.1, 0.4] & 626 \;[6.4, 7.5] \\
InternW0 & 161 \;[4.6, 6.2] & 528 \;[16.3, 19.0] & 91 \;[2.5, 3.7] & 780 \;[8.1, 9.3] \\
FastWAM & 27 \;[0.6, 1.3] & 222 \;[6.5, 8.4] & 16 \;[0.3, 0.9] & 265 \;[2.6, 3.3] \\
\bottomrule\end{tabularx}
\end{table}

\begin{table}[!htbp]
\centering
\caption{\textbf{Task scores and setting-level success rates.} For each task and system, the first number is the mean task score (0--100) over both settings and the gray pair the success rate (\%) in the \emph{Clean}\,/\,\emph{Random} setting. Each setting has 15 episodes per task for an agent and 150 for an end-to-end policy.}
\label{tab:per-task-appendix}
\begingroup
\fontsize{7.2}{8.6}\selectfont
\setlength{\tabcolsep}{1.5pt}
\renewcommand{\arraystretch}{0.96}
\begin{tabularx}{\linewidth}{@{}l*{3}{>{\centering\arraybackslash}X}!{\color{black!30}\vrule}*{6}{>{\centering\arraybackslash}X}@{}}
\toprule
\textbf{Task} & GPT-6 & DeepSeek & Gemini & KinRT & $\pi_{0.5}$ & G0.5 & X-VLA & InternW0 & FastWAM \\
\midrule
\rowcolor{M2AMindT}\multicolumn{10}{@{}l}{\strut\textbf{\textit{\textcolor{M2AMind}{Mind}}}}\\
E1 & 76.9\enspace\textcolor{black!55}{80\,/\,67} & 22.9\enspace\textcolor{black!55}{13\,/\,27} & 7.0\enspace\textcolor{black!55}{7\,/\,7} & 11.9\enspace\textcolor{black!55}{7\,/\,7} & 3.7\enspace\textcolor{black!55}{1\,/\,1} & 0.4\enspace\textcolor{black!55}{0\,/\,0} & 0.2\enspace\textcolor{black!55}{0\,/\,0} & 0.0\enspace\textcolor{black!55}{0\,/\,0} & 2.7\enspace\textcolor{black!55}{0\,/\,2} \\
E2 & 64.7\enspace\textcolor{black!55}{60\,/\,67} & 24.3\enspace\textcolor{black!55}{20\,/\,20} & 10.0\enspace\textcolor{black!55}{7\,/\,7} & 2.6\enspace\textcolor{black!55}{1\,/\,1} & 2.2\enspace\textcolor{black!55}{0\,/\,0} & 2.9\enspace\textcolor{black!55}{0\,/\,0} & 3.0\enspace\textcolor{black!55}{0\,/\,2} & 2.6\enspace\textcolor{black!55}{0\,/\,1} & 0.9\enspace\textcolor{black!55}{1\,/\,0} \\
P1 & 46.8\enspace\textcolor{black!55}{53\,/\,33} & 0.4\enspace\textcolor{black!55}{0\,/\,0} & 1.5\enspace\textcolor{black!55}{0\,/\,0} & 1.0\enspace\textcolor{black!55}{1\,/\,1} & 1.3\enspace\textcolor{black!55}{1\,/\,1} & 0.1\enspace\textcolor{black!55}{0\,/\,0} & 0.0\enspace\textcolor{black!55}{0\,/\,0} & 0.8\enspace\textcolor{black!55}{1\,/\,1} & 0.4\enspace\textcolor{black!55}{1\,/\,0} \\
P2 & 0.3\enspace\textcolor{black!55}{0\,/\,0} & 0.0\enspace\textcolor{black!55}{0\,/\,0} & 0.0\enspace\textcolor{black!55}{0\,/\,0} & 17.8\enspace\textcolor{black!55}{30\,/\,1} & 17.2\enspace\textcolor{black!55}{27\,/\,1} & 4.6\enspace\textcolor{black!55}{3\,/\,1} & 0.0\enspace\textcolor{black!55}{0\,/\,0} & 3.1\enspace\textcolor{black!55}{3\,/\,0} & 0.0\enspace\textcolor{black!55}{0\,/\,0} \\
S1 & 23.8\enspace\textcolor{black!55}{33\,/\,13} & 0.0\enspace\textcolor{black!55}{0\,/\,0} & 0.2\enspace\textcolor{black!55}{0\,/\,0} & 0.4\enspace\textcolor{black!55}{0\,/\,0} & 0.3\enspace\textcolor{black!55}{0\,/\,0} & 0.2\enspace\textcolor{black!55}{0\,/\,0} & 0.8\enspace\textcolor{black!55}{1\,/\,0} & 1.4\enspace\textcolor{black!55}{1\,/\,0} & 0.0\enspace\textcolor{black!55}{0\,/\,0} \\
S2 & 100.0\enspace\textcolor{black!55}{100\,/\,100} & 20.0\enspace\textcolor{black!55}{13\,/\,27} & 10.0\enspace\textcolor{black!55}{13\,/\,7} & 4.7\enspace\textcolor{black!55}{5\,/\,5} & 7.3\enspace\textcolor{black!55}{9\,/\,5} & 1.7\enspace\textcolor{black!55}{3\,/\,1} & 0.0\enspace\textcolor{black!55}{0\,/\,0} & 10.0\enspace\textcolor{black!55}{11\,/\,9} & 0.0\enspace\textcolor{black!55}{0\,/\,0} \\
F1 & 78.9\enspace\textcolor{black!55}{67\,/\,87} & 22.4\enspace\textcolor{black!55}{13\,/\,20} & 7.2\enspace\textcolor{black!55}{13\,/\,0} & 3.2\enspace\textcolor{black!55}{3\,/\,3} & 5.8\enspace\textcolor{black!55}{8\,/\,3} & 8.6\enspace\textcolor{black!55}{12\,/\,5} & 5.0\enspace\textcolor{black!55}{7\,/\,3} & 9.7\enspace\textcolor{black!55}{7\,/\,11} & 4.1\enspace\textcolor{black!55}{8\,/\,0} \\
F2 & 53.3\enspace\textcolor{black!55}{67\,/\,40} & 0.0\enspace\textcolor{black!55}{0\,/\,0} & 3.3\enspace\textcolor{black!55}{7\,/\,0} & 5.7\enspace\textcolor{black!55}{5\,/\,7} & 3.7\enspace\textcolor{black!55}{4\,/\,3} & 0.7\enspace\textcolor{black!55}{0\,/\,1} & 0.0\enspace\textcolor{black!55}{0\,/\,0} & 13.7\enspace\textcolor{black!55}{10\,/\,17} & 0.0\enspace\textcolor{black!55}{0\,/\,0} \\
M1 & 0.0\enspace\textcolor{black!55}{0\,/\,0} & 0.0\enspace\textcolor{black!55}{0\,/\,0} & 0.0\enspace\textcolor{black!55}{0\,/\,0} & 0.1\enspace\textcolor{black!55}{0\,/\,0} & 0.0\enspace\textcolor{black!55}{0\,/\,0} & 0.0\enspace\textcolor{black!55}{0\,/\,0} & 0.0\enspace\textcolor{black!55}{0\,/\,0} & 0.3\enspace\textcolor{black!55}{0\,/\,0} & 0.0\enspace\textcolor{black!55}{0\,/\,0} \\
M2 & 94.0\enspace\textcolor{black!55}{87\,/\,100} & 33.8\enspace\textcolor{black!55}{33\,/\,33} & 10.6\enspace\textcolor{black!55}{7\,/\,13} & 9.1\enspace\textcolor{black!55}{14\,/\,1} & 3.2\enspace\textcolor{black!55}{4\,/\,0} & 8.8\enspace\textcolor{black!55}{10\,/\,3} & 5.0\enspace\textcolor{black!55}{8\,/\,0} & 19.6\enspace\textcolor{black!55}{19\,/\,17} & 4.0\enspace\textcolor{black!55}{6\,/\,1} \\
\rowcolor{M2AActT}\multicolumn{10}{@{}l}{\strut\textbf{\textit{\textcolor{M2AAct}{Act}}}}\\
DR1 & 79.4\enspace\textcolor{black!55}{87\,/\,67} & 10.0\enspace\textcolor{black!55}{7\,/\,7} & 4.1\enspace\textcolor{black!55}{0\,/\,0} & 25.9\enspace\textcolor{black!55}{25\,/\,13} & 20.9\enspace\textcolor{black!55}{15\,/\,11} & 31.5\enspace\textcolor{black!55}{25\,/\,24} & 0.9\enspace\textcolor{black!55}{0\,/\,0} & 29.5\enspace\textcolor{black!55}{17\,/\,29} & 10.4\enspace\textcolor{black!55}{3\,/\,1} \\
DR2 & 91.7\enspace\textcolor{black!55}{100\,/\,80} & 21.4\enspace\textcolor{black!55}{13\,/\,0} & 12.0\enspace\textcolor{black!55}{7\,/\,0} & 15.9\enspace\textcolor{black!55}{0\,/\,2} & 14.8\enspace\textcolor{black!55}{0\,/\,0} & 48.7\enspace\textcolor{black!55}{45\,/\,28} & 29.4\enspace\textcolor{black!55}{18\,/\,7} & 44.6\enspace\textcolor{black!55}{43\,/\,21} & 7.2\enspace\textcolor{black!55}{0\,/\,0} \\
DR3 & 80.0\enspace\textcolor{black!55}{100\,/\,60} & 26.7\enspace\textcolor{black!55}{27\,/\,27} & 40.0\enspace\textcolor{black!55}{33\,/\,47} & 79.5\enspace\textcolor{black!55}{85\,/\,73} & 78.7\enspace\textcolor{black!55}{81\,/\,76} & 51.0\enspace\textcolor{black!55}{51\,/\,51} & 75.7\enspace\textcolor{black!55}{79\,/\,72} & 72.7\enspace\textcolor{black!55}{69\,/\,76} & 61.4\enspace\textcolor{black!55}{84\,/\,39} \\
DR4 & 0.0\enspace\textcolor{black!55}{0\,/\,0} & 0.0\enspace\textcolor{black!55}{0\,/\,0} & 0.0\enspace\textcolor{black!55}{0\,/\,0} & 0.0\enspace\textcolor{black!55}{0\,/\,0} & 0.0\enspace\textcolor{black!55}{0\,/\,0} & 0.0\enspace\textcolor{black!55}{0\,/\,0} & 0.0\enspace\textcolor{black!55}{0\,/\,0} & 0.0\enspace\textcolor{black!55}{0\,/\,0} & 0.0\enspace\textcolor{black!55}{0\,/\,0} \\
PR1 & 100.0\enspace\textcolor{black!55}{100\,/\,100} & 42.5\enspace\textcolor{black!55}{40\,/\,40} & 48.0\enspace\textcolor{black!55}{40\,/\,40} & 30.4\enspace\textcolor{black!55}{20\,/\,17} & 28.8\enspace\textcolor{black!55}{17\,/\,15} & 5.7\enspace\textcolor{black!55}{4\,/\,2} & 46.1\enspace\textcolor{black!55}{45\,/\,41} & 45.7\enspace\textcolor{black!55}{43\,/\,37} & 13.0\enspace\textcolor{black!55}{21\,/\,0} \\
PR2 & 52.2\enspace\textcolor{black!55}{47\,/\,40} & 12.9\enspace\textcolor{black!55}{7\,/\,7} & 18.5\enspace\textcolor{black!55}{27\,/\,0} & 3.2\enspace\textcolor{black!55}{0\,/\,1} & 3.4\enspace\textcolor{black!55}{0\,/\,2} & 4.6\enspace\textcolor{black!55}{1\,/\,2} & 2.5\enspace\textcolor{black!55}{1\,/\,0} & 4.2\enspace\textcolor{black!55}{1\,/\,4} & 1.4\enspace\textcolor{black!55}{0\,/\,1} \\
PR3 & 0.0\enspace\textcolor{black!55}{0\,/\,0} & 0.0\enspace\textcolor{black!55}{0\,/\,0} & 0.0\enspace\textcolor{black!55}{0\,/\,0} & 0.1\enspace\textcolor{black!55}{0\,/\,0} & 1.1\enspace\textcolor{black!55}{2\,/\,0} & 0.2\enspace\textcolor{black!55}{0\,/\,0} & 0.3\enspace\textcolor{black!55}{0\,/\,0} & 0.2\enspace\textcolor{black!55}{0\,/\,0} & 0.1\enspace\textcolor{black!55}{0\,/\,0} \\
PR4 & 0.0\enspace\textcolor{black!55}{0\,/\,0} & 0.0\enspace\textcolor{black!55}{0\,/\,0} & 0.0\enspace\textcolor{black!55}{0\,/\,0} & 18.0\enspace\textcolor{black!55}{10\,/\,26} & 10.0\enspace\textcolor{black!55}{11\,/\,9} & 17.3\enspace\textcolor{black!55}{14\,/\,21} & 64.0\enspace\textcolor{black!55}{67\,/\,61} & 6.3\enspace\textcolor{black!55}{9\,/\,4} & 0.0\enspace\textcolor{black!55}{0\,/\,0} \\
PR5 & 1.0\enspace\textcolor{black!55}{0\,/\,0} & 0.0\enspace\textcolor{black!55}{0\,/\,0} & 0.0\enspace\textcolor{black!55}{0\,/\,0} & 0.9\enspace\textcolor{black!55}{0\,/\,0} & 1.1\enspace\textcolor{black!55}{0\,/\,0} & 0.7\enspace\textcolor{black!55}{0\,/\,0} & 0.1\enspace\textcolor{black!55}{0\,/\,0} & 0.1\enspace\textcolor{black!55}{0\,/\,0} & 0.0\enspace\textcolor{black!55}{0\,/\,0} \\
PR6 & 0.0\enspace\textcolor{black!55}{0\,/\,0} & 0.0\enspace\textcolor{black!55}{0\,/\,0} & 0.0\enspace\textcolor{black!55}{0\,/\,0} & 0.0\enspace\textcolor{black!55}{0\,/\,0} & 0.0\enspace\textcolor{black!55}{0\,/\,0} & 0.0\enspace\textcolor{black!55}{0\,/\,0} & 0.0\enspace\textcolor{black!55}{0\,/\,0} & 0.0\enspace\textcolor{black!55}{0\,/\,0} & 0.0\enspace\textcolor{black!55}{0\,/\,0} \\
\rowcolor{M2ACoordT}\multicolumn{10}{@{}l}{\strut\textbf{\textit{\textcolor{M2ACoord}{Mind2Act}}}}\\
CP01 & 61.2\enspace\textcolor{black!55}{60\,/\,60} & 1.0\enspace\textcolor{black!55}{0\,/\,0} & 0.3\enspace\textcolor{black!55}{0\,/\,0} & 10.9\enspace\textcolor{black!55}{7\,/\,9} & 11.8\enspace\textcolor{black!55}{10\,/\,7} & 10.5\enspace\textcolor{black!55}{6\,/\,10} & 2.6\enspace\textcolor{black!55}{2\,/\,1} & 3.1\enspace\textcolor{black!55}{3\,/\,1} & 5.4\enspace\textcolor{black!55}{8\,/\,1} \\
CP02 & 18.3\enspace\textcolor{black!55}{20\,/\,13} & 0.0\enspace\textcolor{black!55}{0\,/\,0} & 0.0\enspace\textcolor{black!55}{0\,/\,0} & 0.5\enspace\textcolor{black!55}{1\,/\,0} & 0.8\enspace\textcolor{black!55}{1\,/\,0} & 0.0\enspace\textcolor{black!55}{0\,/\,0} & 0.0\enspace\textcolor{black!55}{0\,/\,0} & 0.4\enspace\textcolor{black!55}{0\,/\,1} & 0.0\enspace\textcolor{black!55}{0\,/\,0} \\
CP03 & 0.0\enspace\textcolor{black!55}{0\,/\,0} & 0.0\enspace\textcolor{black!55}{0\,/\,0} & 0.0\enspace\textcolor{black!55}{0\,/\,0} & 0.8\enspace\textcolor{black!55}{0\,/\,0} & 1.2\enspace\textcolor{black!55}{0\,/\,0} & 0.2\enspace\textcolor{black!55}{0\,/\,0} & 0.0\enspace\textcolor{black!55}{0\,/\,0} & 0.3\enspace\textcolor{black!55}{0\,/\,0} & 0.0\enspace\textcolor{black!55}{0\,/\,0} \\
CP04 & 0.7\enspace\textcolor{black!55}{0\,/\,0} & 0.0\enspace\textcolor{black!55}{0\,/\,0} & 0.0\enspace\textcolor{black!55}{0\,/\,0} & 0.3\enspace\textcolor{black!55}{0\,/\,0} & 0.3\enspace\textcolor{black!55}{0\,/\,0} & 0.0\enspace\textcolor{black!55}{0\,/\,0} & 0.2\enspace\textcolor{black!55}{0\,/\,0} & 0.1\enspace\textcolor{black!55}{0\,/\,0} & 0.0\enspace\textcolor{black!55}{0\,/\,0} \\
CP05 & 24.4\enspace\textcolor{black!55}{13\,/\,27} & 1.2\enspace\textcolor{black!55}{0\,/\,0} & 0.0\enspace\textcolor{black!55}{0\,/\,0} & 0.8\enspace\textcolor{black!55}{0\,/\,0} & 1.2\enspace\textcolor{black!55}{0\,/\,0} & 0.1\enspace\textcolor{black!55}{0\,/\,0} & 0.3\enspace\textcolor{black!55}{0\,/\,0} & 0.8\enspace\textcolor{black!55}{0\,/\,0} & 0.0\enspace\textcolor{black!55}{0\,/\,0} \\
CP06 & 12.6\enspace\textcolor{black!55}{7\,/\,13} & 0.6\enspace\textcolor{black!55}{0\,/\,0} & 0.6\enspace\textcolor{black!55}{0\,/\,0} & 22.9\enspace\textcolor{black!55}{23\,/\,11} & 26.3\enspace\textcolor{black!55}{22\,/\,19} & 15.6\enspace\textcolor{black!55}{16\,/\,9} & 2.4\enspace\textcolor{black!55}{1\,/\,1} & 21.1\enspace\textcolor{black!55}{15\,/\,17} & 1.1\enspace\textcolor{black!55}{1\,/\,0} \\
CP07 & 0.0\enspace\textcolor{black!55}{0\,/\,0} & 0.0\enspace\textcolor{black!55}{0\,/\,0} & 0.0\enspace\textcolor{black!55}{0\,/\,0} & 0.0\enspace\textcolor{black!55}{0\,/\,0} & 0.0\enspace\textcolor{black!55}{0\,/\,0} & 0.0\enspace\textcolor{black!55}{0\,/\,0} & 0.0\enspace\textcolor{black!55}{0\,/\,0} & 0.0\enspace\textcolor{black!55}{0\,/\,0} & 0.0\enspace\textcolor{black!55}{0\,/\,0} \\
CP08 & 15.1\enspace\textcolor{black!55}{20\,/\,7} & 0.0\enspace\textcolor{black!55}{0\,/\,0} & 0.0\enspace\textcolor{black!55}{0\,/\,0} & 4.4\enspace\textcolor{black!55}{5\,/\,0} & 7.7\enspace\textcolor{black!55}{12\,/\,0} & 0.3\enspace\textcolor{black!55}{0\,/\,0} & 0.3\enspace\textcolor{black!55}{0\,/\,0} & 13.4\enspace\textcolor{black!55}{19\,/\,5} & 0.3\enspace\textcolor{black!55}{0\,/\,0} \\
CP09 & 0.0\enspace\textcolor{black!55}{0\,/\,0} & 0.0\enspace\textcolor{black!55}{0\,/\,0} & 0.0\enspace\textcolor{black!55}{0\,/\,0} & 0.0\enspace\textcolor{black!55}{0\,/\,0} & 0.0\enspace\textcolor{black!55}{0\,/\,0} & 0.0\enspace\textcolor{black!55}{0\,/\,0} & 0.0\enspace\textcolor{black!55}{0\,/\,0} & 0.0\enspace\textcolor{black!55}{0\,/\,0} & 0.0\enspace\textcolor{black!55}{0\,/\,0} \\
CP10 & 0.8\enspace\textcolor{black!55}{0\,/\,0} & 0.4\enspace\textcolor{black!55}{0\,/\,0} & 0.1\enspace\textcolor{black!55}{0\,/\,0} & 2.5\enspace\textcolor{black!55}{1\,/\,1} & 1.5\enspace\textcolor{black!55}{0\,/\,1} & 3.8\enspace\textcolor{black!55}{1\,/\,4} & 1.1\enspace\textcolor{black!55}{0\,/\,0} & 2.1\enspace\textcolor{black!55}{1\,/\,1} & 1.7\enspace\textcolor{black!55}{1\,/\,1} \\
\bottomrule
\end{tabularx}
\endgroup
\end{table}

\begin{table}[!htbp]\centering\small
\caption{\textbf{Successful episodes by difficulty level.} Counts out of 100 episodes per level (50 \emph{Clean} and 50 \emph{Random}) for two end-to-end policies, on the tasks where at least one of them succeeds repeatedly. The level totals equal the task counts of the release.}
\label{tab:difficulty-counts}
\setlength{\tabcolsep}{5pt}\setlength{\aboverulesep}{0pt}\setlength{\belowrulesep}{0pt}
\begin{tabularx}{\linewidth}{@{}l>{\raggedright\arraybackslash}X*{3}{r}>{\hspace{12pt}}r*{2}{r}@{}}\toprule
\rowcolor{M2ASage}\rule{0pt}{3.2ex} & & \multicolumn{3}{c}{$\pi_{0.5}$} & \multicolumn{3}{>{\hspace{12pt}}c@{}}{G0.5}\\
\cmidrule(lr){3-5}\cmidrule(l{16pt}){6-8}
\rowcolor{M2ASage}ID & Task & Easy & Med. & Hard & Easy & Med. & Hard\\\midrule\addlinespace[0.65ex]
DR1 & Conveyor pickup & 34 & 5 & 0 & 70 & 4 & 0 \\
DR2 & Light-cued response & 0 & 0 & 0 & 37 & 39 & 34 \\
DR3 & Timed passage through cyclic gates & 76 & 81 & 79 & 51 & 51 & 51 \\
PR1 & Resistance adjustment & 23 & 13 & 13 & 6 & 3 & 0 \\
PR4 & Cloth folding and flat placement & 28 & 1 & 1 & 51 & 0 & 1 \\
P2 & Steamer-stack order restoration & 22 & 18 & 3 & 5 & 1 & 0 \\
S2 & Hidden-hazard navigation & 20 & 2 & 0 & 5 & 0 & 0 \\
F1 & Unknown button mapping & 17 & 0 & 0 & 25 & 0 & 0 \\
M2 & Card-matching game & 6 & 0 & 0 & 19 & 1 & 0 \\
CP01 & Conveyor meal assembly & 20 & 5 & 1 & 24 & 0 & 0 \\
CP06 & Produce weighing on a balance & 46 & 16 & 0 & 30 & 5 & 2 \\
CP08 & Block structure from a shown blueprint & 17 & 1 & 0 & 0 & 0 & 0 \\
\addlinespace[0.4ex]\bottomrule\end{tabularx}
\end{table}

\FloatBarrier
\section{Diagnostic Details}
\label{app:diagnostics-detail}
This appendix defines the diagnostic steps and failure codes used in Section~\ref{sec:diagnostics}, gives the task-level results behind the per-view discussion, lists what each diagnosis layer settles for the three diagnosed systems, gives the model profile cards of three representative systems and the stop reasons of the agent episodes, and closes with the behavioral profiles.

\subsection{Diagnostic Steps and Failure Codes}
\label{subsec:error-taxonomy}
For each failed episode, we look for the first error from which the system could no longer recover. An error that the system later corrects does not count. We assign the failure to the earliest stage that explains the observed outcome. For example, choosing the wrong object to grasp is a reasoning error. Choosing the correct object but failing to hold it is an execution error.

We defined the failure codes by mapping the steps of each of the 30 tasks and identifying what could go wrong at each step. Table~\ref{tab:error-taxonomy} lists the resulting 21 codes. Eighteen describe failures in four stages: perceiving the scene and reasoning about the next action (Mind), executing the action (Act), and checking its physical outcome to decide what to do next (Mind-Action composition). The remaining three codes apply to the episode as a whole: the system exhausts its physics or decision budget, stalls or repeats an action, or leaves too little evidence to identify the failure. These codes are not assigned to a side. Each task uses only the codes relevant to its steps. The side of a failure (Mind, Act or composition) follows from its stage, whatever the view of the task. A misread cue in an Act task counts as a Mind failure. The assignment is a classification of the first unrecoverable error and is not a causal test.

\begin{table}[!htb]\centering\small
\definecolor{vMind}{HTML}{416D62}\definecolor{vAct}{HTML}{36566F}\definecolor{vM2A}{HTML}{8A6A2F}\definecolor{fGreyT}{HTML}{5A6270}%
\caption{\textbf{Failure codes used in episode diagnosis.} The table groups 18 codes by the stage at which the failure occurs. Three episode-level codes cover an exhausted budget, stalled or repeated actions, and insufficient evidence.}
\label{tab:error-taxonomy}
\setlength{\tabcolsep}{3pt}
\begin{tabularx}{\linewidth}{@{}lY@{\hspace{10pt}}c@{\hspace{4pt}}lY@{}}\toprule
\multicolumn{2}{@{}l}{\textbf{\textcolor{vMind}{Perceive}}\enspace\textcolor{fGreyT}{\itshape Mind}} && \multicolumn{2}{@{}l}{\textbf{\textcolor{vAct}{Execute}}\enspace\textcolor{fGreyT}{\itshape Act}}\\[1pt]
\texttt{\footnotesize M-PERCEIVE} & misrecognized an object, color or reading && \texttt{\footnotesize A-REACH} & approach or positioning failed\\
\addlinespace[3pt]\multicolumn{2}{@{}l}{\textbf{\textcolor{vMind}{Reason}}\enspace\textcolor{fGreyT}{\itshape Mind}} && \texttt{\footnotesize A-GRASP} & grasp failed or not held\\[1pt]
\texttt{\footnotesize M-GOAL} & misunderstood the goal of the task && \texttt{\footnotesize A-PLACE} & placement inaccurate or unstable\\
\texttt{\footnotesize M-MEMORY} & forgot or misremembered information && \texttt{\footnotesize A-PRESS} & press, turn or other interaction had no effect\\
\texttt{\footnotesize M-RULE} & inferred a wrong rule or mapping && \texttt{\footnotesize A-PRECISION} & error in continuous fine control\\
\texttt{\footnotesize M-SPATIAL} & wrong spatial mapping or layout && \texttt{\footnotesize A-TIMING} & action too early, too late or too slow\\
\texttt{\footnotesize M-PLAN} & wrong plan or step order && \texttt{\footnotesize A-COLLISION} & collision with or disturbance of the scene\\\cmidrule(l){4-5}
\texttt{\footnotesize M-PREDICT} & wrong prediction of how the scene changes && \multicolumn{2}{@{}l}{\textbf{\textcolor{fGreyT}{Episode level}}\enspace\textcolor{fGreyT}{\itshape not attributed}}\\[1pt]
\addlinespace[3pt]\multicolumn{2}{@{}l}{\textbf{\textcolor{vM2A}{Close the loop}}\enspace\textcolor{fGreyT}{\itshape composition}} && \texttt{\footnotesize X-TIMEOUT} & physics or decision budget ran out\\[1pt]
\texttt{\footnotesize C-NO-VERIFY} & moved on without confirming the outcome && \texttt{\footnotesize X-STALL} & stalled or kept repeating the same actions\\
\texttt{\footnotesize C-NO-UPDATE} & did not update the next step from feedback && \texttt{\footnotesize X-UNCERTAIN} & evidence not sufficient to decide\\
\texttt{\footnotesize C-NO-REPAIR} & did not recover after a failure && &\\
\texttt{\footnotesize C-PREREQ} & acted before a physical prerequisite held && &\\
\bottomrule\end{tabularx}
\end{table}

\subsection{Task-Level Results by View}
\label{app:per-view-detail}
This section gives the task-level success counts behind the per-view discussion of Section~\ref{subsec:task-results}.

\paragraph{Success is concentrated in a few tasks.} For three of the four VLAs, one task, passing through gates that open and close on a cycle (DR3), accounts for 36--41\% of all successes: 238 of 628 for KinRT, 236 of 574 for $\pi_{0.5}$ and 227 of 626 for X-VLA. For X-VLA, three tasks (DR3, cloth folding PR4 and resistance adjustment PR1) account for 549 of its 626 successes. Gemini-4-Argon shows the same pattern on a smaller scale: resistance adjustment and gate passage together contribute 24 of its 42 successes.

\paragraph{Eight tasks are completed by no system.} Workpiece tracking (M1), tray carrying with loose balls (DR4), oriented stamping (PR5), the domino chain (PR6), moving-rack loading (CP03), temperature-controlled cooking (CP04), bridge building (CP07) and whiteboard restoration (CP09) have no successful episode among the 56,700. On four of them (DR4, PR6, CP07 and CP09), every system also has a mean task score of zero, meaning that not even the first scored stage is reached. Each of these tasks has fifty scripted demonstrations per level that pass the same evaluator, confirming platform feasibility.

\subsubsection*{Mind}
\paragraph{End-to-end policies.} The VLAs complete few Mind episodes, and their successes are concentrated in a narrow set of tasks. KinRT's 132 Mind successes include 46 on steamer-stack restoration (P2), 22 on card matching (M2), and 20 on repeated-item recognition (E1). Of the 105 successes of $\pi_{0.5}$, 43 are on P2 and 22 on hazard navigation (S2). These successes occur mostly at the Easy level, where the least information must be retained (Table~\ref{tab:difficulty-counts} for $\pi_{0.5}$ and G0.5), and mostly in \emph{Clean} scenes (Section~\ref{subsec:difficulty-results}). Across the four VLAs, workpiece tracking (M1) is never completed and tray-layout restoration (S1) only once. A policy that conditions on the current observation or on a short history cannot recover information that has left its view, and the Mind results of the VLAs are consistent with this limitation. Evaluating memory-augmented policies~\cite{shi2025memoryvla,torne2026mem,li2025cronusvla} on the Mind view provides a direct test.

\paragraph{Agents.} GPT-6 Astra succeeds in 52.7\% of Mind episodes, but its performance is uneven within three of the five categories: 100\% on hazard navigation (S2) against 23.3\% on tray-layout restoration (S1), 93.3\% on card matching (M2) against 0\% on workpiece tracking (M1), and 43.3\% on meal orders (P1) against 0\% on steamer-stack restoration (P2). The tasks it solves best are those in which the action is a button press or a simple placement (S2, F1, M2, E2, E1), whereas the tasks it fails additionally require lifting covers, routing a workpiece through stations, or unstacking and restacking baskets. Its Mind score therefore reflects the reach of its motion tools as well as its memory, and because each category contains only two tasks, category-level differences should not be read as differences between kinds of memory. DeepSeek-V4.1-Flash shows the same pattern on a smaller scale: its 33 Mind successes all come from five of these tasks (M2 10, E1 6, E2 6, S2 6 and F1 5). Gemini-4-Argon obtains its 13 on six tasks.
\subsubsection*{Act}
\paragraph{End-to-end policies.} Act is the view in which the VLAs succeed most often, and their success comes mainly from dynamic response. Averaged over tasks, the four VLAs reach 22--28\% on the four dynamic tasks and 4--18\% on the six precision tasks. Although they are trained on the same demonstration collection, the policies differ sharply from task to task, and no single policy leads throughout: G0.5 leads on DR1 and DR2, KinRT on DR3, and X-VLA on PR1 and PR4. G0.5 leads light-cued response (DR2, 110 of 300 episodes) but is last on resistance adjustment (PR1, 9), where X-VLA leads with 130. X-VLA also completes 192 cloth-folding episodes (PR4), against 30 to 54 for the other three. Apart from the zero-success tasks DR4, PR5 and PR6, pen tracing (PR3) is completed only three times in 1,200 VLA episodes.

\paragraph{Agents.} GPT-6 Astra completes every resistance-adjustment episode and most episodes of the three timed tasks DR1, DR2, and DR3 (76.7\%, 90.0\%, and 80.0\%), yet it completes no episode of the five tasks that require continuous contact or a carried load (DR4, PR3, PR4, PR5, and PR6). The Hard level of DR2 allows 2.6\,s to press a lit button, whereas the agent's mean request time on Act tasks is 9.2\,s. Because simulation pauses during inference, the agent can still meet the deadline. These results show that the agent selects and times actions correctly in simulation time. They do not test real-time reaction.
\subsubsection*{Mind2Act}
Table~\ref{tab:composed-completion} lists the completed Mind2Act episodes of every system by task.

\begin{table}[!htbp]\centering\footnotesize
\caption{\textbf{Completed Mind2Act episodes by task.} Each task has 30 episodes per agent and 300 per end-to-end policy, pooled over the three difficulty levels and both settings.}
\label{tab:composed-completion}
\setlength{\tabcolsep}{2.6pt}
\begin{tabularx}{\linewidth}{@{}l>{\raggedright\arraybackslash}X*{9}{r}@{}}\toprule
\rowcolor{M2ASage}ID & Task & GPT-6 & DeepSeek & Gemini & KinRT & $\pi_{0.5}$ & G0.5 & X-VLA & InternW0 & FastWAM \\\midrule
CP01 & Conveyor meal assembly & 18 & \textcolor{black!35}{0} & \textcolor{black!35}{0} & 24 & 26 & 24 & 4 & 5 & 13 \\
CP02 & Rotating-cake decoration & 5 & \textcolor{black!35}{0} & \textcolor{black!35}{0} & 1 & 2 & \textcolor{black!35}{0} & \textcolor{black!35}{0} & 1 & \textcolor{black!35}{0} \\
CP03 & Moving-rack loading & \textcolor{black!35}{0} & \textcolor{black!35}{0} & \textcolor{black!35}{0} & \textcolor{black!35}{0} & \textcolor{black!35}{0} & \textcolor{black!35}{0} & \textcolor{black!35}{0} & \textcolor{black!35}{0} & \textcolor{black!35}{0} \\
CP04 & Temperature-controlled cooking & \textcolor{black!35}{0} & \textcolor{black!35}{0} & \textcolor{black!35}{0} & \textcolor{black!35}{0} & \textcolor{black!35}{0} & \textcolor{black!35}{0} & \textcolor{black!35}{0} & \textcolor{black!35}{0} & \textcolor{black!35}{0} \\
CP05 & Piano replay & 6 & \textcolor{black!35}{0} & \textcolor{black!35}{0} & \textcolor{black!35}{0} & \textcolor{black!35}{0} & \textcolor{black!35}{0} & \textcolor{black!35}{0} & \textcolor{black!35}{0} & \textcolor{black!35}{0} \\
CP06 & Produce weighing on a balance & 3 & \textcolor{black!35}{0} & \textcolor{black!35}{0} & 51 & 62 & 37 & 2 & 48 & 1 \\
CP07 & Bridge building for a toy car & \textcolor{black!35}{0} & \textcolor{black!35}{0} & \textcolor{black!35}{0} & \textcolor{black!35}{0} & \textcolor{black!35}{0} & \textcolor{black!35}{0} & \textcolor{black!35}{0} & \textcolor{black!35}{0} & \textcolor{black!35}{0} \\
CP08 & Block structure from a shown blueprint & 4 & \textcolor{black!35}{0} & \textcolor{black!35}{0} & 8 & 18 & \textcolor{black!35}{0} & \textcolor{black!35}{0} & 35 & \textcolor{black!35}{0} \\
CP09 & Whiteboard restoration & \textcolor{black!35}{0} & \textcolor{black!35}{0} & \textcolor{black!35}{0} & \textcolor{black!35}{0} & \textcolor{black!35}{0} & \textcolor{black!35}{0} & \textcolor{black!35}{0} & \textcolor{black!35}{0} & \textcolor{black!35}{0} \\
CP10 & Return or dodge incoming balls & \textcolor{black!35}{0} & \textcolor{black!35}{0} & \textcolor{black!35}{0} & 4 & 1 & 8 & \textcolor{black!35}{0} & 2 & 2 \\
\midrule
\multicolumn{2}{@{}l}{\textit{All ten tasks}} & \textbf{36} & \textbf{0} & \textbf{0} & \textbf{88} & \textbf{109} & \textbf{69} & \textbf{6} & \textbf{91} & \textbf{16} \\
\bottomrule\end{tabularx}
\end{table}

\paragraph{The composition gap of the strongest system.} GPT-6 Astra succeeds in 52.7\% of Mind and 39.0\% of Act episodes. If a Mind2Act workflow required one independent success of each kind, 20.5\% of its Mind2Act episodes would succeed. It completes only 12.0\% (36 of 300, 95\% Wilson interval [8.8,~16.2]\%). Eighteen of the 36 successes come from conveyor meal assembly (CP01), and it completes none of CP03, CP04, CP07, CP09 or CP10. The two weaker agents complete no Mind2Act episode, which gives a 95\% Wilson interval of [0, 1.3]\% for each of their success rates.

\paragraph{For the VLAs, composition is not the first bottleneck.} The VLAs complete 0.2--3.6\% of Mind2Act episodes, close to their Mind success rates (1.0--4.4\%) and far below their Act rates. For KinRT, $\pi_{0.5}$ and G0.5, the Mind2Act rate exceeds the product of the Mind and Act rates (for example, 3.6\% against 0.4\% for $\pi_{0.5}$), and for X-VLA the two are about equal (0.2\%). This reflects concentration: produce weighing (CP06) and conveyor meal assembly (CP01) supply 75 of KinRT's 88 Mind2Act successes, 88 of the 109 of $\pi_{0.5}$, and 61 of the 69 of G0.5, mostly at the Easy level. At Easy, CP01 has a single order and CP06 a single item to weigh, so little must be remembered and a practiced motion carries most of the task. For these policies, the two workflows then behave much like Act tasks. InternW0 departs from this pattern: block structure from a shown blueprint (CP08) contributes 35 of its 91 Mind2Act successes, more than any other end-to-end policy completes on that task, while produce weighing (CP06) contributes 48.

\paragraph{Mind2Act workflows mostly fail early.} On the four workflows that no system completes (CP03, CP04, CP07 and CP09), the mean task score of every system is at most 1.2 points out of 100. The profile card of G0.5 (Fig.~\ref{fig:profile-card-g05}) shows why for one policy: 202 of its 300 bridge-building episodes (CP07) never grasp a piece. Elsewhere, partial credit adds little. For every system, the Mind2Act score exceeds the success rate by at most 1.5 points, so under the current scoring rules a failed Mind2Act episode has usually failed early rather than near completion. For most systems evaluated here, these four workflows therefore end at the first physical step, before the composition they were designed to test. InternW0 on cooking (CP04) is the exception: in 271 of its 300 episodes it switches on the heat before all ingredients are in the pot, which is a composition error.

\paragraph{What the view-level comparison shows.} Conveyor meal assembly (CP01) reuses the conveyor interception of DR1 and, like P1, depends on remembered food orders. Yet GPT-6 Astra completes CP01 more often (60.0\%) than P1 (43.3\%), and every VLA that completes any CP01 episode does so more often than the product of its P1 and DR1 rates would predict (8.0\% for KinRT against 0.1\%). A Mind2Act task is thus not simply the conjunction of two single-view tasks, and a lower Mind2Act score cannot by itself be attributed to composition. The views instead show a narrower finding: each system's ability is uneven across Mind and Act, the systems differ in which is weaker, and every system succeeds less often on the Mind2Act workflows than on its stronger single view. 

\subsection{What Each Diagnosis Layer Settles}
\label{subsec:rule-layer-counts}
Table~\ref{tab:rule-layer-counts} lists, for the three systems of Appendix~\ref{subsec:profile-cards}, how many failed episodes the rule layer settles and how many pass to the LLM judge. Where the judge is uncertain whether the evidence decides the first unrecoverable step, or where repeated judge calls disagree, human reviewers settle the code. The lower block compares the attribution of the rule-settled failures with that of all failures. For the two end-to-end policies the rule layer attributes more failures to the Mind side than to any other, because scorer records reveal a misremembered item more readily than video reveals a motor fault. Over all failures, execution errors are the largest group.

\begin{table}[!htb]\centering\small
\caption{\textbf{Failures settled by each diagnosis layer.} Top: failed episodes settled by the rule layer and passed to the LLM judge. Bottom: share of failures attributed to each side, among the failures the rules settle and among all failures. The remainder carry episode-level codes.}
\label{tab:rule-layer-counts}
\setlength{\tabcolsep}{5pt}
\begin{tabular}{@{}lrrr@{}}\toprule
System & Failed & Rule layer & LLM judge\\\midrule
G0.5 (VLA) & 8,470 & 3,230 (38.1\%) & 5,240\\
InternW0 (WAM) & 8,220 & 2,760 (33.6\%) & 5,460\\
GPT-6 Astra (agent) & 589 & 131 (22.2\%) & 458\\
\bottomrule\end{tabular}\\[6pt]
\begin{tabular}{@{}lrrrrrr@{}}\toprule
 & \multicolumn{3}{c}{Rule layer only} & \multicolumn{3}{c}{All layers}\\\cmidrule(lr){2-4}\cmidrule(l){5-7}
System & Mind & Act & Comp. & Mind & Act & Comp.\\\midrule
G0.5 & 44.9\% & 31.8\% & 7.7\% & 27.7\% & 41.4\% & 6.1\%\\
InternW0 & 53.0\% & 23.4\% & 23.4\% & 32.8\% & 37.2\% & 10.5\%\\
GPT-6 Astra & 27.5\% & 54.2\% & 18.3\% & 30.4\% & 55.3\% & 7.1\%\\
\bottomrule\end{tabular}
\end{table}

\subsection{Model Profile Cards of Three System Types}
\label{subsec:profile-cards}
\begin{figure}[!htbp]\centering
\resizebox{\linewidth}{!}{\input{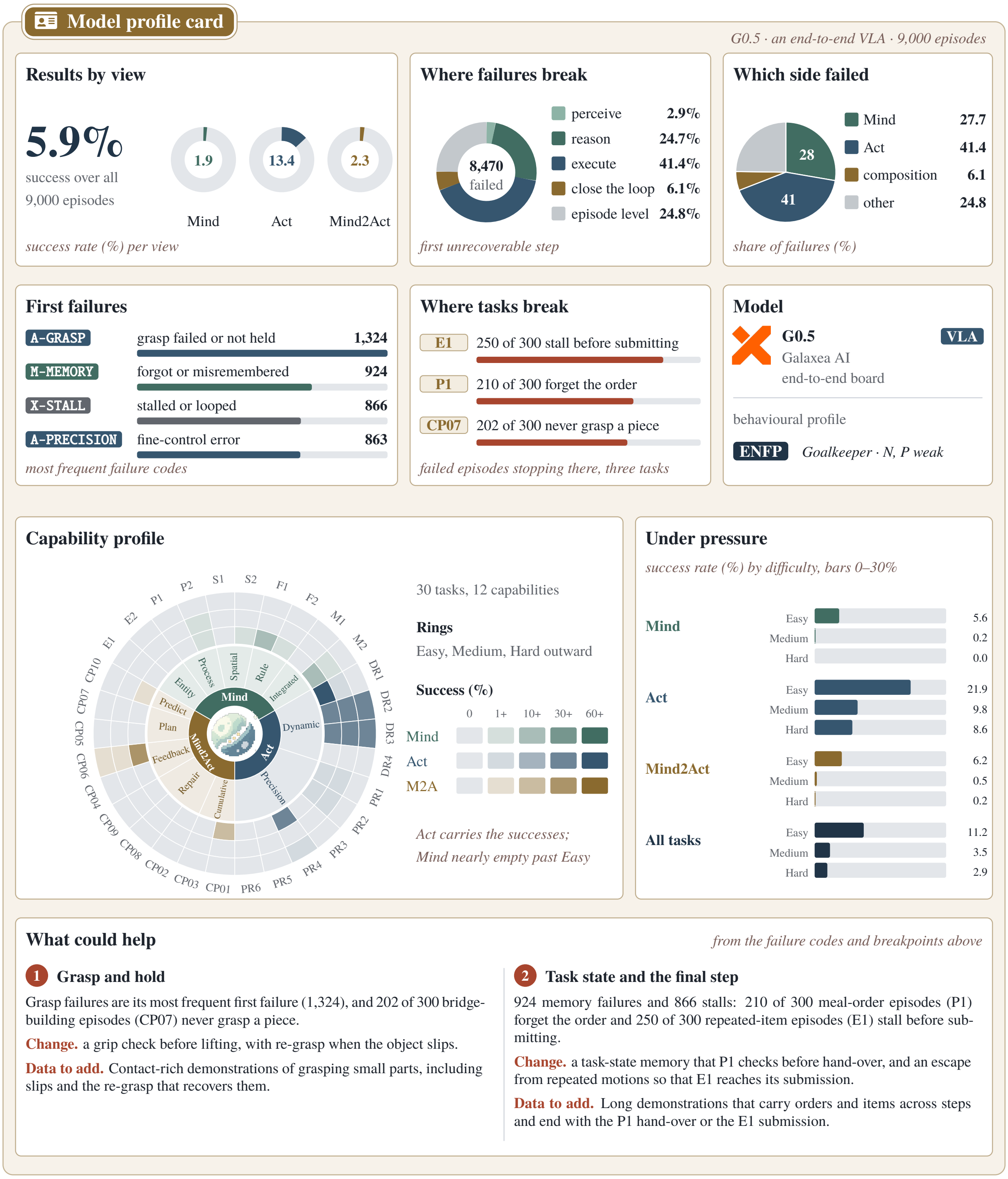}}
\caption{\textbf{Model profile card of G0.5 (VLA).} Panels as in Fig.~\ref{fig:profile-card-example}. \emph{Other} in the side panel collects failures with an episode-level code, such as stalls and cases the evidence cannot decide. The memory and lifestyle letters of the behavioral profile (N, P) are weak leans (Appendix~\ref{app:profiles-fun}).}
\label{fig:profile-card-g05}
\end{figure}

\begin{figure}[!htbp]\centering
\resizebox{\linewidth}{!}{\input{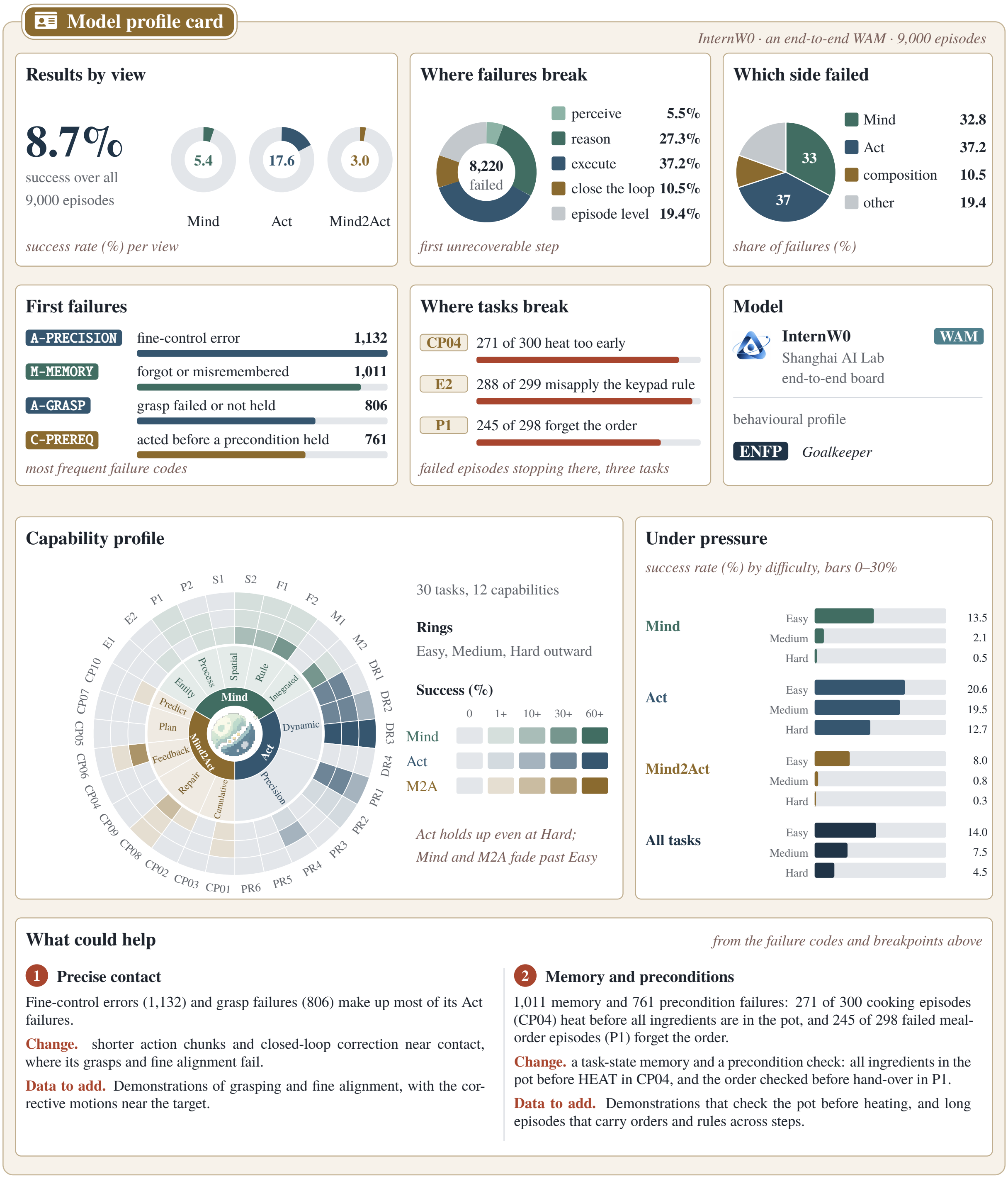}}
\caption{\textbf{Model profile card of InternW0 (WAM).} Panels as in Fig.~\ref{fig:profile-card-g05}.}
\label{fig:profile-card-internw0}
\end{figure}

\begin{figure}[!htbp]\centering
\resizebox{\linewidth}{!}{\input{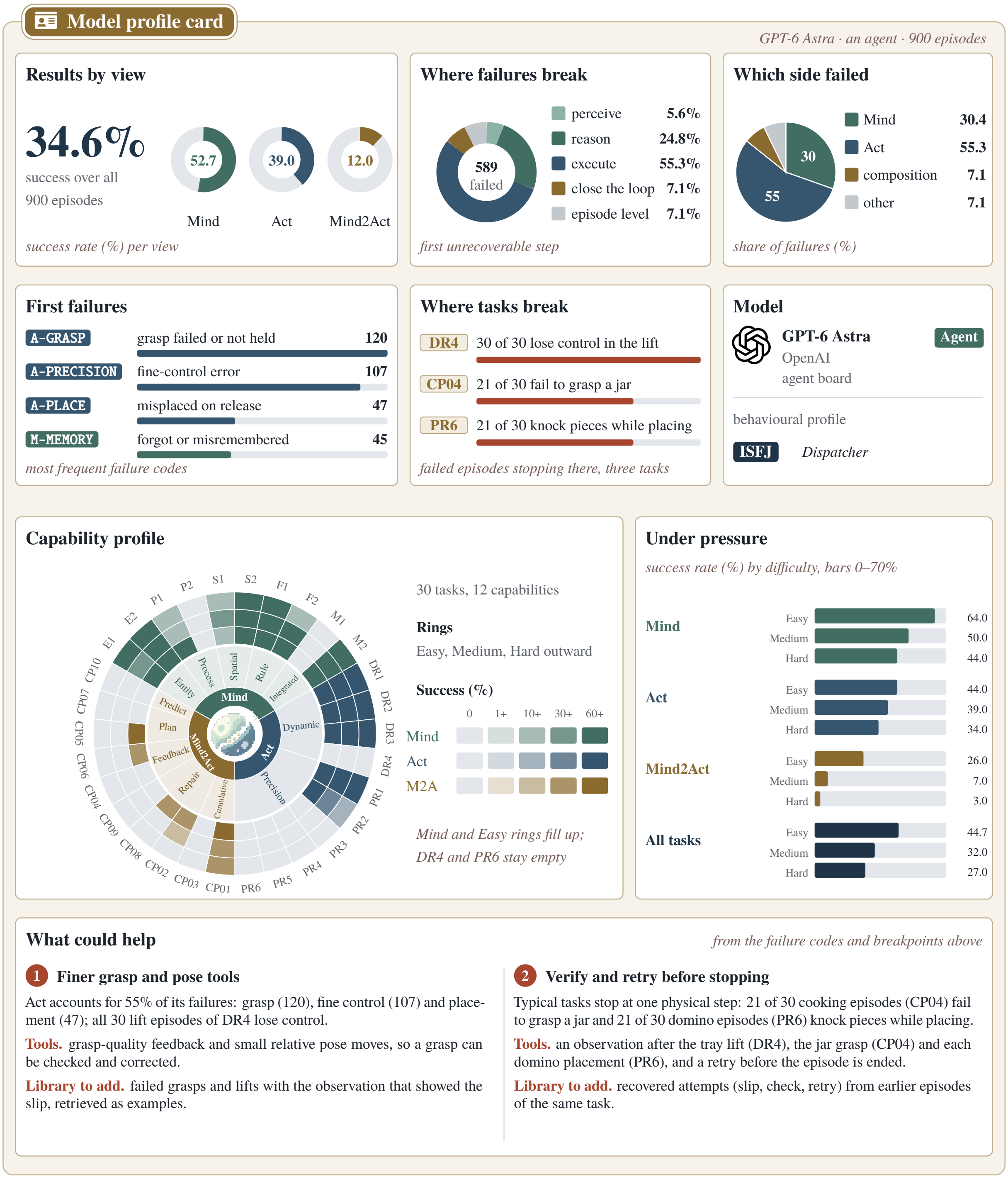}}
\caption{\textbf{Model profile card of GPT-6 Astra (agent).} Panels as in Fig.~\ref{fig:profile-card-g05}. Each task and level has ten episodes, and the difficulty bars run to 70\%.}
\label{fig:profile-card-astra}
\end{figure}

Figs.~\ref{fig:profile-card-g05}--\ref{fig:profile-card-astra} give the model profile cards of G0.5, InternW0 and GPT-6 Astra, one representative system of each evaluated type (a VLA, a WAM and an agent), in the format of Fig.~\ref{fig:profile-card-example}. Unlike the examples in Figs.~\ref{fig:diagnosis-card} and~\ref{fig:profile-card-example}, these cards report measured results. The success rates are those of Table~\ref{tab:main-results}. The failure panels cover every failed episode: 8,470 for G0.5, 8,220 for InternW0 and 589 for GPT-6 Astra. Rules settle the episodes they can decide, the judge reviews the rest, and human reviewers settle the cases the judge marks as uncertain (Section~\ref{subsec:diagnosis-harness}).

\paragraph{Reading the three cards.} The two end-to-end policies share a profile but break in different ways. G0.5 most often fails to grasp or hold an object (1,324 episodes) and stalls in 866 episodes. In repeated-item recognition (E1), 250 of its 300 episodes stop before the batch is submitted. InternW0 stalls less, but in 761 episodes it proceeds before a precondition holds: in 271 of its 300 cooking episodes (CP04), it switches on the heat before all recipe ingredients are in the pot. Fine-control errors are its most frequent code (1,132 episodes), and it misapplies the keypad rule in 288 of 299 failed episodes of E2. GPT-6 Astra fails mostly on the execution side (55\% of its failures, against 30\% on the Mind side), led by grasp (120) and fine-control errors (107): every tray-carrying episode (DR4) is lost during the lift, and 21 of 30 domino-chain episodes (PR6) end with pieces knocked over while they are placed. The cards therefore point to different remedies: grasp demonstrations and task-state memory for G0.5, contact-rich demonstrations and precondition checks for InternW0, and finer motion tools with a check-and-retry step before stopping for the agent.

\FloatBarrier
\subsection{Stop Reasons of Agent Episodes}
\label{subsec:qual-agents}
Table~\ref{tab:agent-stops} gives the reasons for which the episodes of the three agents ended, as used in Section~\ref{subsec:failure-recovery}. A stop by the agent is its own decision to end the episode. The completion flag is set by the environment, not by the agent. It records how the episode ended and does not imply a full score, which comes from the task evaluator. The two budget rows count episodes that reach the limit of 150 decisions or of simulated time. The stop reasons are read from the termination field of each retained episode result. On average, an episode uses 79.6 decisions for GPT-6 Astra, 111.8 for DeepSeek-V4.1-Flash and about 33 for Gemini-4-Argon.

\begin{table}[!htb]\centering\small
\caption{\textbf{Stop reasons of agent episodes.} Number of episodes by stop reason in the \emph{Clean} and \emph{Random} settings, 450 episodes each.}
\label{tab:agent-stops}
\setlength{\tabcolsep}{5pt}
\begin{tabular}{@{}lrrrrrr@{}}\toprule
 & \multicolumn{2}{c}{GPT-6 Astra} & \multicolumn{2}{c}{DeepSeek-V4.1-Flash} & \multicolumn{2}{c}{Gemini-4-Argon}\\\cmidrule(lr){2-3}\cmidrule(lr){4-5}\cmidrule(l){6-7}
Stop reason & Clean & Random & Clean & Random & Clean & Random\\\midrule
Stopped by the agent & 223 & 237 & 175 & 210 & 409 & 420\\
Decision budget exhausted & 15 & 14 & 160 & 126 & 0 & 0\\
Physics budget exhausted & 37 & 38 & 50 & 44 & 4 & 2\\
Completion flag recorded & 175 & 161 & 64 & 70 & 37 & 28\\
Adjudicated manually & 0 & 0 & 1 & 0 & 0 & 0\\\midrule
Total & 450 & 450 & 450 & 450 & 450 & 450\\
\bottomrule\end{tabular}
\end{table}
\FloatBarrier
\subsection{Behavioral Profiles, Just for Fun}
\label{app:profiles-fun}
Each model profile card also carries a four-letter behavioral profile, such as the ENFP of Fig.~\ref{fig:profile-card-example}. It borrows the MBTI format to turn four comparisons of success rates into a code that is easy to remember and share. Each letter asks one MBTI-style question of a robot policy and answers it with two task groups of the benchmark (Table~\ref{tab:profile-letters}). The system takes the letter of the group on which it succeeds more often, so every profile has four letters. Where the two rates differ by less than one percentage point (G0.5's N and P), the card keeps the letter and marks the lean as weak. A profile says how a system tends to succeed and fail, not how well it performs. The numbers behind it are in the quantitative sections of the card.

\begin{table}[!htb]\centering\footnotesize
\caption{\textbf{Definition of the four profile letters.} Each dimension keeps the question that MBTI asks about a person and answers it with two task groups of the benchmark. The system takes the letter of the group on which it does better. Letter~1 compares the two single-capability views, and letters~2--4 each compare two task groups within one view (column \emph{Measured on}).}
\label{tab:profile-letters}
\definecolor{lA}{HTML}{A8452E}\definecolor{lB}{HTML}{416D62}\definecolor{lC}{HTML}{36566F}\definecolor{lD}{HTML}{8A6A2F}
\setlength{\tabcolsep}{3.5pt}
\begin{tabularx}{\linewidth}{@{}l c >{\raggedright\arraybackslash\hsize=0.8\hsize}X >{\raggedright\arraybackslash\hsize=1.2\hsize}X l@{}}\toprule
Dimension & Letter & In MBTI, a person & In Mind2Act World, a policy & Measured on\\\midrule
\multirow{2}{*}{1 \;Energy} & \textbf{\textcolor{lA}{E}} & turns outward, to action & Act-led: executes better than it reasons & Act view\\
 & \textbf{\textcolor{lA}{I}} & turns inward, to thought & Mind-led: reasons better than it executes & Mind view\\\addlinespace[2pt]
\multirow{2}{*}{2 \;Information} & \textbf{\textcolor{lB}{S}} & attends to concrete facts & keeps track of \emph{what} each object is & E1, E2\\
 & \textbf{\textcolor{lB}{N}} & attends to the overall pattern & keeps track of \emph{where} things are & S1, S2\\\addlinespace[2pt]
\multirow{2}{*}{3 \;Decisions} & \textbf{\textcolor{lC}{T}} & decides by objective criteria & meets a fixed tolerance (precision) & PR1--PR6\\
 & \textbf{\textcolor{lC}{F}} & decides by the situation & acts as the scene changes (reactivity) & DR1--DR4\\\addlinespace[2pt]
\multirow{2}{*}{4 \;Lifestyle} & \textbf{\textcolor{lD}{J}} & prefers a settled plan & follows its own plan (feedforward) & CP01, 03, 05, 07, 10\\
 & \textbf{\textcolor{lD}{P}} & stays open and adapts & lets outcomes set the next step (feedback) & CP02, 04, 06, 08, 09\\\bottomrule
\end{tabularx}
\end{table}

\paragraph{How the letters are borrowed.} Each MBTI dimension is matched to the closest contrast in the benchmark. For example, extraverts get things done out in the world and introverts think things over, so a policy that does better on Act is an E and one that does better on Mind is an I. The other three letters follow Table~\ref{tab:profile-letters} in the same way. The information letter uses only entity memory (E1, E2) and spatial memory (S1, S2). The other Mind categories mix several kinds of memory and stay out of the letters.

\begin{figure}[!htb]\centering
\resizebox{\linewidth}{!}{\input{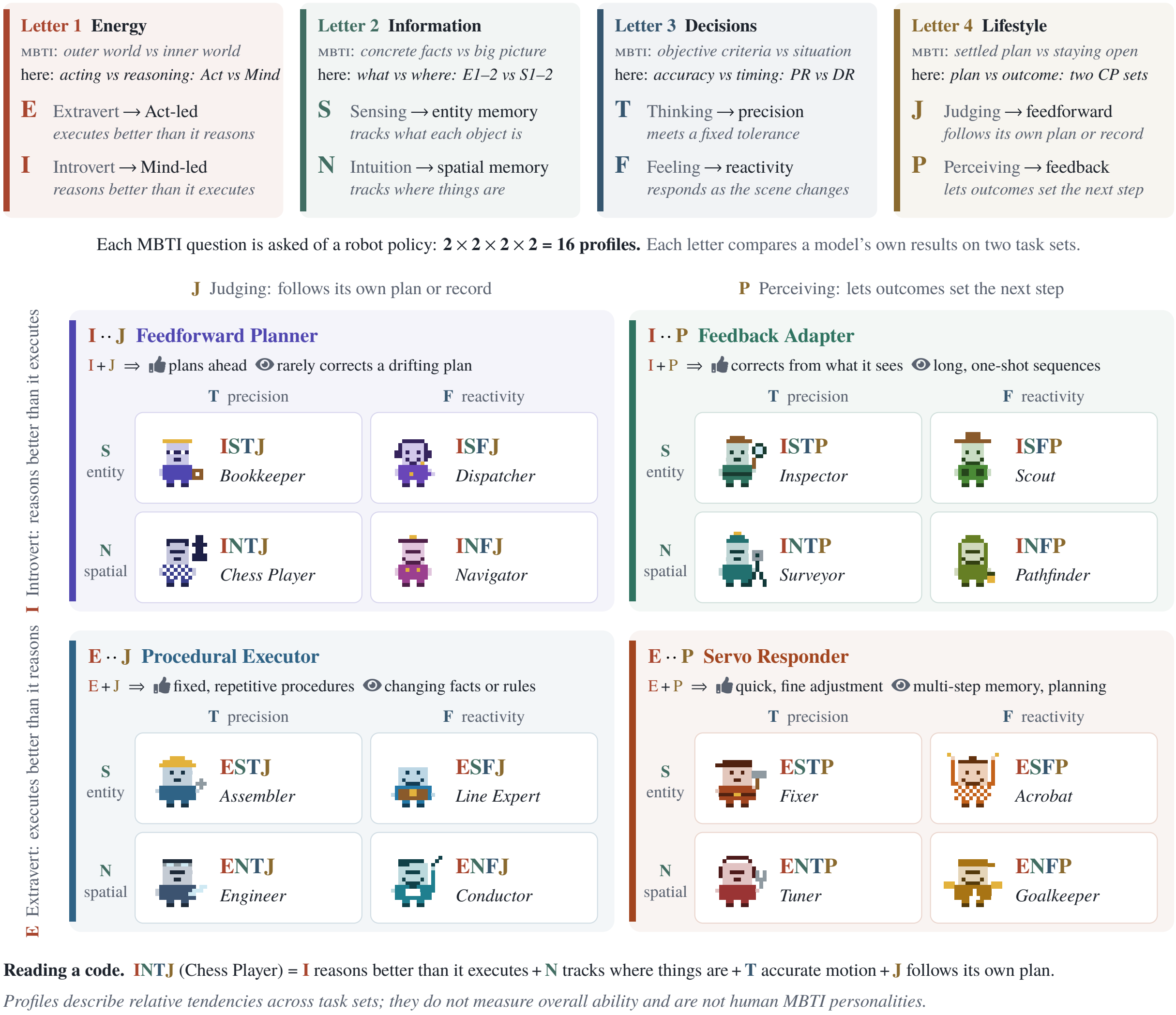}}
\caption{\textbf{The sixteen behavioral profiles.} Each letter keeps the question that MBTI asks about a person (top: energy, information, decisions, lifestyle) and answers it with two task groups of the benchmark. Each letter keeps its color wherever it appears. Letter~1 compares the Act and Mind views, and letters~2--4 each compare two task groups within one view, giving $2\times2\times2\times2=16$ profiles. The first and last letters define the four families, and each family box gives a playful reading of its two letters as a strength (\faThumbsUp) and a weak spot (\faEye). Within a family, rows give the memory letter (S/N) and columns the execution letter (T/F). Codes, nicknames and pixel characters follow the project website.}
\label{fig:profile-codes}
\end{figure}

As in MBTI, the first and last letters sort the codes into four families. \emph{Feedforward Planners} (I\,\textperiodcentered\textperiodcentered\,J) think more than they do and stick to a plan made in advance. \emph{Feedback Adapters} (I\,\textperiodcentered\textperiodcentered\,P) think more than they do and correct course step by step. \emph{Procedural Executors} (E\,\textperiodcentered\textperiodcentered\,J) do more than they think and follow a fixed routine. \emph{Servo Responders} (E\,\textperiodcentered\textperiodcentered\,P) have steady hands that adjust as the scene changes. The middle letters add the details: what the system remembers better (S or N) and whether its hands are more accurate (T) or more timely (F). Since every end-to-end policy evaluated here succeeds more often on Act than on Mind (Table~\ref{tab:main-results}), all of them start with E, while GPT-6 Astra starts with I.

\paragraph{Meet the three diagnosed systems.} InternW0 is an \textbf{ENFP}, a Goalkeeper among the Servo Responders, and holds all four letters firmly: it succeeds more often on Act than on Mind (17.6\% against 5.4\%), on spatial than on entity memory (5.3\% against 0.2\%), on dynamic response than on precise execution (31.8\% against 8.1\%), and on feedback-driven than on plan-driven workflows (5.6\% against 0.5\%). G0.5 is an ENFP too, though only just on the memory and lifestyle letters (0.0\% against 0.8\%, and 2.1\% against 2.5\%). GPT-6 Astra is an \textbf{ISFJ}, a Dispatcher among the Feedforward Planners: Mind-led (52.7\% against 39.0\% on Act), better at entity than spatial memory (68.3\% against 61.7\%), more successful on dynamic response than on precise execution (61.7\% against 23.9\%, though its F says only that the timed tasks go better than the precision tasks, not that it reacts fast), and better on plan-driven than on feedback-driven workflows (16.0\% against 8.0\%). In short, the two Goalkeepers do relatively best on dynamic response, and the Dispatcher completes plan-driven workflows twice as often as feedback-driven ones.

\FloatBarrier

\begingroup
\raggedbottom
\setlength{\emergencystretch}{2em}
\setlength{\parskip}{3pt}
\titlespacing*{\subsection}{0pt}{0pt}{2pt}
\titlespacing*{\paragraph}{0pt}{6pt}{1em}

\newcommand{\taskcardfooter}[1]{%
  \noindent\textbf{Mind2Act focus:} #1\qquad\mbox{\textbf{Usage:} Train \& Eval.}\par}

\newcommand{\taskcardgap}{\par\vspace{2mm}}

\FloatBarrier
\section{Task Details}
\label{app:tasks}
This appendix is the reference for the tasks. It explains where they come from, what each one asks, and how it gets harder. Mind tasks emphasize retaining information and selecting actions, Act tasks emphasize physical precision and timely response, and Mind2Act tasks combine remembered or inferred task state with execution outcomes that constrain subsequent actions. Table~\ref{tab:task-summary} lists all thirty tasks. Every task has 50 demonstrations per level (4,500 in total) and is covered by all nine evaluated systems (Table~\ref{tab:per-task-sr}).

The three levels describe changes in task demands rather than a calibrated difficulty ranking. Demonstration and evaluation seeds are disjoint. The common observation and scoring protocol in Section~\ref{sec:protocol} applies throughout.

\subsection{From Real-World Requirements to Tasks}
\label{app:real-world-mapping}
Each task is built from a requirement that makes a real activity hard for a robot. Table~\ref{tab:task-summary} names, for every task, the real-world source it is built from, its capability category (for Mind2Act, the demand pairing) and the variables raised with difficulty. Fig.~\ref{fig:realworld-mapping} groups the same tasks into the five application domains of Table~\ref{tab:application-domains}.
\begin{figure}[!htb]\centering
\resizebox{\linewidth}{!}{\input{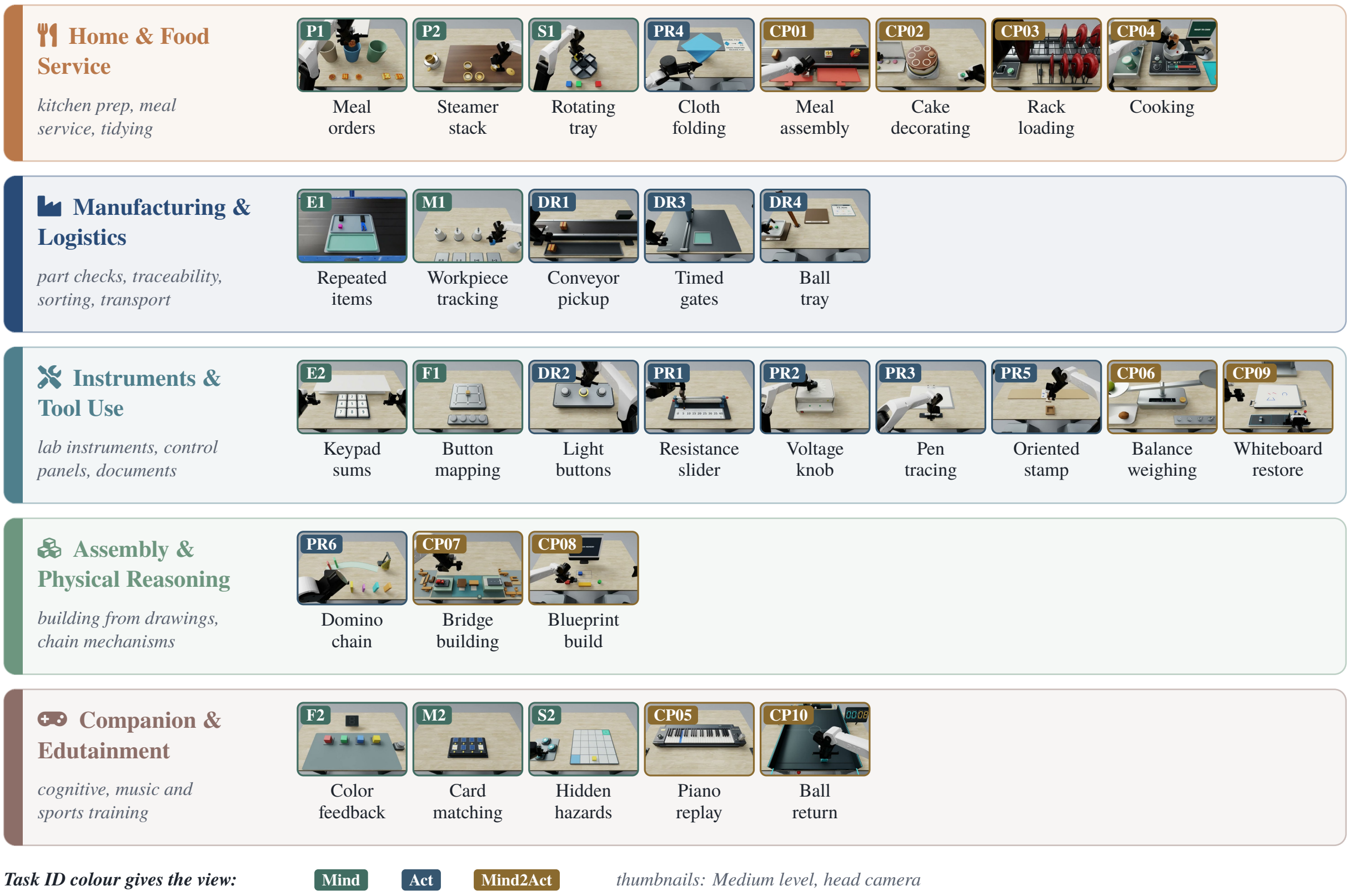}}%
\caption{\textbf{Tasks grouped by application domain.} Each task sits under the domain whose real-world requirement it is built from. The color of each task ID gives its view.}
\label{fig:realworld-mapping}
\end{figure}
\begin{table}[!htbp]
\centering
\caption{\textbf{Task summary of Mind2Act World.} Each view contains ten tasks, and every task has Easy, Medium, and Hard levels. \emph{Source} names the real-world requirement each task is built from; Table~\ref{tab:application-domains} groups the tasks by application domain. The last column gives the capability category (for Mind2Act, the memory or inference demand $\leftrightarrow$ the execution demand) and, after the semicolon, the main variables raised with difficulty. Instructions and success conditions are given on the task cards (Appendix~\ref{subsec:task-cards}).}
\label{tab:task-summary}
\begingroup
\fontsize{8.2}{9.9}\selectfont
\setlength{\tabcolsep}{3.5pt}
\renewcommand{\arraystretch}{1.08}
\begin{tabularx}{\linewidth}{@{}l l >{\raggedright\arraybackslash}p{0.2\linewidth} >{\raggedright\arraybackslash}p{0.19\linewidth} >{\raggedright\arraybackslash}X@{}}
\toprule
\textbf{View} & \textbf{ID} & \textbf{Task} & \textbf{Source} & \textbf{Category or demand pairing; difficulty variables} \\
\midrule
\multirow{10}{*}{\emph{Mind}}
 & E1 & Repeated-item recognition & Part inspection & Entity memory; batches and objects \\
 & E2 & Object-coded keypad arithmetic & Calculation from a reference & Entity memory (object-value bindings); bindings, expression length, repeats \\
 & F1 & Unknown button mapping & Unfamiliar control panels & Feedback-rule learning; grid size and route length \\
 & F2 & Color-block arrangement by feedback & Mastermind-style trial & Feedback-rule learning (hidden arrangement); blocks and submission budget \\
 & P1 & Incremental meal orders & Food-service orders & Process memory; containers, deposits, and new orders \\
 & P2 & Steamer-stack order restoration & Dim-sum steamer service & Process memory (layer order); layers and distractor steamers \\
 & S1 & Tray-layout restoration & Tray and station setup & Spatial memory; rotation segments and reversals \\
 & S2 & Hidden-hazard navigation & Remembered safe routes & Spatial memory (hidden map); board size and hazard cells \\
 & M1 & Workpiece tracking and route replay & Workpiece routing & Integrated memory; exchanges and repeated visits \\
 & M2 & Card-matching game & Memory card game & Integrated memory (pattern-location pairs); 2/4/8 pairs \\
\midrule
\multirow{10}{*}{\emph{Act}}
 & DR1 & Conveyor pickup & Conveyor picking & Dynamic response; number of parcels \\
 & DR2 & Light-cued response & Signal-light operation & Dynamic response; response window \\
 & DR3 & Timed passage through cyclic gates & Interlocked access doors & Dynamic response; open window, second gate \\
 & DR4 & Tray carrying with loose balls & Serving trays & Dynamic response (stability); balls, pillars, allowed drift \\
 & PR1 & Resistance adjustment & Instrument adjustment & Precise execution; tick resolution, interpolation and tolerance \\
 & PR2 & Voltage setting & Instrument adjustment & Precise execution; tick resolution, interpolation and tolerance \\
 & PR3 & Pen tracing & Tool-guided contact & Precise execution; path curvature \\
 & PR4 & Cloth folding and flat placement & Folding linens & Precise execution (deformable); fold type and outline \\
 & PR5 & Oriented stamping & Document stamping & Precise execution (tool orientation);\newline frame size and angle tolerance \\
 & PR6 & Domino chain to a bell & Domino setup & Precise execution (placement for a physical chain); dominoes and size variation \\
\midrule
\multirow{10}{*}{\emph{Mind2Act}}
 & CP01 & Conveyor meal assembly & Meal assembly line & Remaining demand $\leftrightarrow$ conveyor interception; orders, meal-tray stack \\
 & CP02 & Rotating-cake decoration & Food decoration & Layout bindings, device state $\leftrightarrow$ placement on a rotating surface; macarons, rotation speed, box, turntable and cover steps \\
 & CP03 & Moving-rack loading & Dish-rack loading & Loading progress, gear $\leftrightarrow$ moving-slot insertion; plates, speed, clearance \\
 & CP04 & Temperature-controlled cooking & Cooking with appliances & Recipe order, added ingredients $\leftrightarrow$ bimanual jar handling, thermal feedback; ingredients, drawer and pot cover, keep-warm window \\
 & CP05 & Piano replay & Sequence reproduction & Note sequence $\leftrightarrow$ effective strikes with tools; keys, sequence length, mallets, volume \\
 & CP06 & Produce weighing on a balance & Weighing goods & Mass inferred from beam deflection $\leftrightarrow$ placing weights, sliding rider; items, tolerance band \\
 & CP07 & Bridge building for a toy car & Construction with limited stock & Material selection and build plan $\leftrightarrow$ stacking and fitting parts; span, stock, distractors \\
 & CP08 & Block structure from a shown blueprint & Assembly from drawings & Remembered blueprint, actual layer state $\leftrightarrow$ supported stacking, rework; blocks, layers, distractors \\
 & CP09 & Whiteboard restoration & Restoring a drawn board & Remembered original, current board $\leftrightarrow$ bimanual erasing, uncapping, redrawing; strokes, colors \\
 & CP10 & Return or dodge incoming balls & Paddle ball game & Color rule, predicted trajectory $\leftrightarrow$ timed blocker placement; balls, rebounds, countdown \\
\bottomrule
\end{tabularx}
\endgroup
\end{table}

\FloatBarrier
\subsection{Task Cards}
\label{subsec:task-cards}
\label{app:metrics}
\label{subsec:success-predicates}
Each card shows four frames from one expert recording. The frames illustrate the task and may come from a lower level than the listed instruction. The card also gives the English instruction as stored in the demonstrations (the Hard-level text where the levels differ, with a per-episode target inserted in PR1 and PR2), the condition for final success, and the Easy/Medium/Hard configuration. The category follows Section~\ref{sec:benchmark}. Success is evaluated on simulator states and events, some of them checked throughout the episode. A note in parentheses gives events that are irreversible or unscored. Every task is used for both training and evaluation, with 50 demonstrations per task and level (Section~\ref{subsec:data-statistics}). The score values of each task and level are listed on the task pages of the project website.
\par\vspace{2mm}
\definecolor{CardMind}{HTML}{416D62}\definecolor{CardAct}{HTML}{36566F}\definecolor{CardCP}{HTML}{8A6A2F}
\newcommand{\mcard}[7]{%
\par\noindent\begin{minipage}{\linewidth}\phantomsection\label{app:task:#1}%
{\color{#3}\rule[-0.2ex]{2pt}{2.4ex}}\enspace{\small\bfseries\color{#3}#1}\enspace{\small\bfseries #2}\hfill{\footnotesize\itshape\color{Muted}#4}\par\vspace{1.5pt}
\begin{minipage}[t]{0.555\linewidth}\vspace{0pt}\resizebox{\linewidth}{!}{\input{figures/task_cards/#1.tex}}\end{minipage}\hfill
\begin{minipage}[t]{0.43\linewidth}\vspace{0pt}\fontsize{8}{9.6}\selectfont\raggedright
\textbf{Instruction.} #5\par\vspace{2pt}\textbf{Success.} #7\par\vspace{2pt}\textbf{E / M / H.} #6\end{minipage}
\end{minipage}\par\vspace{2.6mm}}

\subsubsection*{Mind tasks}
\label{app:cards-mind}
{\footnotesize\color{Muted}Ten tasks that test memory and reasoning with routine execution.}\par\vspace{2mm}
\par\noindent\begin{minipage}{\linewidth}\phantomsection\label{app:task:E1}%
{\color{CardMind}\rule[-0.2ex]{2pt}{2.4ex}}\enspace{\small\bfseries\color{CardMind}E1}\enspace{\small\bfseries Repeated-item recognition}\hfill{\footnotesize\itshape\color{Muted}Entity memory}\par\vspace{1.5pt}
\begin{minipage}[t]{0.555\linewidth}\vspace{0pt}\resizebox{\linewidth}{!}{\input{figures/task_cards/E1.tex}}\end{minipage}\hfill
\begin{minipage}[t]{0.43\linewidth}\vspace{0pt}\fontsize{8}{9.6}\selectfont\raggedright
\textbf{Instruction.} \par\vspace{2pt}\textbf{Success.} C\par\vspace{2pt}\textbf{E / M / H.} V\end{minipage}
\end{minipage}\par\vspace{2.6mm}
{\emph{Hard:} Observe and remember each batch of items. After the first batch, place the one item that appeared in any of the previous five batches into the submission tray.}%
{3/5/7 batches, 4/4/6 objects per batch and 2/4/6 submissions. Repeats come from the previous 1, up to 3, or up to 5 batches.}%
{Every repeated item stably deposited in its submission tray (first batch is observation-only).}
\par\noindent\begin{minipage}{\linewidth}\phantomsection\label{app:task:E2}%
{\color{CardMind}\rule[-0.2ex]{2pt}{2.4ex}}\enspace{\small\bfseries\color{CardMind}E2}\enspace{\small\bfseries Object-coded keypad arithmetic}\hfill{\footnotesize\itshape\color{Muted}Entity memory}\par\vspace{1.5pt}
\begin{minipage}[t]{0.555\linewidth}\vspace{0pt}\resizebox{\linewidth}{!}{\input{figures/task_cards/E2.tex}}\end{minipage}\hfill
\begin{minipage}[t]{0.43\linewidth}\vspace{0pt}\fontsize{8}{9.6}\selectfont\raggedright
\textbf{Instruction.} \par\vspace{2pt}\textbf{Success.} C\par\vspace{2pt}\textbf{E / M / H.} V\end{minipage}
\end{minipage}\par\vspace{2.6mm}
{Observe the visible entity-value associations, remember them, decode the physical expression, and enter its exact result using only the closed gripper fingertips on the nine keys.}%
{3/4/6 object-value bindings shown, then hidden; the board then shows an arithmetic expression over the same objects (Medium and Hard repeat objects), whose result is typed on keys 1--9.} %
{All digits of the exact result entered in order after effective press-release, with no wrong, missing or extra press.}
\par\noindent\begin{minipage}{\linewidth}\phantomsection\label{app:task:F1}%
{\color{CardMind}\rule[-0.2ex]{2pt}{2.4ex}}\enspace{\small\bfseries\color{CardMind}F1}\enspace{\small\bfseries Unknown button mapping}\hfill{\footnotesize\itshape\color{Muted}Feedback-rule learning}\par\vspace{1.5pt}
\begin{minipage}[t]{0.555\linewidth}\vspace{0pt}\resizebox{\linewidth}{!}{\input{figures/task_cards/F1.tex}}\end{minipage}\hfill
\begin{minipage}[t]{0.43\linewidth}\vspace{0pt}\fontsize{8}{9.6}\selectfont\raggedright
\textbf{Instruction.} \par\vspace{2pt}\textbf{Success.} C\par\vspace{2pt}\textbf{E / M / H.} V\end{minipage}
\end{minipage}\par\vspace{2.6mm}
{\emph{Hard:} Probe the four unmarked buttons and remember how each moves the slider. After the slider resets and the goal appears, reach the visible goal on the 4 by 4 grid using the fewest presses.}%
{A linear track, a 3 by 3 grid, or a 4 by 4 grid, with 2/4/4 buttons. Shortest routes are 1--3/2--3/4--6 steps. Exploration cost is recorded separately.}%
{Slider reaches the visible goal after the slider reset. A held button does not retrigger, and out-of-bounds input has no effect.}
\par\noindent\begin{minipage}{\linewidth}\phantomsection\label{app:task:F2}%
{\color{CardMind}\rule[-0.2ex]{2pt}{2.4ex}}\enspace{\small\bfseries\color{CardMind}F2}\enspace{\small\bfseries Color-block arrangement by feedback}\hfill{\footnotesize\itshape\color{Muted}Feedback-rule learning}\par\vspace{1.5pt}
\begin{minipage}[t]{0.555\linewidth}\vspace{0pt}\resizebox{\linewidth}{!}{\input{figures/task_cards/F2.tex}}\end{minipage}\hfill
\begin{minipage}[t]{0.43\linewidth}\vspace{0pt}\fontsize{8}{9.6}\selectfont\raggedright
\textbf{Instruction.} \par\vspace{2pt}\textbf{Success.} C\par\vspace{2pt}\textbf{E / M / H.} V\end{minipage}
\end{minipage}\par\vspace{2.6mm}
{Place one colored block in each slot, then press and release the confirmation button on the right. The brief feedback gives the number of correct positions. Infer the hidden permutation in as few attempts as possible.}%
{3/4/5 blocks and slots; 5/6/7 valid submissions.}%
{Correct arrangement found within the submission budget: one released, supported block per slot, stable $\ge$0.5\,s, then button released (press on an incomplete layout is not a submission).}
\par\noindent\begin{minipage}{\linewidth}\phantomsection\label{app:task:P1}%
{\color{CardMind}\rule[-0.2ex]{2pt}{2.4ex}}\enspace{\small\bfseries\color{CardMind}P1}\enspace{\small\bfseries Incremental meal orders}\hfill{\footnotesize\itshape\color{Muted}Process memory}\par\vspace{1.5pt}
\begin{minipage}[t]{0.555\linewidth}\vspace{0pt}\resizebox{\linewidth}{!}{\input{figures/task_cards/P1.tex}}\end{minipage}\hfill
\begin{minipage}[t]{0.43\linewidth}\vspace{0pt}\fontsize{8}{9.6}\selectfont\raggedright
\textbf{Instruction.} \par\vspace{2pt}\textbf{Success.} C\par\vspace{2pt}\textbf{E / M / H.} V\end{minipage}
\end{minipage}\par\vspace{2.6mm}
{\emph{Hard:} Remember the initial orders and two additional orders shown later. Place all requested ingredients into the indicated containers; each additional order adds to earlier demand.}%
{2/2/3 containers and 4/5/7 deposits. All levels start with two orders requesting four items. Later additions are none, one order adding one item, or two orders adding two and one items.}%
{Each container holds exactly the requested types and quantities.}
\par\noindent\begin{minipage}{\linewidth}\phantomsection\label{app:task:P2}%
{\color{CardMind}\rule[-0.2ex]{2pt}{2.4ex}}\enspace{\small\bfseries\color{CardMind}P2}\enspace{\small\bfseries Steamer-stack order restoration}\hfill{\footnotesize\itshape\color{Muted}Process memory}\par\vspace{1.5pt}
\begin{minipage}[t]{0.555\linewidth}\vspace{0pt}\resizebox{\linewidth}{!}{\input{figures/task_cards/P2.tex}}\end{minipage}\hfill
\begin{minipage}[t]{0.43\linewidth}\vspace{0pt}\fontsize{8}{9.6}\selectfont\raggedright
\textbf{Instruction.} \par\vspace{2pt}\textbf{Success.} C\par\vspace{2pt}\textbf{E / M / H.} V\end{minipage}
\end{minipage}\par\vspace{2.6mm}
{\emph{Hard:} Remove the lid, remember the dim sum order as you unstack the baskets from top to bottom onto the mats, then restack them in the same order from bottom to top on the left steamer pot and replace the lid without using any distractor baskets.}%
{3/4/4 layers to restore; 0/0/2 distractor baskets that look the same outside but hold different dumplings.}%
{Correct original baskets restacked in original order on the left pot, stable, lid placed; the final stack holds no extra basket (setting one aside temporarily is allowed).}
\par\noindent\begin{minipage}{\linewidth}\phantomsection\label{app:task:S1}%
{\color{CardMind}\rule[-0.2ex]{2pt}{2.4ex}}\enspace{\small\bfseries\color{CardMind}S1}\enspace{\small\bfseries Tray-layout restoration}\hfill{\footnotesize\itshape\color{Muted}Spatial memory}\par\vspace{1.5pt}
\begin{minipage}[t]{0.555\linewidth}\vspace{0pt}\resizebox{\linewidth}{!}{\input{figures/task_cards/S1.tex}}\end{minipage}\hfill
\begin{minipage}[t]{0.43\linewidth}\vspace{0pt}\fontsize{8}{9.6}\selectfont\raggedright
\textbf{Instruction.} \par\vspace{2pt}\textbf{Success.} C\par\vspace{2pt}\textbf{E / M / H.} V\end{minipage}
\end{minipage}\par\vspace{2.6mm}
{Remember the four colored slots, observe the carousel rotation, then place each colored block into its original matching slot after the carousel stops.}%
{1--2 slow rotation segments in one direction, 3--4 segments with reversals, or 5--6 faster segments with repeated reversals. All levels use four colors, four placements and the same placement tolerances.}%
{All four blocks in their original tray-relative cells.}
\par\noindent\begin{minipage}{\linewidth}\phantomsection\label{app:task:S2}%
{\color{CardMind}\rule[-0.2ex]{2pt}{2.4ex}}\enspace{\small\bfseries\color{CardMind}S2}\enspace{\small\bfseries Hidden-hazard navigation}\hfill{\footnotesize\itshape\color{Muted}Spatial memory}\par\vspace{1.5pt}
\begin{minipage}[t]{0.555\linewidth}\vspace{0pt}\resizebox{\linewidth}{!}{\input{figures/task_cards/S2.tex}}\end{minipage}\hfill
\begin{minipage}[t]{0.43\linewidth}\vspace{0pt}\fontsize{8}{9.6}\selectfont\raggedright
\textbf{Instruction.} \par\vspace{2pt}\textbf{Success.} C\par\vspace{2pt}\textbf{E / M / H.} V\end{minipage}
\end{minipage}\par\vspace{2.6mm}
{Remember the briefly lit forbidden cells. Press and release the arrow buttons to move one cell per press. Reach the ring without entering a forbidden cell or leaving the board. Do not touch the slider.}%
{$4\times4$/$5\times5$/$6\times6$ boards with 3/7/12 hazard cells; shortest routes of 6/10/14 steps in the first instance.}%
{Slider supported and inside the board throughout, never touching a hazard, fully in the goal and stable $\ge$0.5\,s (any hazard contact or leaving the board precludes success).}
\par\noindent\begin{minipage}{\linewidth}\phantomsection\label{app:task:M1}%
{\color{CardMind}\rule[-0.2ex]{2pt}{2.4ex}}\enspace{\small\bfseries\color{CardMind}M1}\enspace{\small\bfseries Workpiece tracking and route replay}\hfill{\footnotesize\itshape\color{Muted}Integrated memory}\par\vspace{1.5pt}
\begin{minipage}[t]{0.555\linewidth}\vspace{0pt}\resizebox{\linewidth}{!}{\input{figures/task_cards/M1.tex}}\end{minipage}\hfill
\begin{minipage}[t]{0.43\linewidth}\vspace{0pt}\fontsize{8}{9.6}\selectfont\raggedright
\textbf{Instruction.} \par\vspace{2pt}\textbf{Success.} C\par\vspace{2pt}\textbf{E / M / H.} V\end{minipage}
\end{minipage}\par\vspace{2.6mm}
{Remember the target location and station route. Track the target through the cover exchanges, uncover it, and move it through the required stations in order.}%
{4/4/6 stations and 3/5/7 exchanges, including 0/1/2 exchanges unrelated to the target. Replay requires 4/5/6 placements. Medium and Hard include at least one and two revisits, respectively.}%
{Target retrieved and initial station plus visits replayed in order, each placement released and stable (temporary tabletop placements do not count).}
\par\noindent\begin{minipage}{\linewidth}\phantomsection\label{app:task:M2}%
{\color{CardMind}\rule[-0.2ex]{2pt}{2.4ex}}\enspace{\small\bfseries\color{CardMind}M2}\enspace{\small\bfseries Card-matching game}\hfill{\footnotesize\itshape\color{Muted}Integrated memory}\par\vspace{1.5pt}
\begin{minipage}[t]{0.555\linewidth}\vspace{0pt}\resizebox{\linewidth}{!}{\input{figures/task_cards/M2.tex}}\end{minipage}\hfill
\begin{minipage}[t]{0.43\linewidth}\vspace{0pt}\fontsize{8}{9.6}\selectfont\raggedright
\textbf{Instruction.} \par\vspace{2pt}\textbf{Success.} C\par\vspace{2pt}\textbf{E / M / H.} V\end{minipage}
\end{minipage}\par\vspace{2.6mm}
{Press and release a card's yellow button to turn it face up, revealing at most two unmatched cards at a time. Matching pictures stay face up; mismatches turn face down automatically. Wait until the flip or pair result finishes before selecting another card. Remember the pictures and their positions and match every pair.}%
{$2\times2$/$2\times4$/$4\times4$ layouts with 2/4/8 pairs.}%
{All pairs matched and held face-up (mismatches are exploration, not failure).}
\needspace{6\baselineskip}
\subsubsection*{Act tasks}
\label{app:cards-act}
{\footnotesize\color{Muted}Ten tasks that test timing and precision, with the operation stated in the instruction.}\par\vspace{2mm}
\par\noindent\begin{minipage}{\linewidth}\phantomsection\label{app:task:DR1}%
{\color{CardAct}\rule[-0.2ex]{2pt}{2.4ex}}\enspace{\small\bfseries\color{CardAct}DR1}\enspace{\small\bfseries Conveyor pickup}\hfill{\footnotesize\itshape\color{Muted}Dynamic response}\par\vspace{1.5pt}
\begin{minipage}[t]{0.555\linewidth}\vspace{0pt}\resizebox{\linewidth}{!}{\input{figures/task_cards/DR1.tex}}\end{minipage}\hfill
\begin{minipage}[t]{0.43\linewidth}\vspace{0pt}\fontsize{8}{9.6}\selectfont\raggedright
\textbf{Instruction.} \par\vspace{2pt}\textbf{Success.} C\par\vspace{2pt}\textbf{E / M / H.} V\end{minipage}
\end{minipage}\par\vspace{2.6mm}
{\emph{Hard:} Pick up all 7 packages on the conveyor belt and place them in the central collection tray.}%
{3/5/7 packages arrive on the conveyor, and all of them must be collected into the central tray.}%
{All required packages stably deposited within budget, none missed (a package that leaves the reach region counts as missed).}
\par\noindent\begin{minipage}{\linewidth}\phantomsection\label{app:task:DR2}%
{\color{CardAct}\rule[-0.2ex]{2pt}{2.4ex}}\enspace{\small\bfseries\color{CardAct}DR2}\enspace{\small\bfseries Light-cued response}\hfill{\footnotesize\itshape\color{Muted}Dynamic response}\par\vspace{1.5pt}
\begin{minipage}[t]{0.555\linewidth}\vspace{0pt}\resizebox{\linewidth}{!}{\input{figures/task_cards/DR2.tex}}\end{minipage}\hfill
\begin{minipage}[t]{0.43\linewidth}\vspace{0pt}\fontsize{8}{9.6}\selectfont\raggedright
\textbf{Instruction.} \par\vspace{2pt}\textbf{Success.} C\par\vspace{2pt}\textbf{E / M / H.} V\end{minipage}
\end{minipage}\par\vspace{2.6mm}
{\emph{Hard:} Press the currently illuminated button within 2.6 seconds and release within 1 second. Respond correctly to all 10 cues; never press an unlit button.}%
{Cue-to-press deadlines are 4.0/3.2/2.6\,s. Button geometry, ten-cue count and release timing stay fixed.}%
{Ten correct press-release events, each released within 1\,s. A repeated cue needs a separate event.}
\par\noindent\begin{minipage}{\linewidth}\phantomsection\label{app:task:DR3}%
{\color{CardAct}\rule[-0.2ex]{2pt}{2.4ex}}\enspace{\small\bfseries\color{CardAct}DR3}\enspace{\small\bfseries Timed passage through cyclic gates}\hfill{\footnotesize\itshape\color{Muted}Dynamic response}\par\vspace{1.5pt}
\begin{minipage}[t]{0.555\linewidth}\vspace{0pt}\resizebox{\linewidth}{!}{\input{figures/task_cards/DR3.tex}}\end{minipage}\hfill
\begin{minipage}[t]{0.43\linewidth}\vspace{0pt}\fontsize{8}{9.6}\selectfont\raggedright
\textbf{Instruction.} \par\vspace{2pt}\textbf{Success.} C\par\vspace{2pt}\textbf{E / M / H.} V\end{minipage}
\end{minipage}\par\vspace{2.6mm}
{\emph{Hard:} Grasp the eraser, time the moving gates, and pass through the blue then the orange opening to place the eraser stably in the green tray behind them. Withdraw the whole gripper past the outer gate before they close. Do not touch any gate, frame or rail.}%
{One gate fully open for 6.6/6.0\,s per 11.2/10.6\,s cycle on Easy/Medium; on Hard, two gates with 18.4\,s and 9.2\,s cycles whose shared openings alternate between 3.6\,s, too short for the measured 5.1\,s action chain, and 6.4\,s.}%
{The eraser is carried through each open gate in order and left supported in the tray. Dropping or throwing it into the tray earns only the gate-passage credit. When a gate first closes after the eraser has passed, both arms must be clear of its passage on the near side (arm contact with a gate, frame or rail during that closure precludes success).}
\par\noindent\begin{minipage}{\linewidth}\phantomsection\label{app:task:DR4}%
{\color{CardAct}\rule[-0.2ex]{2pt}{2.4ex}}\enspace{\small\bfseries\color{CardAct}DR4}\enspace{\small\bfseries Tray carrying with loose balls}\hfill{\footnotesize\itshape\color{Muted}Dynamic response (stability)}\par\vspace{1.5pt}
\begin{minipage}[t]{0.555\linewidth}\vspace{0pt}\resizebox{\linewidth}{!}{\input{figures/task_cards/DR4.tex}}\end{minipage}\hfill
\begin{minipage}[t]{0.43\linewidth}\vspace{0pt}\fontsize{8}{9.6}\selectfont\raggedright
\textbf{Instruction.} \par\vspace{2pt}\textbf{Success.} C\par\vspace{2pt}\textbf{E / M / H.} V\end{minipage}
\end{minipage}\par\vspace{2.6mm}
{\emph{Hard:} Carry the tray of four balls by its handle from START to DESTINATION around the two staggered columns, keeping each ball inside its marked circle. Set the tray down stably, release it and retreat. Do not touch the balls or hit anything.}%
{2/3/4 balls, 0/1/2 pillars, and allowed drift radii of 30/25/20\,mm.}%
{All balls stay on the tray and inside their circles throughout; tray delivered within 20\,mm, vertical error $<$3\,mm, released, stable for the last 2\,s (a ball leaving its circle, with 0.5\,mm tolerance, is a latched failure).}
\par\noindent\begin{minipage}{\linewidth}\phantomsection\label{app:task:PR1}%
{\color{CardAct}\rule[-0.2ex]{2pt}{2.4ex}}\enspace{\small\bfseries\color{CardAct}PR1}\enspace{\small\bfseries Resistance adjustment}\hfill{\footnotesize\itshape\color{Muted}Precise execution}\par\vspace{1.5pt}
\begin{minipage}[t]{0.555\linewidth}\vspace{0pt}\resizebox{\linewidth}{!}{\input{figures/task_cards/PR1.tex}}\end{minipage}\hfill
\begin{minipage}[t]{0.43\linewidth}\vspace{0pt}\fontsize{8}{9.6}\selectfont\raggedright
\textbf{Instruction.} \par\vspace{2pt}\textbf{Success.} C\par\vspace{2pt}\textbf{E / M / H.} V\end{minipage}
\end{minipage}\par\vspace{2.6mm}
{Adjust the resistance between the left terminal and wiper to $R$\,$\Omega$, then release the slider and keep that reading for 0.5 seconds. \emph{($R$ per episode.)}}%
{One target per episode on a numbered major tick, an unnumbered minor tick, or between minor ticks: 7/21/21 distinct targets, e.g.\ 10/16/35.75\,$\Omega$.}%
{Reading within $\pm$0.2/0.15/0.12\,$\Omega$, held for 0.5\,s after release.}
\par\noindent\begin{minipage}{\linewidth}\phantomsection\label{app:task:PR2}%
{\color{CardAct}\rule[-0.2ex]{2pt}{2.4ex}}\enspace{\small\bfseries\color{CardAct}PR2}\enspace{\small\bfseries Voltage setting}\hfill{\footnotesize\itshape\color{Muted}Precise execution}\par\vspace{1.5pt}
\begin{minipage}[t]{0.555\linewidth}\vspace{0pt}\resizebox{\linewidth}{!}{\input{figures/task_cards/PR2.tex}}\end{minipage}\hfill
\begin{minipage}[t]{0.43\linewidth}\vspace{0pt}\fontsize{8}{9.6}\selectfont\raggedright
\textbf{Instruction.} \par\vspace{2pt}\textbf{Success.} C\par\vspace{2pt}\textbf{E / M / H.} V\end{minipage}
\end{minipage}\par\vspace{2.6mm}
{Adjust the output voltage to $V$ V, then release the knob and keep that reading for 1 second. \emph{($V$ per episode.)}}%
{One target per episode on a numbered major tick, an unnumbered minor tick, or between minor ticks: 12/33/33 distinct targets, e.g.\ 160/165/147\,V.}%
{Reading within $\pm$1/0.5/0.5\,V for 1\,s after release.}
\par\noindent\begin{minipage}{\linewidth}\phantomsection\label{app:task:PR3}%
{\color{CardAct}\rule[-0.2ex]{2pt}{2.4ex}}\enspace{\small\bfseries\color{CardAct}PR3}\enspace{\small\bfseries Pen tracing}\hfill{\footnotesize\itshape\color{Muted}Precise execution}\par\vspace{1.5pt}
\begin{minipage}[t]{0.555\linewidth}\vspace{0pt}\resizebox{\linewidth}{!}{\input{figures/task_cards/PR3.tex}}\end{minipage}\hfill
\begin{minipage}[t]{0.43\linewidth}\vspace{0pt}\fontsize{8}{9.6}\selectfont\raggedright
\textbf{Instruction.} \par\vspace{2pt}\textbf{Success.} C\par\vspace{2pt}\textbf{E / M / H.} V\end{minipage}
\end{minipage}\par\vspace{2.6mm}
{Grasp the marker and trace the complete visible path, then lift.}%
{A straight line, an S curve with two half-waves, or a curve with four half-waves. All paths are 350\,mm long with $\pm3$\,mm centerline tolerance. Curved paths have minimum radii of about 54/25\,mm.}%
{At least 98\% of the reference path lies within 3\,mm of ink and the RMS distance from the ink to the path is at most 2\,mm. Only ink from actual contact counts.}
\par\noindent\begin{minipage}{\linewidth}\phantomsection\label{app:task:PR4}%
{\color{CardAct}\rule[-0.2ex]{2pt}{2.4ex}}\enspace{\small\bfseries\color{CardAct}PR4}\enspace{\small\bfseries Cloth folding and flat placement}\hfill{\footnotesize\itshape\color{Muted}Precise execution}\par\vspace{1.5pt}
\begin{minipage}[t]{0.555\linewidth}\vspace{0pt}\resizebox{\linewidth}{!}{\input{figures/task_cards/PR4.tex}}\end{minipage}\hfill
\begin{minipage}[t]{0.43\linewidth}\vspace{0pt}\fontsize{8}{9.6}\selectfont\raggedright
\textbf{Instruction.} \par\vspace{2pt}\textbf{Success.} C\par\vspace{2pt}\textbf{E / M / H.} V\end{minipage}
\end{minipage}\par\vspace{2.6mm}
{\emph{Hard:} Fold the hexagonal cloth once into the displayed trapezoid, align its paired edges, release it fully and leave it stable.}%
{Midline fold into a two-layer rectangle; diagonal fold into a two-layer triangle; a single fold of a hexagon along its symmetry axis into a two-layer trapezoid. All pieces have the same area (400\,cm$^2$), mass, and material.}%
{Required fold made, layers and corners aligned within tolerance, inside the target frame, released and lying flat and stable.}
\par\noindent\begin{minipage}{\linewidth}\phantomsection\label{app:task:PR5}%
{\color{CardAct}\rule[-0.2ex]{2pt}{2.4ex}}\enspace{\small\bfseries\color{CardAct}PR5}\enspace{\small\bfseries Oriented stamping}\hfill{\footnotesize\itshape\color{Muted}Precise execution}\par\vspace{1.5pt}
\begin{minipage}[t]{0.555\linewidth}\vspace{0pt}\resizebox{\linewidth}{!}{\input{figures/task_cards/PR5.tex}}\end{minipage}\hfill
\begin{minipage}[t]{0.43\linewidth}\vspace{0pt}\fontsize{8}{9.6}\selectfont\raggedright
\textbf{Instruction.} \par\vspace{2pt}\textbf{Success.} C\par\vspace{2pt}\textbf{E / M / H.} V\end{minipage}
\end{minipage}\par\vspace{2.6mm}
{Pick up square stamp A and stamp once, fully inside the blue box, with its arrow aligned to the target arrow. Do not drag or stamp twice. Return the stamp to its holder in its original orientation and release it.}%
{Frames of 100/80/74\,mm and orientation tolerances of $\pm5^\circ$/$\pm2^\circ$/$\pm1^\circ$; targets sampled from 8/32/320 directions; minimum centered margins of 14.6/4.6/1.6\,mm.}%
{Exactly one imprint with all four rotated corners inside the frame and arrow within tolerance, no smear; stamp returned, gripper withdrawn. The arrow defines the orientation over the full 360\textdegree.}
\par\noindent\begin{minipage}{\linewidth}\phantomsection\label{app:task:PR6}%
{\color{CardAct}\rule[-0.2ex]{2pt}{2.4ex}}\enspace{\small\bfseries\color{CardAct}PR6}\enspace{\small\bfseries Domino chain to a bell}\hfill{\footnotesize\itshape\color{Muted}Precise execution}\par\vspace{1.5pt}
\begin{minipage}[t]{0.555\linewidth}\vspace{0pt}\resizebox{\linewidth}{!}{\input{figures/task_cards/PR6.tex}}\end{minipage}\hfill
\begin{minipage}[t]{0.43\linewidth}\vspace{0pt}\fontsize{8}{9.6}\selectfont\raggedright
\textbf{Instruction.} \par\vspace{2pt}\textbf{Success.} C\par\vspace{2pt}\textbf{E / M / H.} V\end{minipage}
\end{minipage}\par\vspace{2.6mm}
{Stand the dominos anywhere inside the continuous curve band, in order along the curve. Push the first one so the chain of contacts rings the bell. Repair fallen or stalled segments using as few pushes as possible.}%
{5/7/9 dominoes; heights of 10.24 (uniform)/8.32--12.00/7.68--13.44\,cm; band spans of 41.3/57.3/76.1\,cm.}%
{Dominoes released and held in the same valid order for 0.5\,s, the first one pushed, and the chain falling along the band to strike the bell.}
\needspace{6\baselineskip}
\subsubsection*{Mind2Act tasks}
\label{app:cards-mind2act}
{\footnotesize\color{Muted}Ten tasks that combine remembered state and physical outcomes in one loop.}\par\vspace{2mm}
\par\noindent\begin{minipage}{\linewidth}\phantomsection\label{app:task:CP01}%
{\color{CardCP}\rule[-0.2ex]{2pt}{2.4ex}}\enspace{\small\bfseries\color{CardCP}CP01}\enspace{\small\bfseries Conveyor meal assembly}\hfill{\footnotesize\itshape\color{Muted}Cumulative progress}\par\vspace{1.5pt}
\begin{minipage}[t]{0.555\linewidth}\vspace{0pt}\resizebox{\linewidth}{!}{\input{figures/task_cards/CP01.tex}}\end{minipage}\hfill
\begin{minipage}[t]{0.43\linewidth}\vspace{0pt}\fontsize{8}{9.6}\selectfont\raggedright
\textbf{Instruction.} \par\vspace{2pt}\textbf{Success.} C\par\vspace{2pt}\textbf{E / M / H.} V\end{minipage}
\end{minipage}\par\vspace{2.6mm}
{At the start, watch the reference meal trays pass on the conveyor and remember the three food types on each. Then prepare the corresponding empty meal trays on the table and fill each with the same three food types from the continuously moving supply.}%
{1/2/3 orders with 3/6/9 food items. Easy provides one meal tray. Medium and Hard require retrieving 2 and 3 meal trays from the left stack, respectively. Belt speed remains 0.025\,m/s across all levels.}%
{All requested foods stably in the correct meal trays; Medium and Hard also require the meal trays retrieved from the stack (no recirculation or replacement supply).}
\par\noindent\begin{minipage}{\linewidth}\phantomsection\label{app:task:CP02}%
{\color{CardCP}\rule[-0.2ex]{2pt}{2.4ex}}\enspace{\small\bfseries\color{CardCP}CP02}\enspace{\small\bfseries Rotating-cake decoration}\hfill{\footnotesize\itshape\color{Muted}Prerequisites and repair}\par\vspace{1.5pt}
\begin{minipage}[t]{0.555\linewidth}\vspace{0pt}\resizebox{\linewidth}{!}{\input{figures/task_cards/CP02.tex}}\end{minipage}\hfill
\begin{minipage}[t]{0.43\linewidth}\vspace{0pt}\fontsize{8}{9.6}\selectfont\raggedright
\textbf{Instruction.} \par\vspace{2pt}\textbf{Success.} C\par\vspace{2pt}\textbf{E / M / H.} V\end{minipage}
\end{minipage}\par\vspace{2.6mm}
{\emph{Hard:} Remember each macaron and its place on the still cake shown for three seconds. Open the ingredient-box lid, press the green button and recreate the arrangement as the cake rotates. Then press the red button, wait for the cake to stop, and place the cake cover over it.}%
{4/6/6 macarons at speeds of $12/20/28^{\circ}$/s. Easy starts open and running. Medium and Hard add opening the box lid and pressing the green button. Hard also adds the stopping and covering sequence.}%
{Full layout placed and all preparation and cleanup steps done; Hard also requires the turntable stopped and at rest before the cover is placed.}
\par\noindent\begin{minipage}{\linewidth}\phantomsection\label{app:task:CP03}%
{\color{CardCP}\rule[-0.2ex]{2pt}{2.4ex}}\enspace{\small\bfseries\color{CardCP}CP03}\enspace{\small\bfseries Moving-rack loading}\hfill{\footnotesize\itshape\color{Muted}Cumulative progress}\par\vspace{1.5pt}
\begin{minipage}[t]{0.555\linewidth}\vspace{0pt}\resizebox{\linewidth}{!}{\input{figures/task_cards/CP03.tex}}\end{minipage}\hfill
\begin{minipage}[t]{0.43\linewidth}\vspace{0pt}\fontsize{8}{9.6}\selectfont\raggedright
\textbf{Instruction.} \par\vspace{2pt}\textbf{Success.} C\par\vspace{2pt}\textbf{E / M / H.} V\end{minipage}
\end{minipage}\par\vspace{2.6mm}
{\emph{Hard:} Load 7 upright plates into the moving empty slots without disturbing other plates. Tilt the lever fully toward the back of the machine, away from the robot, release it, then press the green button before loading.}%
{3/5/7 plates at 12/16/20\,mm/s. All levels start with the green button; Medium first sets the lever upright in the middle, Hard tilts it fully back. Support gaps are approximately 36.5/30.5/26.5\,mm. Hard reverses direction with a 0.6\,s end transition while insertions occur at cruise speed.}%
{Every plate supported by both side supports and the base, earlier plates undisturbed.}
\par\noindent\begin{minipage}{\linewidth}\phantomsection\label{app:task:CP04}%
{\color{CardCP}\rule[-0.2ex]{2pt}{2.4ex}}\enspace{\small\bfseries\color{CardCP}CP04}\enspace{\small\bfseries Temperature-controlled cooking}\hfill{\footnotesize\itshape\color{Muted}Feedback-driven adjustment}\par\vspace{1.5pt}
\begin{minipage}[t]{0.555\linewidth}\vspace{0pt}\resizebox{\linewidth}{!}{\input{figures/task_cards/CP04.tex}}\end{minipage}\hfill
\begin{minipage}[t]{0.43\linewidth}\vspace{0pt}\fontsize{8}{9.6}\selectfont\raggedright
\textbf{Instruction.} \par\vspace{2pt}\textbf{Success.} C\par\vspace{2pt}\textbf{E / M / H.} V\end{minipage}
\end{minipage}\par\vspace{2.6mm}
{\emph{Hard:} Open the drawer. Pour each required jar into the pot in the remembered recipe order, returning jars upright and caps to the tray, then cover the pot. Turn the black knob to HEAT and press the green button; within 2 s after 80\,\textdegree C, switch to KEEP WARM, then press it again.}%
{2/2/3 ordered ingredients from four jars, and HEAT must be switched to KEEP WARM within 4/3/2\,s after the pot reaches 80\,\textdegree C (green band up to 90/87.5/85\,\textdegree C). All levels return the jars upright and their caps to the tray; Medium and Hard add opening the drawer and covering the pot. Geometric tolerances remain unchanged.}%
{Ingredients added in remembered order before HEAT, KEEP WARM selected within the window before shutdown, and jars and caps returned; Medium and Hard also require the drawer opened and the pot covered.}
\par\noindent\begin{minipage}{\linewidth}\phantomsection\label{app:task:CP05}%
{\color{CardCP}\rule[-0.2ex]{2pt}{2.4ex}}\enspace{\small\bfseries\color{CardCP}CP05}\enspace{\small\bfseries Piano replay}\hfill{\footnotesize\itshape\color{Muted}Sequence and plan execution}\par\vspace{1.5pt}
\begin{minipage}[t]{0.555\linewidth}\vspace{0pt}\resizebox{\linewidth}{!}{\input{figures/task_cards/CP05.tex}}\end{minipage}\hfill
\begin{minipage}[t]{0.43\linewidth}\vspace{0pt}\fontsize{8}{9.6}\selectfont\raggedright
\textbf{Instruction.} \par\vspace{2pt}\textbf{Success.} C\par\vspace{2pt}\textbf{E / M / H.} V\end{minipage}
\end{minipage}\par\vspace{2.6mm}
{\emph{Hard:} Set volume to MAX and release the slider. Pick up both side mallets, press the triangle-marked button and watch the demonstration while holding both, then reproduce the twelve key presses in order using either mallet. Return and release both mallets, release all keys, withdraw your arms, and keep volume at MAX.}%
{7/14/21 candidate keys, 8/8/12 events and 1/2/2 mallets. Easy uses the central mallet; Medium and Hard pick up and return both side mallets together, and Hard also sets the volume slider from MIN to MAX and keeps it there. Either mallet may strike.}%
{Effective strikes in the remembered sequence (rhythm need not match), mallets returned and arms withdrawn; Hard also requires the volume kept at MAX.}
\par\noindent\begin{minipage}{\linewidth}\phantomsection\label{app:task:CP06}%
{\color{CardCP}\rule[-0.2ex]{2pt}{2.4ex}}\enspace{\small\bfseries\color{CardCP}CP06}\enspace{\small\bfseries Produce weighing on a balance}\hfill{\footnotesize\itshape\color{Muted}Feedback-driven adjustment}\par\vspace{1.5pt}
\begin{minipage}[t]{0.555\linewidth}\vspace{0pt}\resizebox{\linewidth}{!}{\input{figures/task_cards/CP06.tex}}\end{minipage}\hfill
\begin{minipage}[t]{0.43\linewidth}\vspace{0pt}\fontsize{8}{9.6}\selectfont\raggedright
\textbf{Instruction.} \par\vspace{2pt}\textbf{Success.} C\par\vspace{2pt}\textbf{E / M / H.} V\end{minipage}
\end{minipage}\par\vspace{2.6mm}
{\emph{Hard:} With the empty scale stable, press ZERO. Move all fruit and vegetables from the left tray onto the left scale pan. Balance the scale using the weights on the right and by physically moving the sliding weight, then release them and keep the pointer in the green zone.}%
{1/2/3 produce items; Medium and Hard first press ZERO on the empty scale; weight sets of $2\times50{+}2\times100$\,g/50--400\,g/50--400\,g, with a 0--40\,g sliding weight added on Hard; green zones of $\pm10$/$\pm8$/$\pm5$\,g; 9/16/80 candidate states, with a unique solution on Hard.}%
{All produce on the left pan, the weights needed for balance on the right pan, nothing dropped, and the pointer inside the green zone and stable for 1\,s; Medium and Hard also require ZERO pressed on the empty scale, and Hard the sliding weight at its target position.}
\par\noindent\begin{minipage}{\linewidth}\phantomsection\label{app:task:CP07}%
{\color{CardCP}\rule[-0.2ex]{2pt}{2.4ex}}\enspace{\small\bfseries\color{CardCP}CP07}\enspace{\small\bfseries Bridge building for a toy car}\hfill{\footnotesize\itshape\color{Muted}Sequence and plan execution}\par\vspace{1.5pt}
\begin{minipage}[t]{0.555\linewidth}\vspace{0pt}\resizebox{\linewidth}{!}{\input{figures/task_cards/CP07.tex}}\end{minipage}\hfill
\begin{minipage}[t]{0.43\linewidth}\vspace{0pt}\fontsize{8}{9.6}\selectfont\raggedright
\textbf{Instruction.} \par\vspace{2pt}\textbf{Success.} C\par\vspace{2pt}\textbf{E / M / H.} V\end{minipage}
\end{minipage}\par\vspace{2.6mm}
{\emph{Hard:} Build a stable bridge across the gap with continuous wheel paths on both sides. Choose suitable pieces by shape and size from the stock, which includes extra blocks, and leave unused blocks aside. Release the bridge pieces, then press and release the button and let the car drive across to the opposite bank.}%
{River widths of 18/28/36\,cm; 1 pier and 2 spans/2 piers and 3 spans/2 unequal piers, sloped spans, and U-shaped ramp supports; stock of 8/18/28 pieces, of which 8/13/13 are needed.}%
{Piers, caps and deck stably in place, button triggered, car upright and across the finish (unused distractor stock allowed).}
\par\noindent\begin{minipage}{\linewidth}\phantomsection\label{app:task:CP08}%
{\color{CardCP}\rule[-0.2ex]{2pt}{2.4ex}}\enspace{\small\bfseries\color{CardCP}CP08}\enspace{\small\bfseries Block structure from a shown blueprint}\hfill{\footnotesize\itshape\color{Muted}Prerequisites and repair}\par\vspace{1.5pt}
\begin{minipage}[t]{0.555\linewidth}\vspace{0pt}\resizebox{\linewidth}{!}{\input{figures/task_cards/CP08.tex}}\end{minipage}\hfill
\begin{minipage}[t]{0.43\linewidth}\vspace{0pt}\fontsize{8}{9.6}\selectfont\raggedright
\textbf{Instruction.} \par\vspace{2pt}\textbf{Success.} C\par\vspace{2pt}\textbf{E / M / H.} V\end{minipage}
\end{minipage}\par\vspace{2.6mm}
{\emph{Hard:} Remember the briefly shown front view, then build the structure from memory, centred in the marked area and aligned with the table reference line. If the base is misplaced, move the beam above it aside and rebuild; put the roof ridge-up on the cylinders. Use only reference-colour blocks and release everything.}%
{3 cuboids in 2 layers/3 cuboids and 1 cylinder in 3 layers/6 blocks in 4 layers with a prism roof; 0/1/2 distractors; 6/4/4\,mm tolerance; 1/1/5\,s hold after release.}%
{Colors, shapes, positions, orientation and support match the blueprint; beam and roof each borne by two blocks; stable 1\,s (Hard 5\,s) after release (holding the structure up does not count; $90^\circ$ roof error rejected).}
\par\noindent\begin{minipage}{\linewidth}\phantomsection\label{app:task:CP09}%
{\color{CardCP}\rule[-0.2ex]{2pt}{2.4ex}}\enspace{\small\bfseries\color{CardCP}CP09}\enspace{\small\bfseries Whiteboard restoration}\hfill{\footnotesize\itshape\color{Muted}Prerequisites and repair}\par\vspace{1.5pt}
\begin{minipage}[t]{0.555\linewidth}\vspace{0pt}\resizebox{\linewidth}{!}{\input{figures/task_cards/CP09.tex}}\end{minipage}\hfill
\begin{minipage}[t]{0.43\linewidth}\vspace{0pt}\fontsize{8}{9.6}\selectfont\raggedright
\textbf{Instruction.} \par\vspace{2pt}\textbf{Success.} C\par\vspace{2pt}\textbf{E / M / H.} V\end{minipage}
\end{minipage}\par\vspace{2.6mm}
{\emph{Hard:} Remember the briefly shown original drawing on the whiteboard. Erase the added strokes, uncap the pens of the colors needed, and redraw any original strokes you erased, so the board matches the original.}%
{2/3/4 original elements in 1/2/3 colors; added strokes of one line/a long and a short crossing line/two long lines and an extra circle. The board starts on a raised stand; the reference procedure first lays it flat on the table, though only the final board is scored.}%
{Final board matches the original in color, shape and layout: extra strokes removed, over-erased parts redrawn; Medium/Hard include uncapping.}
\par\noindent\begin{minipage}{\linewidth}\phantomsection\label{app:task:CP10}%
{\color{CardCP}\rule[-0.2ex]{2pt}{2.4ex}}\enspace{\small\bfseries\color{CardCP}CP10}\enspace{\small\bfseries Return or dodge incoming balls}\hfill{\footnotesize\itshape\color{Muted}Predictive decision}\par\vspace{1.5pt}
\begin{minipage}[t]{0.555\linewidth}\vspace{0pt}\resizebox{\linewidth}{!}{\input{figures/task_cards/CP10.tex}}\end{minipage}\hfill
\begin{minipage}[t]{0.43\linewidth}\vspace{0pt}\fontsize{8}{9.6}\selectfont\raggedright
\textbf{Instruction.} \par\vspace{2pt}\textbf{Success.} C\par\vspace{2pt}\textbf{E / M / H.} V\end{minipage}
\end{minipage}\par\vspace{2.6mm}
{Predict each ball from the launcher's settled aim. After two practice balls, use the sliding blocker to return every white ball to the far net and let every red ball pass untouched into the near net. Wait for each ball to settle before the next shot.}%
{5/8/12 scored balls, of which 2/4/6 are red; shots (direct/single-bounce/double-bounce) of 3/2/0 on Easy, 2/4/2 on Medium and 2/5/5 on Hard; countdowns of 4/3/2\,s.}%
{Every scored ball correct: white returned to the far net, red cleared into the near net (two practice balls unscored).}

\endgroup

\end{document}